\documentclass{article}

\usepackage{iclr2027_conference,times}

\usepackage{hyperref}
\usepackage{url}
\usepackage{fvextra}
\usepackage{makecell}
\usepackage{amsmath,amssymb}
\usepackage{graphicx}
\usepackage{booktabs}
\usepackage{placeins}
\usepackage{multirow}
\usepackage{caption}
\usepackage{xcolor}
\graphicspath{{figures/paper_figures/}}

\newcommand{\ci}[2]{#1{\scriptsize$\pm$}{\scriptsize#2}}
\newcommand{\bci}[2]{\textbf{#1}{\scriptsize$\pm$}{\scriptsize\textbf{#2}}}
\newcommand{\probesd}{Probe-SD}
\newcommand{\probesdq}{Probe-SD-Q}

\title{From Hidden States to Stated Confidence: \\ Self-Distillation for Reasoning Models}

\makeatletter
\let\iclr@orig@maketitle\@maketitle
\def\@maketitle{%
  \def\@makefnmark{\hbox{}}%
  \long\def\@makefntext##1{\parindent 1em\noindent ##1}%
  \iclr@orig@maketitle}
\makeatother

\author{\parbox{\dimexpr\textwidth-2\tabcolsep\relax}{\centering
Yadong Xi\textsuperscript{1}\thanks{Corresponding author: \texttt{xiyadong@corp.netease.com}} \quad Rongsheng Zhang\textsuperscript{1} \quad Tangjie Lv\textsuperscript{1} \quad Ziyang Luo\textsuperscript{2} \quad Ruochen Zhao\textsuperscript{3} \\[0.4em]
\textsuperscript{1}NetEase \quad \textsuperscript{2}Amazon \quad \textsuperscript{3}Singapore University of Technology and Design}}

\iclrfinalcopy
\begin{document}

\maketitle

\lhead{}

\begin{abstract}
Reinforcement learning with binary correctness rewards trains correctness, not calibrated confidence.
The confidence that reasoning models verbalize is overconfident and tracks how willing a model is to
commit to an answer rather than how likely the answer is to be right, so post-hoc rescaling rarely
transfers. We look inside the model instead. On factual question answering, a linear probe on the
hidden state between the chain of thought and the answer has 5--38$\times$ lower expected calibration
error (ECE) than the verbalized score across four benchmarks and two model families. Since a probe
needs hidden-state access, we introduce \emph{probe-guided self-distillation} (\probesd): the probe
scores the model's own sampled traces, each trace's stated confidence is overwritten with that score,
and the base checkpoint is finetuned on the result, so the model itself states calibrated confidence
at test time. On Qwen3-14B, \probesd{} cuts out-of-domain ECE from 0.542 to 0.113, beats post-hoc
recalibration and distillation of empirical-accuracy labels, and keeps the confidence stated while
reasoning consistent with the final score. Its question-level variant, \probesdq{}, lowers ECE
further to 0.099, and the gains carry over to multiple-choice QA with shuffled options. The probe
mainly tells which questions the model knows, not which of its sampled answers is right, so as an
answer selector it nearly ties majority voting. We therefore use it as a source of confidence.
\end{abstract}

\section{Introduction}
\label{sec:intro}

Reasoning models---trained with reinforcement learning to think in natural language before
answering \citep{openai2024o1,guo2025deepseekr1}---account for much of the recent progress in
language modeling. A common recipe, RLVR, rewards binary correctness: a lucky guess pays as
well as a confident correct answer, abstention costs as much as an error, and calibration receives
no direct reward \citep{damani2025binaryrewardstraininglms}.
Where the
stakes are high, being right is not enough: a model must also say when it is unsure
\citep{xiong2024llmsexpressuncertaintyempirical}. This is harder on knowledge-heavy factual tasks,
where uncertainty estimation is less reliable than on reasoning-intensive tasks such as mathematics
\citep{tao2025revisitinguncertaintyestimationcalibration}. We therefore study a model's
\emph{verbalized confidence} on factual QA.

Miscalibration here runs deeper than a mis-set scale.
\citet{kumaran2026reportedconfidencellmstracks} finds that a stated confidence predicts whether a
model will submit or abstain far better than whether it will be right, including in reasoning
models: the number reads as a decision about commitment rather than as evidence.
\citet{ji2025calibratingverbaluncertainty} find the same for the wording of an answer: expressed
uncertainty lies along one linear direction, only loosely tied to semantic uncertainty. \textbf{The channel that expresses
uncertainty has its own cause, and that cause is not the evidence for the answer.}
Consequently, temperature scaling \citep{guo2017calibration} and isotonic regression
\citep{zadrozny2002transforming} rescale a signal that measures something else: they fit one
distribution, transfer poorly, and add no discriminative power. This motivates changing the source
of verbalized confidence rather than rescaling it. That raises two questions: does a reasoning model hold an
internal signal calibrated to correctness, and can the model be taught to say it?

\textbf{A calibration signal exists in hidden states.} Hidden states linearly encode output correctness
\citep{alain2018understandingintermediatelayersusing,marks2024geometrytruthemergentlinear,orgad2025llmsknowshowintrinsic,gekhman2025insideouthiddenfactualknowledge},
but prior evidence comes largely from non-reasoning models and focuses on
\emph{discrimination} rather than \emph{calibration}. Across 80 models, the two are
nearly rank-uncorrelated \citep{tao2025revisitinguncertaintyestimationcalibration}. We address this
gap on factual QA with reasoning models. A linear probe at the boundary between the chain of
thought and the answer is substantially better calibrated: its ECE is far lower than that of the
verbalized score and stays low on held-out benchmarks.

\textbf{But the probe is a poor answer selector.} Used as a top-score selector among $N$ samples, the
same linear probe nearly ties majority voting and stays far below the oracle, echoing
\citet{servedio2025hiddenstateshidingsomething}. Its high overall AUROC mainly reflects which
questions the model knows. Within a question, it ranks correct above incorrect samples only weakly.
\textbf{It is therefore more useful for estimating how likely the model is to be right than for
identifying the correct answer among candidates.}

\begin{figure}[t]
\centering
\includegraphics[width=\linewidth]{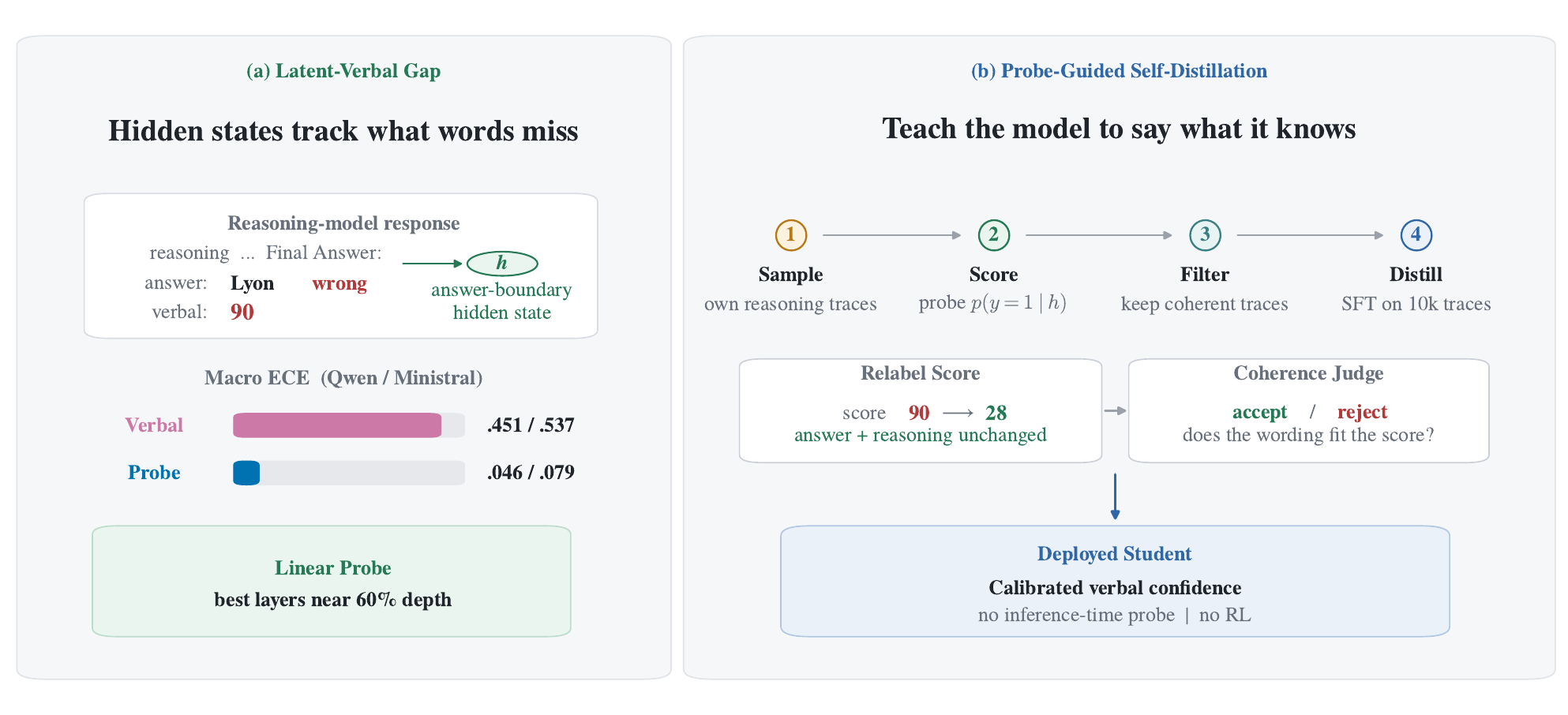}
\caption{Overview. (a) On factual QA, a reasoning model's verbalized confidence is overconfident
(here 90 for a wrong answer), while a linear probe on the hidden state at the answer boundary is
calibrated (ECE averaged over four benchmarks). (b) \probesd{}: the probe scores the model's own
traces, the confidence each trace states is overwritten with the probe score, a judge keeps the
traces whose wording still fits the new score, and the base checkpoint of the same family is
finetuned on them.}
\label{fig:pipeline}
\end{figure}

\textbf{Probe-guided self-distillation (\probesd).} Using a probe at inference requires hidden-state
access, which text-only APIs do not provide, so deployment-oriented evaluations often exclude internal-state
approaches \citep{tao2025revisitinguncertaintyestimationcalibration}. We therefore move it offline: the reasoning model
samples its own traces, the probe scores them, the stated confidence is overwritten with the probe
score, and the base checkpoint of the same family is finetuned on the result
(Figure~\ref{fig:pipeline}), so that only the model remains at test time. A suitable label must
come from the model's own knowledge and be calibrated. An external judge fails the first. Empirical
accuracy over the model's samples is calibrated and even estimates accuracy better, yet trains a less
calibrated student. On Qwen, the probe label wins mainly through how it ranks the training traces
(\S\ref{sec:main-results}). It reads what the model's hidden state encodes, which the student can
learn. Empirical accuracy also carries variation that this state does not encode, which the student
can only memorize. Our contributions are the following:
\begin{itemize}
\item \textbf{Measurement.} On two model families and four benchmarks, a probe at the answer
boundary outperforms the verbalized score on all 24 ECE, AUROC and Brier comparisons, with ECE 5--38$\times$
lower (\S\ref{sec:probe-calibrated}--\S\ref{sec:probe-robust}).
\item \textbf{Boundary.} Used as a top-score selector, the probe nearly ties majority voting, because it
ranks answers within a question only weakly (\S\ref{sec:not-selection}). We therefore use it as a
confidence, not a selector.
\item \textbf{Method.} Supervised finetuning alone transfers the calibration signal into verbalized confidence:
Qwen3-14B's ECE drops from 0.178 to 0.024 in-domain and from 0.542 to 0.113 out-of-domain, and
the question-level variant \probesdq{} lowers it further (\S\ref{sec:main-results}). Without
relabeling, the student stays nearly as miscalibrated as the original model (\S\ref{sec:ablation}). Unlike
ConfTuner, the student's reasoning and final score stay consistent. On multiple-choice QA,
\probesd{} again beats empirical-accuracy labels, and its confidence depends half as much on the
option order (\S\ref{sec:mcqa-main}).
\end{itemize}

\section{Related Work}
\label{sec:related}

\textbf{Verbalized confidence and sampling-consistency proxies.} A model can state a numerical
confidence outright or hedge in its wording
\citep{yang2024verbalizedconfidencescoresllms,xiong2024llmsexpressuncertaintyempirical,yona2024can,belem2024perceptions}.
Stated numbers beat token-level conditional probabilities
\citep{kadavath2022languagemodelsmostlyknow,tian2023justask} but are systematically overconfident
\citep{xiong2024llmsexpressuncertaintyempirical,mei2025reasoning}. A reasoning mode only softens the error
\citep{yoon2025reasoningmodelsbetterexpress,tao2025revisitinguncertaintyestimationcalibration}.
\citet{kumaran2026reportedconfidencellmstracks} identifies the failure: the number predicts
submit-or-abstain decisions better than correctness, so it is behavioral rather than
evidential. Sampling consistency instead scores a question rather than
a trace: self-consistency \citep{wang2023selfconsistency} and semantic entropy
\citep{kuhn2023semantic,farquhar2024detecting} cost an order of magnitude more inference, and
semantic-entropy probes \citep{kossen2024semanticentropyprobes} cut that cost. Offline distillation can amortize self-consistency into a single-generation predictor \citep{zollo2026unsupervisedconfidencecalibration}. Trace length \citep{devic2025tracelength} is free but unscaled. None of these signals is both cheap at inference and on a trustworthy
probability scale. We obtain both by teaching the model to state a calibrated internal signal.

\textbf{Training a model to express calibrated confidence.} These methods differ mainly in what
supplies the target. Most derive it from answer correctness. \citet{mielke2022reducing} recalibrate
the wording of confidence, \citet{lin2022teachingmodelsexpressuncertainty} and
\citet{jang2026confidence} supervise the stated number with empirical accuracy over samples, and
ConfTuner \citep{li2025conftuner} optimizes the final confidence token with a tokenized Brier score.
Our EA-SD baseline uses the empirical-accuracy label within our pipeline, and
\citet{jang2026confidence} also find that calibration does not transfer across task families, which
motivates our factual-QA scope. ConfTuner was developed for non-reasoning models and leaves any
confidence stated in a chain of thought unsupervised. LACIE \citep{stengel2024lacie} targets what a
listener accepts, and HALT \citep{franzmeyer2026highaccuracytalkhalt} uses fragment-level correctness
to learn when to abstain. A reinforcement-learning branch optimizes the stated confidence online:
RLCR \citep{damani2025binaryrewardstraininglms} adds a Brier term to the binary correctness reward,
and SaySelf \citep{xu2024sayselfteachingllmsexpress}, \citet{liu2026rlmf},
\citet{guo2026llmsexpressuncertaintyexplicitly}, \citet{stangel2025rewarding} and \citet{ma2026dcpo}
proceed similarly, at the cost of online sampling and reward design. Few methods use internal
signals. \citet{ji2025calibratingverbaluncertainty} steer activations without correctness targets. In
parallel work, \citet{cacioli2026makingllmssay} uses a hidden-state probe as the confidence target via
LoRA on instruction-tuned models, focusing on discrimination rather than calibration. We instead
distill a calibrated probe into the confidence that reasoning models state, with supervised
finetuning alone.

\textbf{Internal-state probes.} Hidden states carry a linearly readable correctness signal, used for
truthfulness discrimination
\citep{azaria2023internalstatellmknows,orgad2025llmsknowshowintrinsic,burns2023discovering}, activation intervention
\citep{li2024inferencetimeinterventionelicitingtruthful}, and pre-generation awareness of knowledge
boundaries \citep{ni2025knowledgeboundary}. Its strength is contested:
\citet{gekhman2025insideouthiddenfactualknowledge} argue that internal factual knowledge exceeds
what surfaces in the output, while \citet{servedio2025hiddenstateshidingsomething} show that
internal states pick the correct response out of a model's own samples only weakly. These probes
use non-reasoning models and validate discrimination rather than calibration. The two are
nearly rank-uncorrelated across models \citep{tao2025revisitinguncertaintyestimationcalibration}.
\citet{kapoor2024taught} compare probe and verbalized confidence on non-reasoning models, find that
frozen probes generalize poorly, and instead train a separate correctness model. The probes
that do report ECE either target mathematics for early stopping
\citep{zhang2025reasoningmodelsknowtheyre,li2026calibratingoverconfidencesacrificingconfidence}, read
a judge model's assessment of someone else's output
\citep{radharapu2025calibratingllmjudgeslinear}, or reside at the last layer of a non-reasoning model
\citep{liu2024enhancinglanguagemodelfactuality}. All keep the probe in the deployment
path. We instead apply calibration probes to factual QA with reasoning models,
distill their signal offline, and remove the probe from deployment.

\textbf{Self-distillation.} Within knowledge distillation \citep{hinton2015distilling},
self-distillation includes STaR \citep{zelikman2022star} and ReST$^{\text{EM}}$
\citep{singh2024restem}, which bootstrap from self-generated, externally verified trajectories, and
SSD \citep{zhang2026ssd}, which improves performance using only self-generated data. We keep the
paradigm and change the target: probe scores replace the
verbalized confidence score, and the student starts from the base model.

\section{A Calibration Signal in Reasoning-Model Hidden States}
\label{sec:signal}

We test whether reasoning models hold a trustworthy internal correctness signal. Across four
benchmarks and two model families, a post-CoT probe is substantially better calibrated, but used as a
top-score selector it nearly ties majority voting. We evaluate on four factual QA benchmarks---TriviaQA \citep{joshi2017triviaqa}, EntityQuestions \citep{sciavolino2021entityquestions},
NQ-open \citep{kwiatkowski2019natural} and SimpleQA-Verified
\citep{wei2024simpleqa,haas2025simpleqa}---using 1{,}000 questions and 10 traces each,
on Qwen3-14B \citep{qwen3} and Ministral-3-8B-Reasoning-2512 \citep{liu2026ministral3} at
$T{=}0.6$, top-$k{=}20$, top-$p{=}0.95$. We fit an $\ell_2$-regularized logistic probe at the
\emph{last pre-answer token} to trace correctness on 10{,}000 TriviaQA training questions. TriviaQA
is in-domain and the other benchmarks are out-of-domain. Layer, readout, and regularization are
chosen on TriviaQA validation data alone (Appendix~\ref{app:grid}). Qwen3-32B-Instruct grades answers. Because
all labels come from it, we manually audited 700 samples and validated its reliability (Appendix~\ref{app:judge}). We report
ECE \citep{naeini2015obtaining}, AUROC, Brier \citep{Brier1950VERIFICATIONOF} and Coverage (Cov.; the fraction of traces for which the confidence source produces a score); Appendix~\ref{app:metrics} gives definitions.
AUROC pools all traces, so it rewards separating the questions a model knows from those it does
not. The \emph{within-question AUROC} is computed only among the samples of a question that has both
correct and incorrect ones, so it measures whether a score finds the right answer to that question.

\subsection{Internal States Give Better-Calibrated Confidence}
\label{sec:probe-calibrated}

\begin{table}[t]
\caption{Verbalized score (V) versus hidden-state probe (P). All calibration metrics are computed 
on the subset that the source can score. Brier decomposition and missing-score sensitivity analysis are in Appendix~\ref{app:brier}.}
\label{tab:probe-vs-verbal}
\centering\small
\setlength{\tabcolsep}{4pt}
\renewcommand{\arraystretch}{1.15}
\resizebox{0.95\textwidth}{!}{%
\begin{tabular}{@{}ll ccccc ccccc@{}}
\toprule
& & \multicolumn{5}{c}{\emph{Qwen3-14B}} & \multicolumn{5}{c}{\emph{Ministral-3-8B}}\\
\cmidrule(lr){3-7}\cmidrule(l){8-12}
Dataset & & Acc & ECE $\downarrow$ & AUROC $\uparrow$ & Brier $\downarrow$ & Cov.\ $\uparrow$
            & Acc & ECE $\downarrow$ & AUROC $\uparrow$ & Brier $\downarrow$ & Cov.\ $\uparrow$\\
\midrule
\multirow{2}{*}{TriviaQA (ID)}
 & V & \multirow{2}{*}{.729} & \ci{.178}{.020} & \ci{.825}{.017} & \ci{.184}{.017} & .980
     & \multirow{2}{*}{.664} & \ci{.247}{.024} & \ci{.669}{.015} & \ci{.263}{.021} & .957\\
 & P & & \bci{.015}{.009} & \bci{.948}{.009} & \bci{.081}{.008} & \textbf{1.000}
     & & \bci{.023}{.009} & \bci{.919}{.013} & \bci{.105}{.009} & \textbf{.977}\\
\addlinespace[2pt]
\multirow{2}{*}{EntityQ}
 & V & \multirow{2}{*}{.267} & \ci{.529}{.022} & \ci{.812}{.017} & \ci{.455}{.018} & .903
     & \multirow{2}{*}{.207} & \ci{.627}{.022} & \ci{.734}{.016} & \ci{.564}{.019} & .950\\
 & P & & \bci{.014}{.009} & \bci{.913}{.014} & \bci{.097}{.010} & \textbf{1.000}
     & & \bci{.043}{.015} & \bci{.902}{.017} & \bci{.095}{.009} & \textbf{.999}\\
\addlinespace[2pt]
\multirow{2}{*}{NQ-open}
 & V & \multirow{2}{*}{.440} & \ci{.436}{.029} & \ci{.705}{.021} & \ci{.416}{.025} & .723
     & \multirow{2}{*}{.370} & \ci{.513}{.026} & \ci{.643}{.018} & \ci{.490}{.022} & .798\\
 & P & & \bci{.078}{.017} & \bci{.826}{.020} & \bci{.177}{.012} & \textbf{1.000}
     & & \bci{.106}{.018} & \bci{.816}{.020} & \bci{.186}{.012} & \textbf{.994}\\
\addlinespace[2pt]
\multirow{2}{*}{SimpleQA}
 & V & \multirow{2}{*}{.066} & \ci{.661}{.016} & \ci{.627}{.042} & \ci{.532}{.013} & .849
     & \multirow{2}{*}{.061} & \ci{.761}{.013} & \ci{.531}{.040} & \ci{.671}{.012} & .887\\
 & P & & \bci{.075}{.012} & \bci{.704}{.048} & \bci{.076}{.008} & \textbf{1.000}
     & & \bci{.145}{.012} & \bci{.628}{.052} & \bci{.110}{.009} & \textbf{.979}\\
\bottomrule
\end{tabular}}
\end{table}

\textbf{The probe beats the verbalized score on every metric, and the two fail differently.}
Table~\ref{tab:probe-vs-verbal} compares the verbalized score with the probe on the same sampled
traces. Trained
only on TriviaQA, the probe stays calibrated out of domain: its EntityQ ECE stays near the in-domain
value, while the verbalized score's ECE more than doubles. The verbalized score fails by
overconfidence, so its ECE and Brier are high together (Qwen on SimpleQA: 0.661 and 0.532). The
probe lowers ECE while also raising AUROC, so its low error does not come at the cost of
discrimination. Calibration and ranking weaken on NQ-open and on SimpleQA, the benchmark with the
lowest base rate.

\subsection{The Signal Is Robust to Task, Layer, and Readout}
\label{sec:probe-robust}

Both models read out a better-than-chance correctness signal from the early layers, while the best
calibration and ranking coincide on a broad plateau at roughly 60--65\% relative
depth: about L21--L27 for Qwen and L20--L24 for Ministral, with out-of-domain performance degrading
toward the output layers (Appendix~\ref{app:grid}). The signal is
therefore \emph{readable across layers and most stable in the middle}. The probe is equally insensitive
to the readout: probes trained on the residual
stream and on the MLP output differ by far less than the
Probe--Verbal gap (Table~\ref{tab:e2}). Across tasks, a probe trained only on TriviaQA stays
calibrated on three other factual QA benchmarks (Table~\ref{tab:probe-vs-verbal}), and on MATH-500
its ranking still transfers while its probability scale does not (Appendix~\ref{app:math}).

\subsection{A Well-Calibrated Probe Is a Poor Answer Selector}
\label{sec:not-selection}

\begin{figure}[t]
\centering
\includegraphics[width=0.9\linewidth]{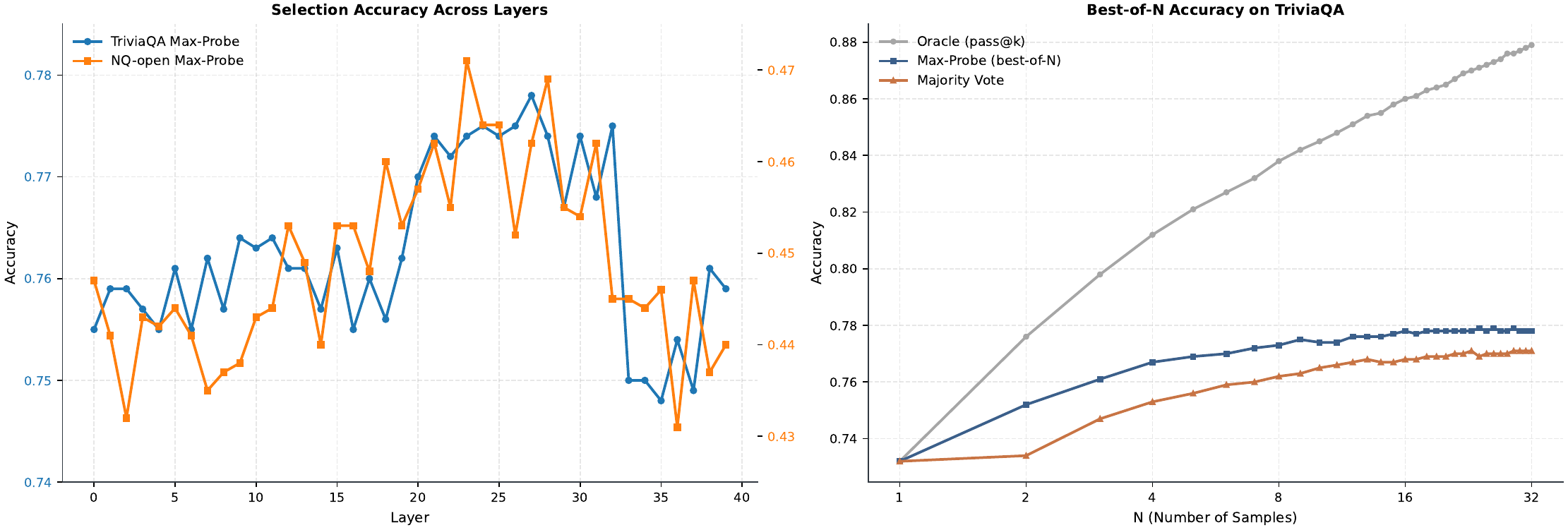}
\caption{\textbf{The probe scores traces well and selects among them poorly.} (a) Max-Probe
selection accuracy for Qwen3-14B on TriviaQA and NQ-open as a function of the layer. (b)
Best-of-$N$ curves on the TriviaQA 32-sample bench:
Max-Probe tracks lexically normalized majority voting while the oracle pass@$N$ pulls ahead.}
\label{fig:select-acc}
\end{figure}

Internal confidence is often used to select or weight answers
\citep{kang2025selfcertainty,fu2025deepconf,taubenfeld2025cisc}, so we test this linear probe as a
top-score selector. Among a question's $N$ samples, we pick the trace with the highest probe score (Max-Probe). Its
accuracy tracks lexically normalized majority voting and stays far below the oracle pass@$N$, which
counts a question as solved if any sample is correct. The pattern holds at every layer
(Figure~\ref{fig:select-acc}a), at every $N$ (Figure~\ref{fig:select-acc}b), and in a 32-sample
extension over both models and all four datasets (Appendix~\ref{app:grid}). The reason lies in what
the probe ranks well. Its high overall AUROC mostly reflects telling apart the questions
the model knows from those it does not. Selection instead requires ranking answers to the same
question, where the probe is much weaker: its within-question AUROC is 0.650 on Qwen and 0.637 on
Ministral, below its overall AUROC on every dataset by 0.07--0.24 and no better than answer frequency
(Appendix~\ref{app:within}). This agrees
with \citet{servedio2025hiddenstateshidingsomething} and \citet{taubenfeld2025cisc}. We therefore use
the probe as an offline calibration teacher, not as a reranker.

\section{Probe-Guided Self-Distillation}
\label{sec:method}

\subsection{Problem Setting}
\label{sec:teacher}

The probe cannot serve a text-only interface, so we move it from inference to offline supervision.
Let the reasoning model $\pi$ generate, for question $q$, a response $r=(\tau,a,s)$, where
$\tau$ is the chain of thought, $a$ the final answer, and $s\in\{1,\ldots,100\}$ the
verbalized confidence. The frozen probe reads the pre-answer hidden state $h$ of that trace and
yields $p=g_\phi(h)\in[0,1]$. Relabeling keeps $a$ and constructs
$r'=(\tau',a,s')$, with $s'$ aligned to $p$ and $\tau'$
preserving the reasoning while staying linguistically consistent with $s'$.
The relabeled traces supervise the same-family base model. We call the signal
providing $s'$ the \emph{teacher} and the supervised base model the \emph{student}.

\subsection{Candidate Sampling and Probe Scoring}
\label{sec:sampling}

We sample $N{=}20$ candidate responses per question from $\pi$ on TriviaQA training questions.
\citet{zhang2026ssd} report that the benefit of self-distillation grows with sampling temperature,
so the training samples use $T{=}1.5$, top-$k{=}20$, top-$p{=}0.95$, while evaluation uniformly
uses $T{=}0.6$. A temperature
ablation (Appendix~\ref{app:hparams}) shows only a
modest gain within the bootstrap half-widths. Because temperature shifts the hidden-state
distribution, we reselect the probe on a high-temperature validation set instead of reusing the
$T{=}0.6$ configuration. Qwen keeps its configuration, and Ministral
switches from L20 MLP to L24 residual. We discard samples whose answer boundary or probe readout is
missing. For a retained trace $i$ we read $h_i$ at 
the pre-answer position and compute
\begin{equation}
\label{eq:label}
p_i=\sigma(w^\top h_i+b),\qquad
s_i'=\operatorname{clip}\!\left(\operatorname{round}(100p_i),1,100\right).
\end{equation}

\subsection{Confidence Relabeling and Expression Coherence}
\label{sec:relabel}

\begin{figure}[t]
\centering\small
\renewcommand{\arraystretch}{1.25}
\begin{tabular}{@{}p{0.09\linewidth}p{0.87\linewidth}@{}}
\toprule
\multicolumn{2}{@{}l@{}}{\textbf{Question:} In 1990, which was the first UK city to be a European City of
Culture? \quad (gold: Glasgow)}\\
\midrule
\textbf{Before} & ``\ldots{} I think the first UK city to be selected was York in 1990. But I'm not
100\% sure. \ldots{} My confidence is around \textcolor{red}{\textbf{85\%}} because I've heard this
before, but I'm not entirely certain if there was another city before it.'' \quad Final answer:
York, \textcolor{red}{\textbf{85}}\\
\midrule
\textbf{After} & ``\ldots{} I think the first UK city to be selected was York in 1990. But I'm not
100\% sure. \ldots{} My confidence is around \textcolor{red}{\textbf{66\%}} because I've heard this
before, but I'm not entirely certain if there was another city before it.'' \quad Final answer:
York, \textcolor{red}{\textbf{66}}\\
\bottomrule
\end{tabular}
\caption{A real relabeled trace from Qwen3-14B (Appendix~\ref{app:audit}). The probe scores this
wrong answer at $p_i{=}0.665$, so relabeling replaces the in-chain and the final confidence (red)
with 66. The answer and all other text are unchanged, and the hedged wording stays consistent with
the new score.}
\label{fig:relabel}
\end{figure}

Relabeling changes numbers only. We replace $s_i$ after the final marker with $s_i'$ and rewrite
the in-chain confidence numbers, leaving the reasoning steps and the final answer unchanged
(Figure~\ref{fig:relabel}). This
can still produce self-contradictory samples, e.g.\ a chain of thought saying ``definitely true''
next to a confidence of 35. We therefore run a coherence filter. With $\bar p_q=N^{-1}\sum_i p_i$ the question's mean probe
score, we examine candidates in increasing order of $|p_i-\bar p_q|$. A candidate is rejected
if its chain of thought contains multiple or inconsistent confidence numbers. The remaining
candidates are rated by Gemma-4-31B-it \citep{gemmateam2026gemma4}, which judges the relabeled
triple of reasoning tone, explicit confidence statements, and final score on a 1--5 coherence scale.
The first candidate rated 5 enters the training set. The filter edits nothing and carries no answer
knowledge, so it enforces text--score consistency without injecting correctness information.

\textbf{Question-level labels (\probesdq).} \probesd{} labels each retained trace with its own probe
score (Eq.~\ref{eq:label}). Its question-level variant, \probesdq{}, labels every retained trace of
question $q$ with the question's mean probe score over all $N$ candidates,
\begin{equation}
\label{eq:qlabel}
s_i'=\operatorname{clip}\!\left(\operatorname{round}(100\,\bar p_q),1,100\right),\qquad
\bar p_q=\frac{1}{N}\sum_{j=1}^{N}p_j,
\end{equation}
and is otherwise identical: it keeps the same retained traces and rewrites their in-chain and final
confidence numbers in the same way. It targets the question-level signal that the probe reads
reliably, rather than the weak differences between samples of the same question (\S\ref{sec:not-selection}).
We recommend \probesdq{}, which is better out of domain (\S\ref{sec:main-results}). \S\ref{sec:ablation}
and \S\ref{sec:analysis} analyze \probesd{}, mostly on Qwen, where its labels already track $\bar p_q$
closely ($r{=}.987$, Appendix~\ref{app:qmean}).

To isolate the role of the teacher signal itself, we construct two controls, EA-SD and SC-SD.
EA-SD relabels a trace with the empirical accuracy of its question across the multiple samples, the
label of \citet{lin2022teachingmodelsexpressuncertainty} and \citet{jang2026confidence}.
SC-SD replaces the probe score with the answer-cluster frequency
$\operatorname{round}(100f_c)$. Both share the sampling, relabeling, filtering, and training of
\probesd{}, differing only in the teacher source.

\subsection{Training Objective and Initialization}
\label{sec:objective}

Standard SFT weights every assistant token equally, so long reasoning traces dilute supervision on
the few score tokens. On Qwen3-14B, we therefore apply a weight $\omega>1$ to the final score span
$\mathcal S_{\mathrm{score}}$, leaving the targets of reasoning and answer tokens unchanged:
\begin{equation}
\mathcal L_{\mathrm{CW}}
=-\sum_{t\in\mathrm{assistant}}w_t\log P_\theta(y_t\mid y_{<t},q),\qquad
w_t=\begin{cases}\omega,&t\in\mathcal S_{\mathrm{score}},\\ 1,&\text{otherwise.}\end{cases}
\end{equation}
Ministral repeats the answer and score inside and outside its thinking block, so standard SFT
suffices. Although the traces come from the Instruct/Reasoning
checkpoint, we initialize the student from the Base checkpoint of the same family and compare Base and Instruct/Reasoning initialization on identical distillation data.

\section{Experiments}
\label{sec:experiments}

\subsection{Setup}
\label{sec:exp-setup}

\textbf{Training and evaluation.} Students start from Qwen3-14B-Base and
Ministral-3-8B-Base. All SFT methods use the same 10{,}000 TriviaQA traces, downsampled from the
retained set (Appendix~\ref{app:downsample}), and three seeds,
with each checkpoint evaluated with 10 samples per question. The out-of-domain result is the macro average
over EntityQ, NQ-open, and SimpleQA. SFT estimates average the three seeds with 95\%
question-level bootstrap half-widths ($B{=}5000$). Verbal, Answer Probability, and post-hoc
recalibration use one generation of the original model.
Per-dataset results are in Appendix~\ref{app:full-results}.

\textbf{Compared methods.} Verbal is the original model's verbalized confidence; Answer Probability
is the mean generation probability of the answer tokens; Verbal-Iso and Verbal-TS+bias fit isotonic
and Platt scaling on TriviaQA and transfer them unchanged out of domain; Surface-Feat fits an L2
logistic readout on surface features---trace length, verbal score, hesitation count, and token
probability. EA-SD uses the question's empirical accuracy as its label, SC-SD the answer-cluster
frequency; Probe-SD-LRM is \probesd{} with Instruct/Reasoning initialization instead of Base,
identical in all else; \probesd{} uses per-trace labels and \probesdq{} question-level labels
(\S\ref{sec:relabel}). ConfTuner \citep{li2025conftuner} trains Qwen3-14B-Instruct's final score
with a tokenized Brier loss on the model's own TriviaQA traces (Appendix~\ref{app:conftuner}).

\subsection{Main Results}
\label{sec:main-results}

\begin{table}[t]
\caption{Main results. TriviaQA is in-domain. Out-of-domain is the macro average of the other three
benchmarks. Readout and rescaling methods reuse the Verbal answers and therefore have no separate
accuracy. ConfTuner: Qwen only, one run.
\probesdq{} is the question-level variant of \S\ref{sec:relabel} (three seeds per model).}
\label{tab:main}
\centering\small
\setlength{\tabcolsep}{2pt}
\renewcommand{\arraystretch}{1.3}
\resizebox{1.0\textwidth}{!}{%
\begin{tabular}{clcccccccc}
\toprule
& & \multicolumn{4}{c}{In-domain (TriviaQA)} & \multicolumn{4}{c}{Out-of-domain (macro of 3)}\\
\cmidrule(lr){3-6}\cmidrule(lr){7-10}
Model & Method & Acc $\uparrow$ & ECE $\downarrow$ & AUROC $\uparrow$ & Brier $\downarrow$
      & Acc $\uparrow$ & ECE $\downarrow$ & AUROC $\uparrow$ & Brier $\downarrow$\\
\midrule
\multirow{11}{*}{Qwen3-14B}
 & Verbal             & \ci{.729}{.024} & \ci{.178}{.020} & \ci{.825}{.017} & \ci{.184}{.017} & \ci{.258}{.013} & \ci{.542}{.013} & \ci{.715}{.017} & \ci{.468}{.011}\\
 & Verbal-Iso         & --- & \ci{.013}{.009} & \ci{.825}{.017} & \ci{.128}{.010} & --- & \ci{.274}{.012} & \ci{.714}{.017} & \ci{.236}{.007}\\
 & Verbal-TS+bias     & --- & \ci{.047}{.011} & \ci{.825}{.017} & \ci{.131}{.010} & --- & \ci{.294}{.011} & \ci{.715}{.017} & \ci{.245}{.009}\\
 & Answer Probability & --- & \ci{.251}{.024} & \ci{.636}{.029} & \ci{.253}{.023} & --- & \ci{.695}{.013} & \ci{.586}{.020} & \ci{.686}{.013}\\
 & Surface-Feat       & --- & \ci{.042}{.010} & \ci{.858}{.016} & \ci{.128}{.010} & --- & \ci{.332}{.012} & \ci{.697}{.017} & \ci{.285}{.007}\\
 & EA-SD              & \bci{.750}{.024} & \ci{.093}{.011} & \ci{.878}{.014} & \ci{.119}{.011} & \ci{.278}{.013} & \ci{.227}{.009} & \ci{.747}{.014} & \ci{.220}{.008}\\
 & SC-SD              & \ci{.748}{.023} & \ci{.063}{.013} & \ci{.876}{.014} & \ci{.116}{.010} & \bci{.281}{.013} & \ci{.236}{.010} & \ci{.753}{.013} & \ci{.222}{.007}\\
 & ConfTuner          & \ci{.737}{.024} & \ci{.048}{.014} & \ci{.913}{.015} & \ci{.100}{.009} & \ci{.258}{.014} & \ci{.125}{.010} & \bci{.815}{.017} & \bci{.136}{.007}\\
 & Probe-SD-LRM       & \ci{.729}{.024} & \ci{.031}{.011} & \ci{.926}{.011} & \ci{.096}{.009} & \ci{.258}{.012} & \ci{.135}{.009} & \ci{.774}{.017} & \ci{.154}{.006}\\
 & \probesd{}         & \ci{.749}{.024} & \ci{.024}{.009} & \ci{.928}{.011} & \ci{.092}{.009} & \ci{.279}{.013} & \ci{.113}{.008} & \ci{.783}{.015} & \ci{.147}{.006}\\
 & \probesdq{}        & \ci{.747}{.023} & \bci{.012}{.006} & \bci{.930}{.010} & \bci{.090}{.008} & \ci{.280}{.013} & \bci{.099}{.008} & \ci{.785}{.015} & \ci{.141}{.006}\\
\midrule
\multirow{10}{*}{Ministral-3-8B}
 & Verbal             & \ci{.664}{.025} & \ci{.247}{.024} & \ci{.669}{.015} & \ci{.263}{.021} & \ci{.213}{.012} & \ci{.634}{.012} & \ci{.636}{.016} & \ci{.575}{.010}\\
 & Verbal-Iso         & --- & \bci{.020}{.014} & \ci{.669}{.015} & \ci{.188}{.009} & --- & \ci{.316}{.012} & \ci{.636}{.016} & \ci{.267}{.005}\\
 & Verbal-TS+bias     & --- & \ci{.067}{.016} & \ci{.669}{.015} & \ci{.193}{.008} & --- & \ci{.341}{.006} & \ci{.636}{.009} & \ci{.280}{.003}\\
 & Answer Probability & --- & \ci{.302}{.025} & \ci{.569}{.027} & \ci{.306}{.025} & --- & \ci{.753}{.013} & \ci{.520}{.018} & \ci{.746}{.012}\\
 & Surface-Feat       & --- & \ci{.042}{.010} & \ci{.770}{.016} & \ci{.174}{.009} & --- & \ci{.329}{.011} & \ci{.673}{.017} & \ci{.276}{.006}\\
 & EA-SD              & \ci{.696}{.024} & \ci{.123}{.015} & \ci{.826}{.014} & \ci{.160}{.012} & \ci{.232}{.012} & \ci{.251}{.009} & \ci{.734}{.015} & \ci{.234}{.007}\\
 & SC-SD              & \bci{.698}{.025} & \ci{.052}{.007} & \ci{.809}{.015} & \ci{.153}{.010} & \ci{.232}{.012} & \ci{.169}{.008} & \ci{.722}{.014} & \ci{.192}{.005}\\
 & Probe-SD-LRM       & \ci{.644}{.026} & \ci{.064}{.011} & \bci{.904}{.012} & \ci{.128}{.008} & \ci{.206}{.012} & \ci{.142}{.009} & \ci{.778}{.016} & \ci{.147}{.006}\\
 & \probesd{}         & \ci{.693}{.025} & \ci{.029}{.009} & \ci{.887}{.013} & \ci{.123}{.009} & \ci{.232}{.012} & \ci{.140}{.008} & \ci{.783}{.015} & \ci{.154}{.005}\\
 & \probesdq{}        & \ci{.693}{.026} & \ci{.031}{.013} & \ci{.894}{.014} & \bci{.119}{.008} & \bci{.233}{.012} & \bci{.089}{.008} & \bci{.789}{.016} & \bci{.131}{.005}\\
\bottomrule
\end{tabular}}
\end{table}

\textbf{\probesd{} improves calibration and outperforms the evaluated post-hoc recalibration methods
out of domain.} Out-of-domain ECE falls from 0.542 to 0.113 on Qwen and from 0.634 to 0.140 on
Ministral, and AUROC rises on both models (Table~\ref{tab:main}). Isotonic recalibration (Verbal-Iso) is competitive in domain, but
its out-of-domain ECE is more than twice that of \probesd{}, and as a monotone rescaling it cannot
change AUROC. \probesd{} improves both by writing the internal teacher's signal into the output.

\textbf{The probe label's ranking is a major source of its gain.} EA-SD and SC-SD
differ from \probesd{} only in the label, yet their out-of-domain ECE is far higher, and \probesdq{} lowers it further (Appendix~\ref{app:qmean}). The probe label is not the more accurate one, however: on held-out samples of
the training questions, empirical accuracy estimates a question's accuracy more closely
(Appendix~\ref{app:mech}). The two labels differ in how they rank the training traces and in their
values, which are less extreme for the probe. We separate the two on Qwen with two hybrid students
(EA-QM and Probe-QM in Table~\ref{tab:2x2}).
One keeps the empirical-accuracy ranking and takes the probe labels' values by quantile, and the
other keeps the probe ranking with the empirical-accuracy values. The probe's
values alone halve in-domain ECE (0.096 to 0.047) but leave out-of-domain ECE unchanged (0.228 to
0.233), as expected of a monotone rescaling. The probe's ranking alone lowers
out-of-domain ECE to 0.145, about 70\% of the way to \probesd{}'s 0.113. This gain sits on the
questions the model never answers correctly: 92\% of the out-of-domain Brier gap between EA-SD and
the probe-ranked hybrid comes from questions with pooled accuracy below 0.05, not from mid-accuracy questions, where
sampling noise in empirical accuracy is largest (Appendix~\ref{app:bins}). A likely reason is
learnability. The probe estimates accuracy from the model's own hidden state, so the student can
learn its ranking from the input. Empirical accuracy also carries accuracy variation that this state
does not encode, which the student can only memorize per training question and cannot carry to new
ones. Consistently, students fit probe-ranked training labels more closely (Appendix~\ref{app:mech}).

\textbf{ConfTuner ranks better, \probesd{} calibrates better and stays consistent.} \probesd{} has
lower ECE in and out of domain, while ConfTuner has the highest out-of-domain AUROC and lowest
out-of-domain Brier. ConfTuner supervises only the
final score. About 22\% of responses also state a confidence in the chain of thought. Among these,
ConfTuner's final score differs from the in-chain one by more than 20 points in 28.0\%, versus
0.5\% for the original model and 0.1\% for \probesd{} (Appendix~\ref{app:conftuner}). \probesd{} keeps the two consistent because
relabeling rewrites both scores together and the coherence filter rejects traces whose wording
disagrees.

\textbf{Base initialization adds accuracy, and SimpleQA remains hard.} Against the same method
initialized from the Instruct/Reasoning checkpoint (Probe-SD-LRM),
\probesd{} raises out-of-domain accuracy from 0.258 to 0.279 on
Qwen and from 0.206 to 0.232 on Ministral, mostly through retrieval (\S\ref{sec:priming-main}).
SimpleQA is the most extreme base-rate shift (about 7\% accuracy versus about 70\% in training).
\probesd{} cuts its ECE about fivefold, but absolute calibration under this shift remains imperfect
(Appendix~\ref{app:full-results}).

\subsection{Component Ablation}
\label{sec:ablation}

\begin{table}[t]
\caption{Component ablation on Qwen3-14B. ``relabel'' is probe relabeling of the verbalized confidence and CW the confidence-weighted loss.}
\label{tab:ablation}
\centering\small
\setlength{\tabcolsep}{4pt}
\renewcommand{\arraystretch}{1.1}
\resizebox{0.87\textwidth}{!}{%
\begin{tabular}{lcccccccccc}
\toprule
& \multicolumn{2}{c}{Components} & \multicolumn{4}{c}{In-domain} & \multicolumn{4}{c}{Out-of-domain}\\
\cmidrule(lr){2-3}\cmidrule(lr){4-7}\cmidrule(lr){8-11}
Variant & relabel & CW & Acc & ECE $\downarrow$ & AUROC $\uparrow$ & Brier $\downarrow$
& Acc & ECE $\downarrow$ & AUROC $\uparrow$ & Brier $\downarrow$\\
\midrule
\probesd{} (full) & yes & yes & \textbf{.749} & \textbf{.024} & \textbf{.928} & \textbf{.092}
& .279 & \textbf{.113} & \textbf{.783} & \textbf{.147}\\
w/o CW            & yes & no  & .748 & .033 & .919 & .097
& .280 & .129 & .768 & .157\\
w/o relabel       & no  & no  & .747 & .159 & .814 & .173
& \textbf{.281} & .522 & .723 & .446\\
\bottomrule
\end{tabular}}
\end{table}

Relabeling is decisive (Table~\ref{tab:ablation}): without it, out-of-domain ECE stays at 0.522, while
accuracy changes by under one percentage point. It changes the confidence scale, not the answers.
The confidence-weighted loss (CW) adds a secondary gain: removing it raises out-of-domain ECE from
0.113 to 0.129 and lowers out-of-domain AUROC from 0.783 to 0.768. The small accuracy gain also
appears without relabeling, so it comes from Base-initialized SFT.

\subsection{Multiple-Choice QA with Shuffled Options}
\label{sec:mcqa-main}

Multiple-choice questions add two things that factual QA lacks. A model can be right by guessing,
so empirical accuracy counts lucky guesses, and the option order can change while the question stays
the same. We rerun the full pipeline on MMLU \citep{hendrycks2021mmlu} with Qwen3-14B: 40 knowledge
subjects, 10{,}235 training and 1{,}502 test questions. Each training sample gets its own random
option order, and the label probe is cross-fitted over two folds of questions. Each test question is
asked in four option orders, with the correct option once in each position. MMLU-Pro
\citep{wang2024mmlupro} is out of domain: 1{,}000 knowledge questions with ten options. The
across-order standard deviation $\sigma_{\text{order}}$ measures how much the stated confidence moves
when only the order changes, net of sampling noise (Appendix~\ref{app:mcqa}).

\begin{table}[t]
\caption{Multiple-choice QA on Qwen3-14B. Students average three seeds (seed standard deviation at
most .004). $\sigma_{\text{order}}$ is the standard deviation of stated confidence across option
orders, net of sampling noise. Full results, including Ministral-3-8B, are in Appendix~\ref{app:mcqa}.}
\label{tab:mcqa-main}
\centering\small
\setlength{\tabcolsep}{5pt}
\begin{tabular}{lcccccc}
\toprule
& \multicolumn{3}{c}{MMLU (in-domain)} & \multicolumn{3}{c}{MMLU-Pro (out-of-domain)}\\
\cmidrule(lr){2-4}\cmidrule(lr){5-7}
Method & ECE $\downarrow$ & AUROC $\uparrow$ & $\sigma_{\text{order}}$ $\downarrow$
       & ECE $\downarrow$ & AUROC $\uparrow$ & $\sigma_{\text{order}}$ $\downarrow$\\
\midrule
Verbal (original)  & .085 & .729 & .018 & .127 & .697 & .016\\
Probe (original)   & .030 & .857 & .051 & .089 & .828 & .066\\
EA-SD              & .109 & .783 & .072 & .130 & .773 & .085\\
\probesd{}         & .023 & .854 & .036 & .058 & .843 & .043\\
\probesdq{}        & .023 & .856 & .035 & .057 & .848 & .041\\
\bottomrule
\end{tabular}
\end{table}

\textbf{The result carries over.} All three students reach the same accuracy (0.84 on MMLU), so
they differ only in the confidence they state (Table~\ref{tab:mcqa-main}). \probesd{}'s ECE is about
five times lower than EA-SD's in domain and half of it on MMLU-Pro, with higher AUROC. Its
confidence also moves half as much across option orders. The original model's verbal score barely
moves, but only because it is near 0.9 for almost every answer. As on TriviaQA, empirical accuracy
estimates each training question's accuracy more closely than the probe (RMSE 0.066 against 0.255)
yet trains the worse student, so the noise that guessing adds to empirical accuracy does not explain
the gap. Ministral-3-8B shows the same pattern with one seed (ECE 0.027 against 0.113).

\textbf{Confident misconceptions remain.} When the model gives the same wrong answer in every
order, every confidence source stays high (0.65--0.89, probe 0.67). Neither the probe nor the
student detects confident misconceptions.

\section{Analysis}
\label{sec:analysis}

We ask what the student inherits from its teacher, where the Base-initialization gain comes from, and
how its confidence behaves under test-time aggregation.

\subsection{The Student Inherits Its Teacher, Including Where Self-Consistency Fails}
\label{sec:fidelity}
\label{sec:label-source}

\begin{figure}[t]
\centering
\includegraphics[width=0.9\linewidth]{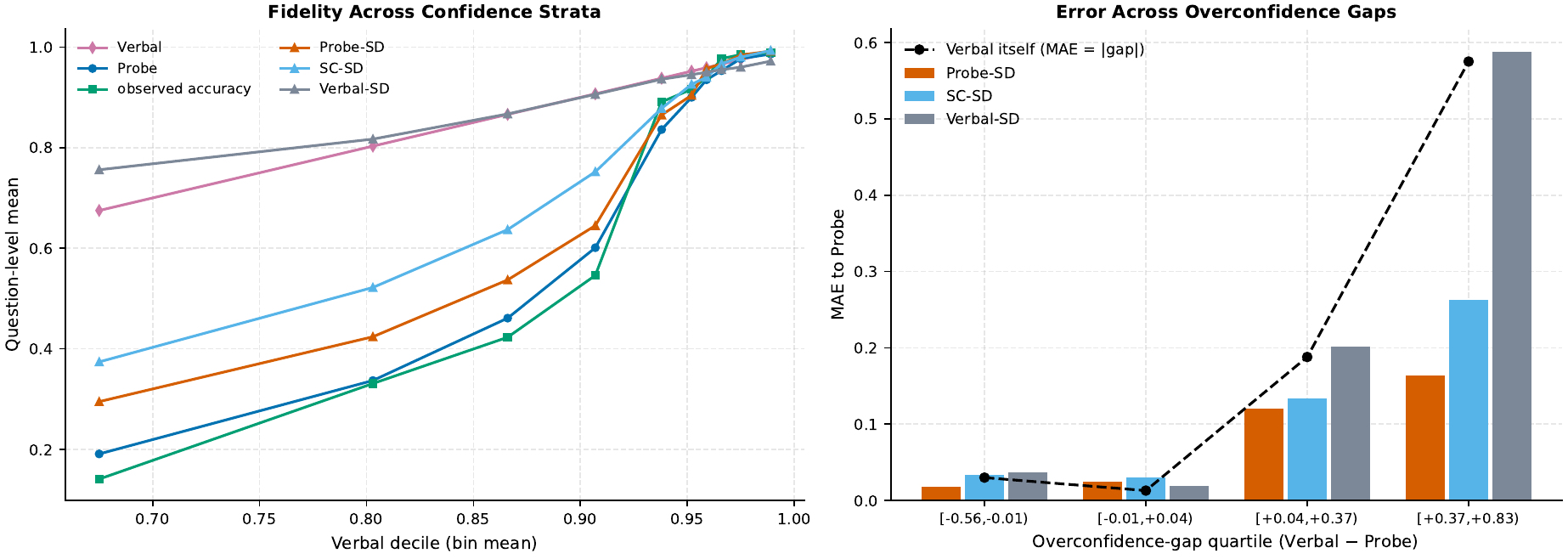}
\caption{\textbf{Each student ends up closest to the teacher it was trained on.} (a) Probe, source
verbal score, observed accuracy, and the three students' verbalized confidence on TriviaQA,
aggregated by source-verbal decile. (b) Mean absolute error of each student relative to Probe on
TriviaQA, by quartile of the overconfidence gap between source verbal and Probe.}
\label{fig:fidelity}
\end{figure}

Each student tracks the teacher it was trained on (Figure~\ref{fig:fidelity}a). On Qwen3-14B,
\probesd{}, SC-SD and Verbal-SD differ only in the confidence target: the probe score, the
answer-cluster frequency, or the source model's own verbalized score. The difference concentrates where
the original model is overconfident. In the top quartile of the source--probe gap, \probesd{} stays
within 0.16 of the probe, whereas Verbal-SD drifts by 0.59 (Figure~\ref{fig:fidelity}b). \probesd{}
therefore does not lower scores uniformly. It corrects the traces where the verbal channel and the
internal teacher disagree. Correlations and MAEs for all datasets are in Appendix~\ref{app:mechanism}.

The same view explains where SC-SD falls behind. The gap between \probesd{} and SC-SD comes from the
minority of questions on which the two labels disagree. Where the mean probe score far exceeds the
answer frequency (13.4\% of questions), answer frequency underestimates accuracy by 0.38, while the
probe stays close to it. \probesd{} follows its teacher there, and SC-SD inherits its label's
error (Appendix~\ref{app:mechanism}).

\subsection{Base Initialization Improves Factual Retrieval}
\label{sec:priming-main}

Base initialization helps mainly by making the model recall the answer. Reasoning helps factual QA
largely by recalling facts into the context \citep{gekhman2026thinkingrecallreasoningunlocks}, so we
compare \probesd{} with the original model along this path on matched TriviaQA questions. A trace
counts as a recall hit when the gold answer or an alias appears in its chain of thought. We split the
accuracy gain into a higher hit rate (retrieval), higher accuracy after a hit (use), and the miss
branch, and sort questions by difficulty using their hit counts. Retrieval accounts for most of the
gain on both models, and on Qwen accuracy after a hit is unchanged. The gain concentrates in
medium-difficulty questions and most hard questions stay hard, so Base initialization raises access
to latent factual knowledge at the margin without moving its boundary (Appendix~\ref{app:priming}).

\subsection{SFT Alone Produces Test-Time Confidence Behavior}
\label{sec:tts-main}

We adapt two diagnostics from RLCR \citep{damani2025binaryrewardstraininglms}, but claim no
equivalence to an RL-trained model (Appendix~\ref{app:rl}). On Qwen3-14B TriviaQA, internalized confidence is more useful for weighting than for
selecting, mirroring the probe in \S\ref{sec:not-selection}: at $N{=}16$, confidence-weighted voting
beats max-confidence selection by 1.3 points (95\% CI $[0.4, 2.3]$) and matches or slightly exceeds
majority voting ($+0.5$, CI $[-0.1, 1.0]$). For mutually exclusive answers, the original model's
confidences sum to more than 2 on 33.3\% of questions, versus 13.3\% for \probesd{}, which rules out
saturated overconfidence (Appendix~\ref{sec:tts}).

\section{Conclusion and Limitations}
\label{sec:conclusion}

Across two model families and four factual QA benchmarks, a linear probe at the pre-answer position
is far better calibrated than the verbalized score. Because a probe needs hidden-state access, we
apply it offline as a teacher: it scores the model's own traces, the stated confidence is overwritten
with the probe score, and the base checkpoint of the same family is finetuned on the result, so the
model itself states calibrated confidence at deployment. Its question-level variant \probesdq{} is
better still out of domain. The same holds on multiple-choice QA with shuffled options, where the
student's confidence is also less sensitive to the option order. On Qwen, the probe label beats empirical accuracy mainly through its
ranking, which reflects what the model's hidden state encodes and which students fit more closely. As an answer selector, by
contrast, the probe nearly ties majority voting.

The claims are limited to factual and multiple-choice knowledge QA with short, checkable answers on
two open-weight reasoning families up to 14B parameters. Neither the probe nor the student detects
answers that the model gives consistently and wrongly (\S\ref{sec:mcqa-main}). On MATH-500, the probe's ranking transfers but its probability scale
does not (Appendix~\ref{app:math}).
Building the
training data requires white-box access and per-model probe configuration (Appendix~\ref{app:lim}).

\label{endofsec7}

\subsection*{Ethics Statement}

This work studies the calibration of confidence reported by language models on factual question
answering, using publicly available benchmarks and open-weight models. No human subjects, private
data, or personally identifying information are involved. Better-calibrated confidence is intended
to support honest communication of uncertainty and safer downstream thresholding. We note one
dual-use consideration: a model that states well-calibrated confidence may be trusted more
readily, so residual miscalibration on out-of-distribution or low base-rate inputs could be more
consequential than the same error from a visibly overconfident model.

\subsection*{Reproducibility Statement}

All datasets and base models used are publicly available. The experimental protocol and supporting
analyses are documented in Appendices~\ref{app:setup}--\ref{app:lim}. Code will be released upon
publication.

\subsection*{AI Use Statement}

We used generative AI tools as research artifacts inside the experimental pipeline: answer
extraction and correctness grading (Qwen3-32B-Instruct), coherence filtering of relabeled traces
(Gemma-4-31B-it), and chain-of-thought fact extraction and labeling for the factual priming audit
(glm-5-2). All of these are documented in the appendices, and the grader was validated against a
human audit. We additionally used generative AI assistance for writing
polish and for code implementation, all of which was reviewed by the authors. We did not use
generative AI to generate research ideas, experimental results, or claims. We take
responsibility for the final content of this work.
\label{endofmaintext}

\bibliography{iclr2027_conference}
\bibliographystyle{iclr2027_conference}

\appendix

\section{Experimental Setup}
\label{app:setup}
\label{sec:probe-setup}

This appendix expands the setup summarized in \S\ref{sec:signal}.

\paragraph{Data and models.} We use four factual QA benchmarks with short, automatically checkable
answers: TriviaQA \citep{joshi2017triviaqa}, EntityQuestions (EntityQ)
\citep{sciavolino2021entityquestions}, NQ-open \citep{kwiatkowski2019natural}, and
SimpleQA-Verified \citep{wei2024simpleqa,haas2025simpleqa}, with 1{,}000 test questions each and
10 traces per question, on Qwen3-14B \citep{qwen3} (40 layers) and Ministral-3-8B-Reasoning-2512
\citep{liu2026ministral3} (34 layers), sampled at $T{=}0.6$, top-$k{=}20$, top-$p{=}0.95$. Probes
are trained on 10{,}000 TriviaQA training questions, so TriviaQA is in-domain and the other three
are out-of-domain throughout. All test sets are held out from configuration selection.
Qwen3-32B-Instruct extracts and grades answers. Its reliability is
audited against human labels in Appendix~\ref{app:judge}.

\paragraph{Readout and probe.} The prompt asks for a complete chain of thought followed by
\texttt{\#\#FINAL ANSWER\#\#\textbackslash n\{answer\}, \{score\}}, with the score an integer from
1 to 100 normalized to $[0,1]$ (Appendix~\ref{app:prompts}). Among seven candidate extraction
positions, the \emph{last token before the answer}---after the chain of thought has ended and
before the answer begins---was stable across settings and always close to optimal, so all
experiments use it. Per layer we fit an $\ell_2$-regularized logistic regression
$p_{\mathrm{probe}}=\sigma(w^\top h+b)$ on the residual stream and on the MLP output separately,
with cross-entropy against trace correctness. TriviaQA validation data alone select layer~24 residual with
$C{=}0.005$ for Qwen3-14B and layer~20 MLP with $C{=}1.0$ for Ministral-3-8B, at 60.0\% and 58.8\%
relative depth. The candidate positions, the full grid, and the selection criterion are in
Appendix~\ref{app:grid}.

\paragraph{Metrics.}\label{app:metrics}
Let $y_i\in\{0,1\}$ indicate whether trace $i$ is correct and let $p_i\in[0,1]$ be the confidence
assigned to that trace. Unless stated otherwise, confidence metrics are computed over the
\emph{scorable subset}: traces for which the source produces a valid confidence score.
Expected calibration error (ECE) \citep{naeini2015obtaining} partitions scores into 10 equal-width
bins and computes
\[
\mathrm{ECE}=\sum_{b=1}^{10}\frac{|I_b|}{n}
\left|\frac{1}{|I_b|}\sum_{i\in I_b}p_i-
\frac{1}{|I_b|}\sum_{i\in I_b}y_i\right|,
\]
where $I_b$ is the set of scored traces in bin $b$. ECE is zero for perfect empirical calibration
and is lower when stated confidence more closely matches observed accuracy; because it depends on
binning, we use the same bins for every method. The Brier score is the mean squared probabilistic
error, $n^{-1}\sum_i(p_i-y_i)^2$. It is a strictly proper score that reflects both calibration and
discrimination, with lower values better. AUROC is the probability that a randomly chosen correct
trace receives a higher score than a randomly chosen incorrect trace (with ties handled in the
standard rank-based manner). It measures ranking rather than probability-scale calibration: 0.5 is
chance and 1 is perfect discrimination. Coverage is the number of traces with a valid confidence
score divided by the total number of generated traces; higher is better. Thus a source may have low
ECE but weak AUROC, or strong AUROC but a miscalibrated scale, so we report both together with Brier
and Coverage. Appendix~\ref{app:missing} tests whether differences in the scorable subset affect the
comparison.
\clearpage
\section{Prompts and Sampling Settings}
\label{app:prompts}

\paragraph{Sampling.} Training data are generated at $T{=}1.5$, top-$k{=}20$, top-$p{=}0.95$
with 20 samples per question, identically for both models. Evaluation uses $T{=}0.6$,
top-$k{=}20$, top-$p{=}0.95$ with 10 samples per question. The verbalized score is parsed as
\texttt{,\textbackslash d\{1,3\}} after the last \texttt{\#\#FINAL ANSWER\#\#} marker and
normalized to $[0,1]$. For Ministral-3-8B we always take the last marker with \texttt{rfind},
because the model rehearses the marker inside its thinking block. All prompts ask the model to
reason step by step about both the answer and its confidence and then emit
\texttt{\#\#FINAL ANSWER\#\#\textbackslash n\{answer\}, \{score\}} with the score an integer
from 1 to 100.

\paragraph{Qwen3-14B prompt.} Used unchanged for both training-data generation and evaluation.

{\scriptsize
\begin{Verbatim}[breaklines=true]
[system] You are a precise assistant. Think step by step about how confident you
         are (score 1-100), then give your answer.
[user]   Format: ##FINAL ANSWER##\n{answer}, {score}

         Question: {question}
\end{Verbatim}
}

\paragraph{Ministral-3-8B-Reasoning-2512 prompt.} We keep the model's own default system prompt
(the structured text + \texttt{thinking(closed)} + text template that triggers
\texttt{mistral-common} reasoning mode) and append the format instruction to the user turn.

{\scriptsize
\begin{Verbatim}[breaklines=true]
[system] # HOW YOU SHOULD THINK AND ANSWER

         First draft your thinking process (inner monologue) until you
         arrive at a response. Format your response using Markdown, and use
         LaTeX for any mathematical equations. Write both your thoughts and
         the response in the same language as the input.

         Your thinking process must follow the template below:
         [THINK]Your thoughts and/or draft, like working through an
         exercise on a scratch paper. Be as casual and as long as you want
         until you are confident to generate the response to the
         user.[/THINK]Here, provide a self-contained response.
[user]   Question: {question}

         Answer the question. Then append your confidence (1-100) on the
         same line.

         Format:
         ##FINAL ANSWER##\n{answer}, {score}
\end{Verbatim}
}

\subsection{Consistency Search and Relabeling}
\label{app:relabel-prompts}

Candidates are visited in the order of \S\ref{sec:relabel} and each one is passed through a
three-step filter chain, stopping at the first candidate that survives. (1)~\textbf{Score
extraction check} (judge with thinking off): scan the chain of thought for every explicit
confidence number. The candidate passes if no score is found, or if exactly one is found and it
agrees with the score after \texttt{\#\#FINAL ANSWER\#\#}. It is rejected if several scores
appear or if any contradicts the final one. (2)~\textbf{Mechanical relabeling} (pure string
operations, no prompt): the verbalized score in
\texttt{\#\#FINAL ANSWER\#\#\textbackslash n\{answer\}, \{old\_score\}} is replaced by
$s_{\mathrm{new}}=\max(1,\min(100,\mathrm{round}(p_{\mathrm{probe}}\times100)))$, and every
1--100 number in the chain of thought that is adjacent to a confidence word (``confident'',
``sure'', ``certain'', ``likely'', ``guess'', \ldots) is replaced by the same value. Dates,
counts and statistics are skipped. Ministral typically restates answer and score inside the
\texttt{[THINK]} block before the final marker, so relabeling rewrites both occurrences and the
rehearsal is kept rather than trimmed. (3)~\textbf{Coherence check} (judge with thinking on):
the relabeled response is rated 1--5 for end-to-end consistency between chain of thought,
confidence wording and final score. A rating of 5 selects the candidate and stops the search.
Otherwise the next candidate is tried. Relabeling must precede the coherence check, because what
step~3 evaluates is the self-consistency of the \emph{relabeled} response. Both judge steps use
Gemma-4-31B-it \citep{gemmateam2026gemma4}.

\paragraph{Step 1: score extraction (thinking off).}

{\scriptsize
\begin{Verbatim}[breaklines=true]
[system] You are a confidence-score extractor. Your task is to scan a model's
  thinking chain and identify every explicit confidence score (1-100) the model
  states about its own answer.

  A confidence score is a number 1-100 that answers "how confident am I?" - it
  represents the model's self-assessed likelihood of being correct. Look for
  patterns like:
  - "score X", "score of X", "score: X", "my score is X"
  - "I'm X% confident/sure/certain", "I am X% sure"
  - "confidence: X", "confidence is X", "confidence level X"
  - "about X% confident", "around X%", "maybe X", "probably X%"
  - "give it a X", "rate it X", "put it at X"

  Do NOT count: years (19xx, 20xx), dates, quantities of things, statistics, or
  any number that is NOT describing the model's own confidence level. If the
  thinking says "score is 1-100" (describing the scale), that is NOT a
  confidence score.

[user]
  ## Thinking chain
  {thinking}

  ## Task
  Identify EVERY distinct confidence score the model explicitly states in this
  thinking chain. Return the count and the complete list.

  Return ONLY a JSON object (no markdown, no extra text):
  {"count": <integer>, "scores": [<integer, ...>]}
\end{Verbatim}
}

\paragraph{Step 3: coherence check (thinking on).}

{\scriptsize
\begin{Verbatim}[breaklines=true]
[system] You are a coherence evaluator. Your task is to judge whether a model's
  entire response - from the beginning of its thinking process to its final
  answer - forms a harmonious, consistent whole where the confidence it
  expresses and the scores it gives all align with each other.

  The model's response contains a thinking chain inside <think>...</think>,
  followed by a final answer in the format "##FINAL ANSWER##\n{answer},
  {score}". The score is a number from 1 to 100 on a confidence scale:
  - 1 means "I am completely guessing, I have no idea."
  - 100 means "I am absolutely certain, beyond any doubt."
  A score of 5 or 10 is extremely low - it means the model is admitting it
  barely knows.

  Your evaluation scope is the ENTIRE response, not just the final score. Judge
  the thinking process itself - how the model arrives at its answer matters, not
  just where it ends up. A model that reasons uncertainly but then gives a high
  score is still contradictory, even if its final conclusion is correct.

  You must check:

  1. **End-to-end coherence**: From the first sentence of thinking to the last,
  the model's expressed certainty should be internally consistent. If it starts
  confident, wavers, then becomes confident again - that is not coherent. If it
  sounds sure in some sentences and clueless in others - that is not coherent.
  The whole process must flow naturally toward the final score.

  2. **Score-tone alignment everywhere**: Every time the model mentions a
  confidence level - whether a numeric score inside the thinking ("I'm about X%
  sure"), a verbal expression ("pretty confident", "not sure", "definitely"), or
  the final score - check that the number and the surrounding tone match. This
  is a hard requirement:
     - Confident words ("definitely", "pretty confident", "I'm sure",
       "certain", "no doubt") + any score below 40 -> ALWAYS at most score 2.
     - Uncertain words ("not sure", "guessing", "no idea", "don't know", "not
       confident", "not certain") + any score above 80 -> ALWAYS at most
       score 2.

  3. **Scale awareness**: You must recognize where numbers fall on the 1-100
  scale. A score of 3, 5, 7, 11, 15 - these are all very low. If the thinking
  calls such a score "high" (e.g., "the score would be high, like 5"), that is a
  direct contradiction. A low score (below 40) should only appear alongside
  hesitant, uncertain reasoning. A high score (above 80) should only appear
  alongside confident, decisive reasoning. Mid-range scores (40-79) should
  appear alongside moderately confident reasoning with some hedging.

  4. **No contradictions allowed for score 5**: Score 5 is reserved for perfect
  alignment. If ANY part of the thinking contradicts ANY part of the confidence
  expression - internal or final - it cannot be a 5.

  Rate the coherence on a 1-5 scale:
  1 = Severe conflict: direct contradiction between the thinking's tone and any
      score mentioned, or violation of the hard requirement in point 2.
  2 = Moderate conflict: significant mismatch, mixed signals that undermine
      alignment, or violation of the hard requirement in point 2.
  3 = Neutral: ambiguous, inconsistently coherent, thinking and scores feel
      loosely related.
  4 = Good alignment: the thinking generally supports the scores throughout,
      with only minor imperfections. No hard requirement violations.
  5 = Perfect alignment: every part of the thinking is tonally consistent with
      every score mentioned, from start to finish. The entire response is a
      harmonious, unified whole.

  Return ONLY a JSON object (no markdown, no extra text):
  {"score": <1-5 integer>, "reason": "<one concise sentence>"}

[user]
  ## Question
  {question}

  ## Model Response (full)
  {raw_response}

  ## Task
  Evaluate the coherence of this model response from start to finish. The
  confidence scale is 1-100, where 5 means near-zero confidence. Check every
  confidence expression - internal scores, verbal tones, and the final score -
  they must all align.

  Return ONLY a JSON object:
  {"score": <1-5 integer>, "reason": "<one concise sentence>"}
\end{Verbatim}
}

\subsection{Audit of Mechanical Relabeling}
\label{app:audit}

A rating of 5 is a strict quality gate: a candidate enters the training set only if the
relabeled chain of thought, the confidence wording and the final score are consistent end to
end. Mechanical relabeling never touches the answer or non-confidence text. It only synchronizes
explicit self-confidence numbers. The audit pool contains 14{,}951 SFT training traces, of which
493 candidates have a verbal--probe gap of 18--22 points. Table~\ref{tab:b1} lists five of them
(Qwen3-14B, $T{=}0.6$, chain-of-thought length filtered to 300--1500 tokens). Grading is by
Qwen3-32B-Instruct and the replacement score is
$\mathrm{round}(p_{\mathrm{probe}}\times100)$.

\begin{table}[h]
\caption{Targeted audit of mechanical relabeling. Candidates were selected by gap, so the table
verifies the replacement mechanism and the direction of the label, not the average size of the
correction.}
\label{tab:b1}
\centering\small
\setlength{\tabcolsep}{4pt}
\begin{tabular}{llcllcc}
\toprule
qid & idx & Gold & Model answer & Graded & Verbal $\to$ new & $p_{\mathrm{probe}}$\\
\midrule
\texttt{qb\_8463}   & 10 & Times Square & Times Square             & correct & 95 $\to$ 73 & .727\\
\texttt{qw\_10780}  & 17 & USA          & United States of America & correct & 95 $\to$ 74 & .742\\
\texttt{jp\_2478}   & 17 & Elizabeth    & Elizabeth                & correct & 99 $\to$ 81 & .806\\
\texttt{sfq\_19580} & 11 & Chris Huhne  & Chris Huhne              & correct & 95 $\to$ 75 & .749\\
\texttt{sfq\_4960}  & 0  & Glasgow      & York                     & \textbf{incorrect} & 85 $\to$ 66 & .665\\
\bottomrule
\end{tabular}
\end{table}

The audit deliberately includes both correct and incorrect samples, showing that relabeling does
not merely lower high scores on questions the model got right: row 5 lowers 85 to 66 while the
answer remains the incorrect \emph{York}.

\paragraph{Example 1 (correct answer, 95 $\to$ 73).} \emph{Which New York landmark is known as
`The Crossroads of the World'?} (gold: Times Square). The chain of thought contains
``\ldots{}maybe it's referring to a different location? \ldots{} Like Grand Central Terminal? No,
I don't think so. \ldots{} I think my confidence here is about 95\%.'' The final answer
\texttt{Times Square, 95} is relabeled to \texttt{Times Square, 73} and the in-chain 95\% is
replaced by 73\%. The answer and all other reasoning text are unchanged.

\paragraph{Example 2 (incorrect answer, 85 $\to$ 66).} \emph{In 1990, which was the first UK
city to be a European City of Culture?} (gold: Glasgow, the model answers York). The chain of
thought contains ``I think the first UK city to be selected was York in 1990. But I'm not 100\%
sure. \ldots{} My confidence is around 85\% because I've heard this before, but I'm not entirely
certain if there was another city before it.'' The final answer \texttt{York, 85} becomes
\texttt{York, 66} and the in-chain 85\% becomes 66\%. The answer is still the incorrect
\emph{York}: the probe acts here as a calibration signal, not an answer selector.
\clearpage
\section{Probe Grid Search and Configuration Selection}
\label{app:grid}

\paragraph{Features and training.} The probe input is the hidden-state readout
$\mathbf{x}\in\mathbb{R}^d$ at the pre-answer position---the token preceding the first answer
token, i.e.\ immediately after the newline following the last
\texttt{\#\#FINAL ANSWER\#\#}---with $d{=}5120$ for Qwen3-14B and $d{=}4096$ for
Ministral-3-8B. That position was chosen from seven candidates: the first token after the chain of
thought ends, the last token before the answer, the first answer token, the last answer token, the
mean over all answer tokens, the token preceding the score, and the score token itself. It was the
only one stable across models, datasets and temperatures while always scoring close to the best
candidate, so \S\ref{sec:probe-setup} fixes it for every experiment. We fit $\ell_2$-regularized
logistic regression
\[
\min_{\mathbf{w},b}\ \frac{1}{N}\sum_{i=1}^{N}\ell\!\left(y_i,\sigma(\mathbf{w}^\top\mathbf{x}_i+b)\right)
+\frac{1}{C}\|\mathbf{w}\|_2^2,
\]
where $y_i\in\{0,1\}$ is the correctness label of sample $i$ as decided by the Qwen3-32B-Instruct
grader. Every configuration is trained with lbfgs and \texttt{max\_iter=2000} under three random
seeds $\{1,2,3\}$, and the reported probe score is the ensemble mean
$p_{\mathrm{probe}}=\frac{1}{3}\sum_{s=1}^{3}\sigma(\mathbf{w}_s^\top\mathbf{x}+b_s)$.

\paragraph{Search space and selection.} We take the Cartesian product of all layers, both
readouts and eleven regularization strengths (Table~\ref{tab:e1}). Probes are trained on
model-generated samples from the TriviaQA training split (Qwen3-14B: 9{,}987 samples, positive
rate 0.735, Ministral-3-8B: 9{,}743 samples, positive rate 0.665), disjoint from the validation
and test splits. Selection uses TriviaQA validation data only, with the composite rank described in
Appendix~\ref{app:hparams}. The selected configurations are
reported in Table~\ref{tab:c2} and only then evaluated on the four held-out test sets. Configurations with $C\ge5$
raise lbfgs convergence warnings and are excluded from consideration.

\paragraph{Sensitivity to the selection rule.} The submitted version also ranked EntityQ and NQ-open
validation data. Dropping them selects the same probe for Qwen at both temperatures, so every Qwen
result is unchanged. For Ministral the evaluation probe moves from L21 MLP ($C{=}2.0$) to L20 MLP
($C{=}1.0$) and the label probe from L22 to L24 residual. All Ministral results in this version use
the new probes. The exact choice matters little: Table~\ref{tab:e-worst} reports, among the 40
configurations per model within 0.005 Brier of the best on TriviaQA validation, the worst test ECE
on each out-of-domain benchmark.

\begin{table}[h]
\caption{Worst out-of-domain test ECE among the 40 probe configurations per model that TriviaQA
validation cannot distinguish from the best (within 0.005 Brier). The verbalized score is in parentheses.}
\label{tab:e-worst}
\centering\small
\begin{tabular}{lccc}
\toprule
Model & EntityQ & NQ-open & SimpleQA\\
\midrule
Qwen3-14B & .079 (.529) & .164 (.436) & .132 (.661)\\
Ministral-3-8B & .093 (.627) & .163 (.513) & .217 (.761)\\
\bottomrule
\end{tabular}
\end{table}

\begin{table}[h]
\caption{Probe grid search space.}
\label{tab:e1}
\centering\small
\begin{tabular}{lcc}
\toprule
Dimension & Qwen3-14B & Ministral-3-8B\\
\midrule
Layers & 40 & 34\\
Regularization $C$ & \multicolumn{2}{c}{$\{0.0025,0.005,0.01,0.025,0.05,0.1,0.25,0.5,1.0,2.0,5.0\}$}\\
Readout & \multicolumn{2}{c}{$\{\text{residual},\ \text{MLP}\}$}\\
Total configurations & 1{,}760 & 1{,}496\\
\bottomrule
\end{tabular}
\end{table}

\paragraph{Layer and readout.} The all-layer sweep shows correctness- and
calibration-related signal already readable in fairly early layers, strengthening with depth and
forming a stable plateau over the middle of the network. Layers close to the output overfit more
easily out of domain. Both readouts carry the calibration signal---Qwen's winner is the residual
stream, Ministral's is the MLP output---and Table~\ref{tab:e2} shows the two are essentially
equivalent. On the macro average Qwen scores the same selection accuracy either way (.405) with
AUROC .848 versus .849, and residual is preferred only on the slightly lower ECE (.046 versus
.048). Ministral's macro selection accuracy is .353 versus .350 with AUROC .821 versus .819. In
both models the residual--MLP difference is far smaller than the Probe--Verbal gap, so the
conclusions do not depend on a single definition of the internal tensor.

\begin{table}[h]
\caption{Residual versus MLP readout, per dataset and macro-averaged over the four test sets.
(i) is the winner under the submission's original selection rule (TriviaQA, EntityQ and NQ-open validation), (ii) the best configuration of
the other readout type (Qwen: (i)~residual L24 $C{=}0.005$, (ii)~MLP L24
$C{=}0.01$, Ministral: (i)~MLP L21 $C{=}2.0$, (ii)~residual L21 $C{=}1.0$). These
test-set numbers describe configurations already fixed on validation data. They are not used for
selection. Bold marks the better of the two within a dataset.}
\label{tab:e2}
\centering\small
\begin{tabular}{llcccc}
\toprule
Dataset & Readout & sel.\ Acc $\uparrow$ & ECE $\downarrow$ & AUROC $\uparrow$ & Brier $\downarrow$\\
\midrule
\multicolumn{6}{l}{\emph{Qwen3-14B}}\\
TriviaQA  & (i) residual & .772 & \textbf{.015} & .948 & .081\\
          & (ii) MLP      & \textbf{.779} & .019 & .948 & .081\\
EntityQ   & (i) residual & \textbf{.310} & \textbf{.014} & \textbf{.913} & \textbf{.097}\\
          & (ii) MLP      & .302 & .026 & .909 & .100\\
NQ-open   & (i) residual & \textbf{.463} & .078 & .826 & .177\\
          & (ii) MLP      & .462 & \textbf{.065} & \textbf{.829} & \textbf{.174}\\
SimpleQA  & (i) residual & \textbf{.076} & \textbf{.075} & .704 & \textbf{.076}\\
          & (ii) MLP      & .075 & .082 & \textbf{.708} & .078\\
Macro     & (i) residual & .405 & \textbf{.046} & .848 & .108\\
          & (ii) MLP      & .405 & .048 & \textbf{.849} & .108\\
\addlinespace[2pt]
\multicolumn{6}{l}{\emph{Ministral-3-8B}}\\
TriviaQA  & (i) MLP      & \textbf{.718} & .014 & .917 & .105\\
          & (ii) residual & .714 & \textbf{.013} & \textbf{.918} & \textbf{.104}\\
EntityQ   & (i) MLP      & .225 & .024 & \textbf{.909} & \textbf{.091}\\
          & (ii) residual & .225 & \textbf{.015} & .901 & .095\\
NQ-open   & (i) MLP      & \textbf{.401} & \textbf{.108} & \textbf{.816} & \textbf{.188}\\
          & (ii) residual & .392 & .126 & .813 & .194\\
SimpleQA  & (i) MLP      & .068 & .146 & .640 & .116\\
          & (ii) residual & \textbf{.069} & \textbf{.127} & \textbf{.645} & \textbf{.109}\\
Macro     & (i) MLP      & \textbf{.353} & .073 & \textbf{.821} & .125\\
          & (ii) residual & .350 & \textbf{.070} & .819 & .125\\
\bottomrule
\end{tabular}
\end{table}

\begin{figure}[h]
\centering
\includegraphics[width=0.82\linewidth]{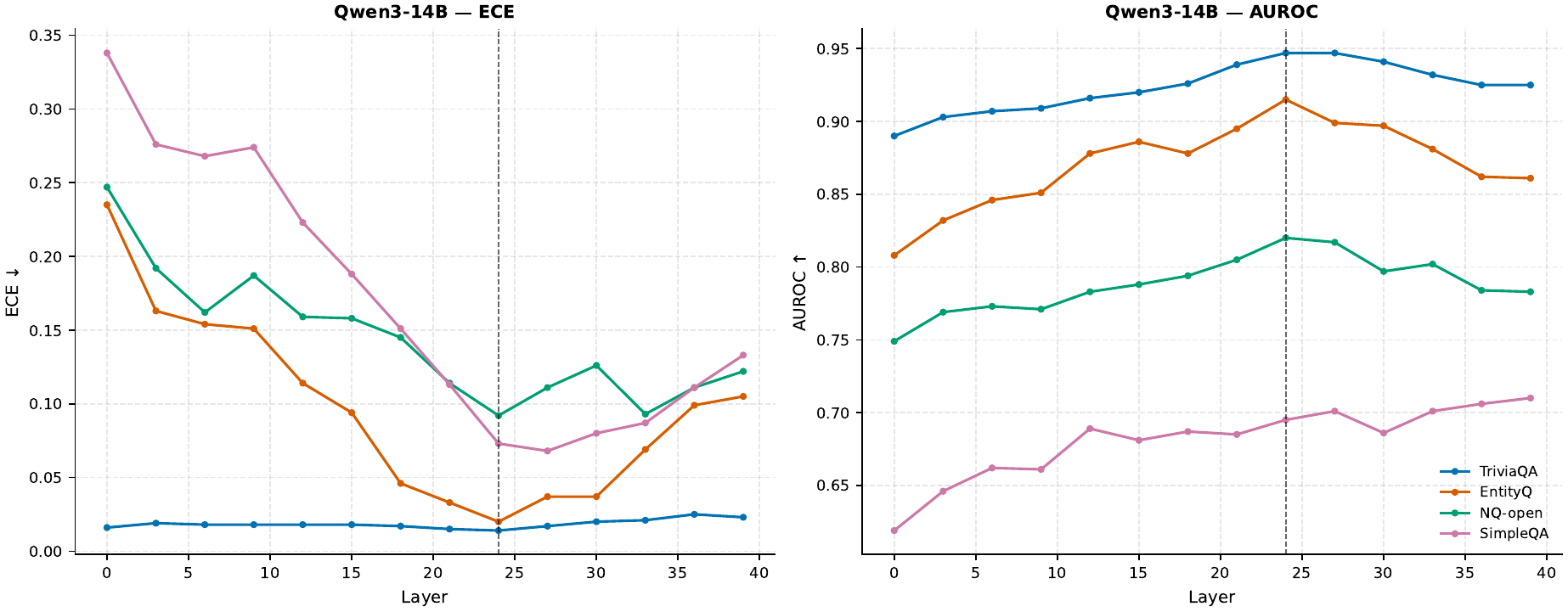}\\[3pt]
\includegraphics[width=0.82\linewidth]{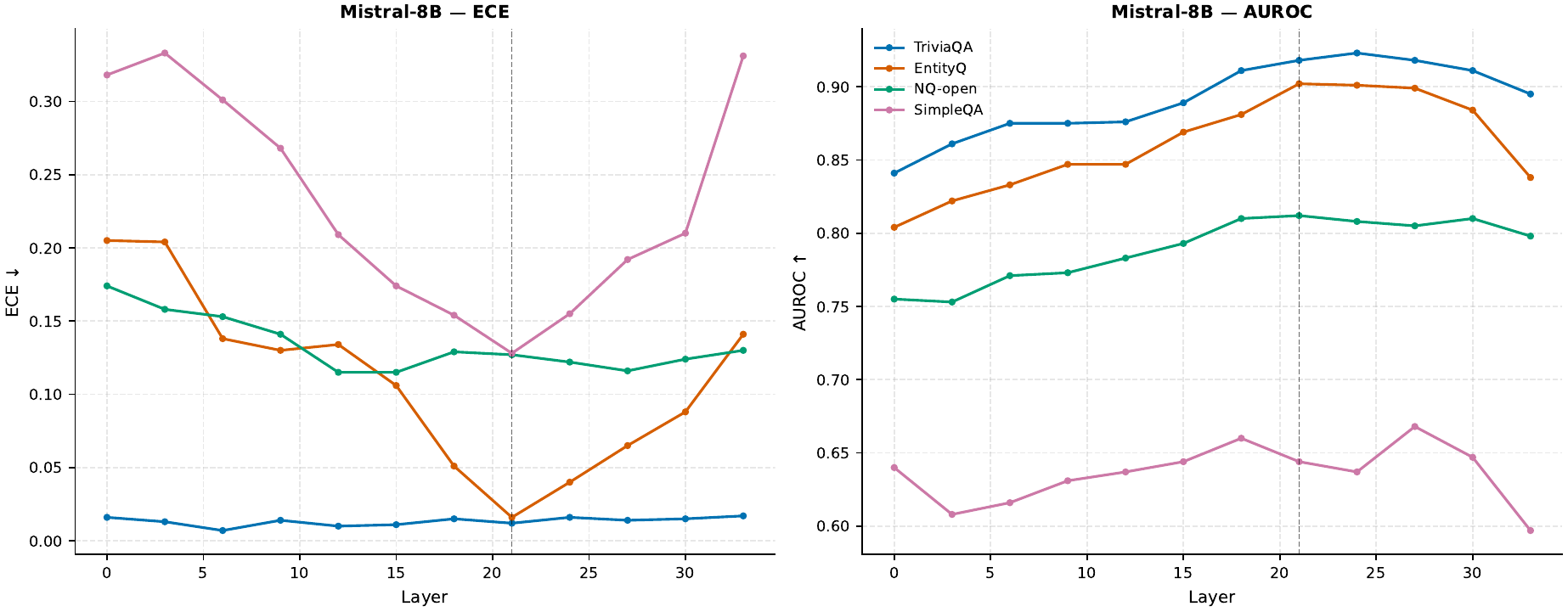}
\caption{Per-layer ECE and AUROC (\S\ref{sec:probe-robust}). Top: Qwen3-14B. Bottom:
Ministral-3-8B. Each line is one dataset and the dashed vertical line marks the best layer.}
\label{fig:layer-curves}
\end{figure}

\paragraph{Best-of-$N$ selection scaling at $N{=}32$.} To rule out a single-dataset artifact we
extend the best-of-$N$ analysis of \S\ref{sec:not-selection} to both models and all four
datasets, sampling 32 independent traces per question at $T{=}0.6$ and comparing selection by
maximum probe score, lexically normalized majority voting, and the oracle pass@$N$.

Selection accuracy forms a broad, flat plateau across layers (Figure~\ref{fig:select-acc}a):
0.748--0.778 over all 40 Qwen layers on TriviaQA and 0.431--0.471 on NQ-open, peaking at L27 and
L23 with under 5 points of variation, so deeper representations do not keep releasing new answer
knowledge. Panel~(b) is more discriminative: on TriviaQA at $N{=}16$, Max-Probe reaches 0.778 and
majority voting 0.768 while the oracle is already at 0.860. At $N{=}32$ the three are 0.778,
0.771, and 0.879.

\begin{table}[h]
\caption{Best-of-$N$ accuracy at $N{=}32$. Bold marks the better of Max-Probe and majority
voting. Oracle is the theoretical ceiling.}
\label{tab:e3}
\centering\small
\begin{tabular}{llccc}
\toprule
Model & Dataset & Oracle (pass@$N$) & Max-Probe & Majority vote\\
\midrule
\multirow{4}{*}{Qwen3-14B}
 & TriviaQA & .879 & \textbf{.778} & .771\\
 & EntityQ  & .447 & \textbf{.315} & .306\\
 & NQ-open  & .695 & .458 & \textbf{.492}\\
 & SimpleQA & .246 & .074 & \textbf{.077}\\
\midrule
\multirow{4}{*}{Ministral-3-8B}
 & TriviaQA & .844 & .717 & \textbf{.728}\\
 & EntityQ  & .363 & .231 & \textbf{.237}\\
 & NQ-open  & .631 & \textbf{.406} & .405\\
 & SimpleQA & .262 & .072 & \textbf{.078}\\
\bottomrule
\end{tabular}
\end{table}

Max-Probe and majority voting trade places with small margins throughout: Qwen is slightly ahead
on TriviaQA and EntityQ (.778 versus .771, .315 versus .306) and slightly behind on NQ-open and
SimpleQA (.458 versus .492, .074 versus .077). Ministral is overtaken on three of four datasets
and leads slightly on NQ-open. Both are far below the oracle on every dataset and the gap widens with $N$
(for Qwen on TriviaQA it grows from .024 at $N{=}2$ to .101 at $N{=}32$).
\clearpage
\section{Grading and Extraction Reliability}
\label{app:judge}

Qwen3-32B extracts the model's final answer and decides its correctness. To assess this pipeline
we sampled 200 already-graded question--answer pairs across 8 model columns and 4 datasets,
pre-screened them with string matching to raise the yield of anomalies, then manually reviewed
all 34 flagged cases plus 10 additional unflagged cases. The manual review inspects the question,
the reference answer, the raw model output, the extraction result and the grading rationale.
String matching is only a screening device and is never treated as ground truth.

\begin{table}[h]
\caption{Extraction and grading reliability under a 200-question human audit.}
\label{tab:f1}
\centering\small
\begin{tabular}{lcc}
\toprule
Metric & String pre-screen & After human review\\
\midrule
Extraction accuracy & 92.0\% (184/200) & \textbf{99.5\% (199/200)}\\
Grading accuracy (answerable) & 93.8\% (181/193) & \textbf{99.5\% (198/199)}\\
Grading precision & 83.1\% & \textbf{98.7\% (77/78)}\\
Grading recall & 100\% & \textbf{100\% (77/77)}\\
True grading errors & --- & \textbf{1/200 (0.5\%)}\\
\bottomrule
\end{tabular}
\end{table}

Extraction and grading both end at 99.5\%, but on different denominators: extraction is 199/200,
whereas grading accuracy is computed on the 199 samples the review confirmed to be answerable
(198/199). The grading denominator moves from 193 at the pre-screen stage to 199 because the
review re-partitioned abstentions, valid synthesized answers, and genuinely gradable responses.

\begin{table}[h]
\caption{Human classification of the audit sample. Left: extraction outcomes over all 200 cases.
Right: attribution of the cases the string pre-screen flagged as grading anomalies.}
\label{tab:f2}
\centering\small
\setlength{\tabcolsep}{4pt}
\begin{tabular}{lcl@{\hskip 8pt}lcl}
\toprule
Extraction outcome & $n$ & Verdict & Flagged grading case & $n$ & Verdict\\
\midrule
Exact or prefix match      & 175 & faithful      & More specific answer contains gold & 5 & correct\\
Substring match            & 9   & correct distillation & Alias not in gold list       & 3 & correct\\
Surface mismatch, valid    & 5   & valid synthesis & Functional description $\equiv$ entity & 2 & correct\\
Empty: explicit abstention & 10  & correct empty & Output contains gold substring     & 3 & correct\\
Empty: answer was given    & 1   & \textbf{true error} & Subset or containment relation & 4 & correct\\
                           &     &               & Diacritic difference               & 1 & correct\\
Total                      & 200 & 199 valid     & \textbf{True grading error}        & 1 & false accept\\
\bottomrule
\end{tabular}
\end{table}

The string pre-screen underestimates extraction quality. All ten empty outputs come from
responses that explicitly said ``Unknown'', ``I don't know'' or ``not available'', or that
rejected the premise of the question, so they are correctly recognized abstentions rather than
missed answers. Five surface-mismatch cases come from responses that ignored the single-answer
format. The extractor distilled an entity or a list from long answers about the inventor of the
CPU, the oldest house in the United States, a Revolutionary War battle, the start and end dates
of the Soviet Union, and the next Olympic Games, and the review confirmed all five as valid
syntheses. The single true extraction error is a response that stated ``Singapore Island, 55''
for which the extractor returned an empty string. Whether or not the answer was right, an
explicitly stated answer should never be discarded.

Abstentions also raise a protocol difference between \texttt{INCORRECT} and
\texttt{NOT\_ATTEMPTED}. Our end-to-end convention keeps explicit abstentions in the accuracy
denominator and counts them as errors, whereas the strict SimpleQA taxonomy would label them
\texttt{NOT\_ATTEMPTED}. This changes the name of the error type but not the end-to-end accuracy
we report.

Of the 19 suspicious grading cases, 18 are false alarms produced by overly strict string rules.
The single true grading error is a question about a 2010 South Africa World Cup stadium where the
model substituted Moses Mabhida Stadium for Cape Town Stadium and the grader mistakenly treated
Moses Mabhida Stadium and Nelson Mandela Bay Stadium as aliases. These are different stadiums, so
the response should have been marked incorrect. All 10 additional unflagged cases were graded
correctly.

Average label noise is therefore far smaller than the gaps between the methods we compare, but a
low marginal error rate does not establish independence of errors. The extractor, the grader and
the probe may still fail together on long-tail entities, ambiguous questions or malformed
outputs. We consequently do not treat automatic grading as unbiased ground truth and do not
exclude correlation between grading error and probe error. The main conclusions are constrained
jointly by the cross-dataset results, this human audit, and the recall metric of
Appendix~\ref{app:priming}, which does not depend on an LLM judge.

\paragraph{Human robustness check for calibration labels.}
To test whether automatic grading can change the calibration conclusions, we randomly sampled 500
rows from the 10,000-row TriviaQA pool behind the Phase A calibration tables and manually reviewed
every row whose extracted answer was not an exact string match to the reference answer or its
aliases. The full-pool and subset metrics are:

\begin{table}[h]
\caption{Full-pool versus 500-row-subset metrics. The subset is used to audit labels, not to
re-estimate the final ECE.}
\label{tab:f3}
\centering\small
\begin{tabular}{lrrrrrr}
\toprule
Source & \multicolumn{2}{c}{ECE} & \multicolumn{2}{c}{AUROC} & \multicolumn{2}{c}{Brier}\\
\cmidrule(lr){2-3}\cmidrule(lr){4-5}\cmidrule(lr){6-7}
 & Full & Subset & Full & Subset & Full & Subset\\
\midrule
Probe  & .0147 & .0363 & .9477 & .9621 & .0809 & .0718\\
Verbal & .1782 & .1905 & .8243 & .8511 & .1839 & .1877\\
Token  & .2514 & .2602 & .6360 & .6181 & .2530 & .2586\\
\bottomrule
\end{tabular}
\end{table}

Across ten resamples of 500 rows, probe ECE ranges from .028 to .053, while AUROC and Brier are
more stable. ECE is therefore sensitive to the audit sample size, so this subset is useful for
testing label reliability but not for reproducing the full-pool ECE.

Among the 500 rows, 258 exactly matched the reference answer and 28 exactly matched an alias. The
remaining 214 were manually reviewed. The grading decision was confirmed for 210 rows, with one
clear grading error, two boundary cases, and one protocol issue in which \textquotedblleft I
don't know\textquotedblright{} was routed as \texttt{INCORRECT}. The review exposed one clear false
rejection caused by a defective reference answer and one boundary false acceptance of an answer
with incorrect extra information, exposing residual risks in both the gold data and semantic
grading.
These 214 rows were selected by string mismatch and therefore do not provide an unbiased estimate
of the overall error rate.

After manually correcting the two semantic decisions judged to require correction, subset ECE is .0363, .1905, and .2602
for the probe, verbal, and token sources, respectively, unchanged from the original values. Probe
AUROC changes from .9621 to .9624 and Brier from .0718 to .0715. The audit correction therefore
does not change the calibration gap between methods. It does not establish independence between
grader and probe errors. Qwen3-32B remains an auditable but fallible proxy for correctness.

\clearpage
\section{Robustness of the Confidence Comparison}
\label{app:brier}

This appendix supplements Table~\ref{tab:probe-vs-verbal} with the per-term Brier decomposition
$\mathrm{Brier}=\mathrm{Rel}-\mathrm{Res}+\mathrm{Unc}$ for both models and both confidence
sources. The identity is \emph{approximate}: it is exact only when the score variance inside
each of the 10 equal-width bins is negligible. Probe scores are tightly concentrated within
bins and reconstruct the Brier score to within $\pm0.001$. Verbalized scores span a wide range
inside the same bin, so the within-bin variance is not captured by Rel or Res and the verbal
rows show a systematic $\pm0.003$--$0.006$ residual. This is a property of the binning, not a
transcription error, and was verified against the source data.

\begin{table}[h]
\caption{Brier score and its decomposition into Reliability ($\downarrow$), Resolution
($\uparrow$) and Uncertainty. Uncertainty depends on the base accuracy of the scorable subset and is
therefore comparable only within a (model, dataset) cell. Bold marks the better value per cell.}
\label{tab:a1}
\centering\small
\begin{tabular}{llcccc}
\toprule
Dataset (Acc) & Source & Brier $\downarrow$ & Rel $\downarrow$ & Res $\uparrow$ & Unc\\
\midrule
\multicolumn{6}{l}{\emph{Qwen3-14B}}\\
TriviaQA (.729) & Verbal & .184 & .052 & .059 & \textbf{.194}\\
                & Probe  & \textbf{.081} & \textbf{.001} & \textbf{.116} & .197\\
EntityQ (.267)  & Verbal & .455 & .297 & .040 & .207\\
                & Probe  & \textbf{.097} & \textbf{.000} & \textbf{.098} & \textbf{.196}\\
NQ-open (.440)  & Verbal & .416 & .207 & .023 & .249\\
                & Probe  & \textbf{.177} & \textbf{.009} & \textbf{.077} & \textbf{.246}\\
SimpleQA (.066) & Verbal & .532 & .480 & .001 & .067\\
                & Probe  & \textbf{.076} & \textbf{.019} & \textbf{.004} & \textbf{.061}\\
\addlinespace[2pt]
\multicolumn{6}{l}{\emph{Ministral-3-8B}}\\
TriviaQA (.664) & Verbal & .263 & .068 & .018 & \textbf{.216}\\
                & Probe  & \textbf{.105} & \textbf{.000} & \textbf{.112} & .218\\
EntityQ (.207)  & Verbal & .564 & .413 & .016 & .170\\
                & Probe  & \textbf{.091} & \textbf{.001} & \textbf{.075} & \textbf{.164}\\
NQ-open (.370)  & Verbal & .490 & .268 & .013 & .238\\
                & Probe  & \textbf{.188} & \textbf{.021} & \textbf{.066} & \textbf{.234}\\
SimpleQA (.061) & Verbal & .671 & .612 & .000 & .059\\
                & Probe  & \textbf{.116} & \textbf{.058} & \textbf{.001} & .059\\
\bottomrule
\end{tabular}
\end{table}

\begin{figure}[h]
\centering
\includegraphics[width=0.49\linewidth]{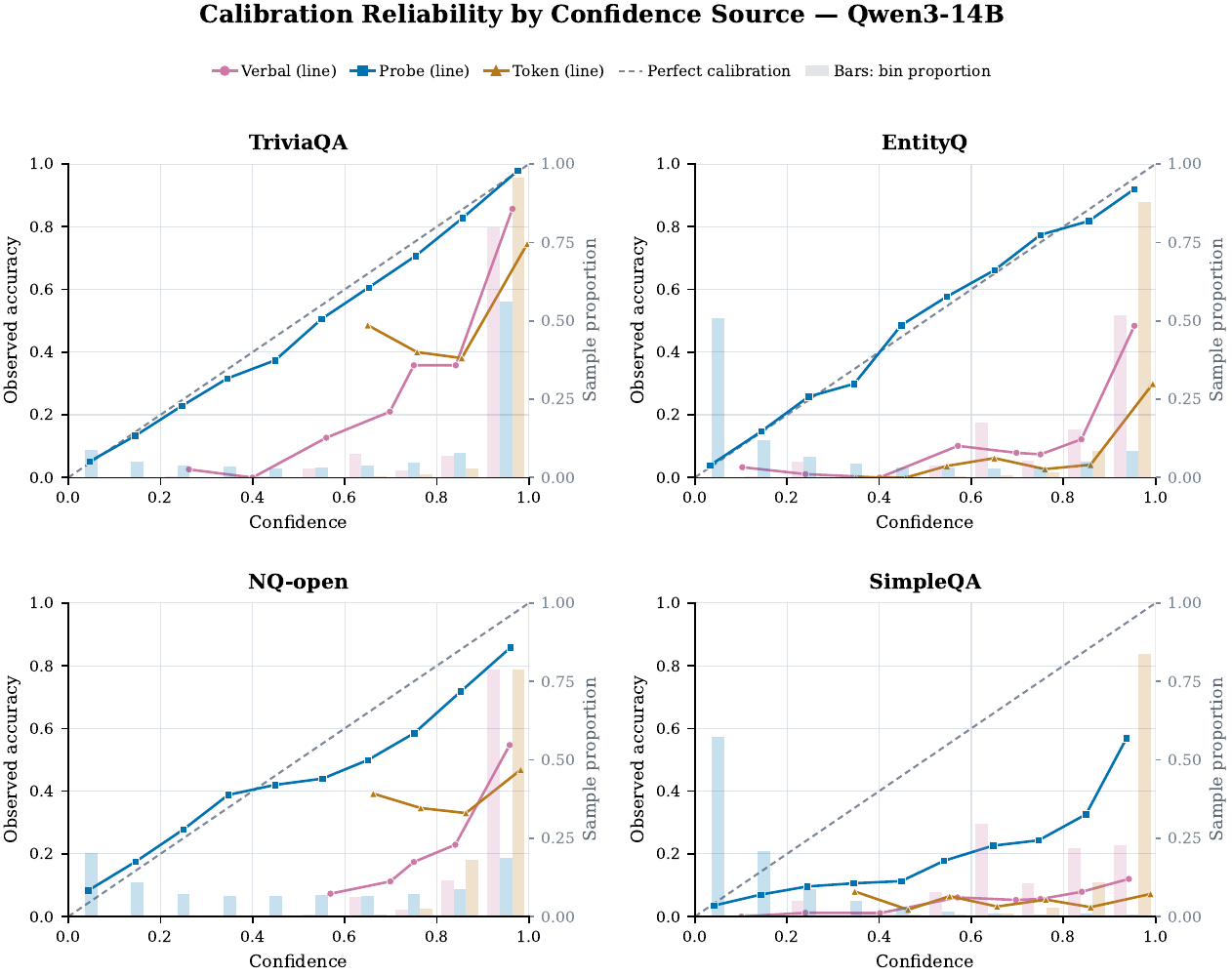}\hfill
\includegraphics[width=0.49\linewidth]{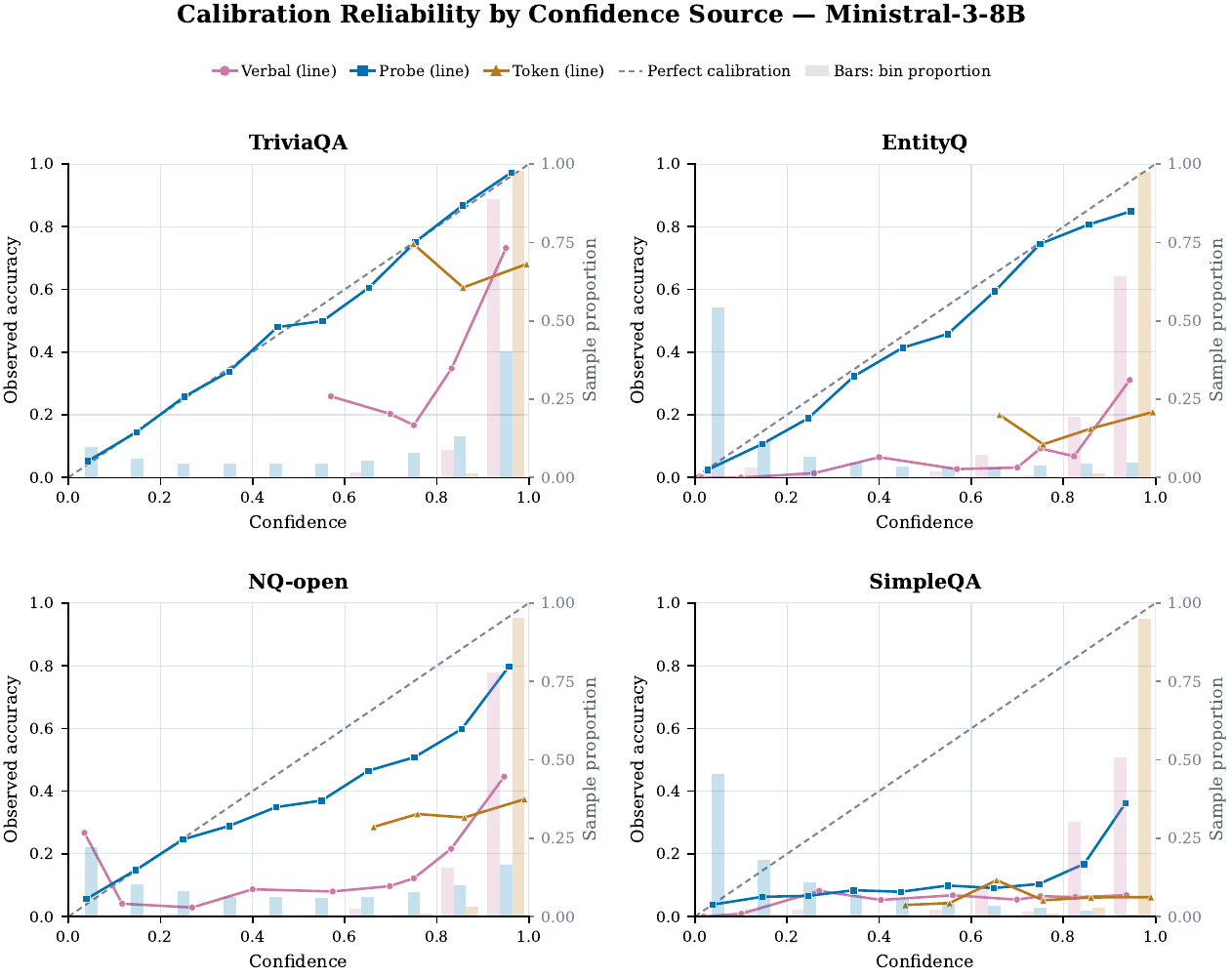}
\caption{Reliability diagrams over four datasets and three sources (verbal / probe / token
probability). Left: Qwen3-14B. Right: Ministral-3-8B. In each panel the solid line connects
per-bin mean confidence to observed accuracy (left axis) and the dashed line is perfect
calibration. Pale bars give the fraction of samples in each of the 10 equal-width confidence
bins (right axis).}
\label{fig:reliability}
\end{figure}

The pattern is identical in both families: the probe's Reliability is essentially zero (its
probability scale is accurate) and its Resolution is the higher of the two (it separates correct
from incorrect traces), whereas the verbalized score has high Reliability error (systematic
overconfidence) and low Resolution. Figure~\ref{fig:reliability} gives the per-bin version of
this comparison: the probe curve lies close to the diagonal, while the verbalized and
token-probability curves concentrate their error in the high-confidence region.

\subsection{Sensitivity to Missing-Score Selection Bias}
\label{app:missing}

ECE, AUROC and Brier can only be computed on samples for which a verbalized score is
successfully parsed. Since format compliance differs across methods, the main tables report
calibration conditional on the $n_{scored}$ subset, and a method with low coverage could benefit
from a different set of admitted samples. We check robustness under three conventions.
\textbf{Convention 1} is the main-table convention with denominator $n_{scored}$. There, the
extract rate measures how often an answer can be extracted and coverage how often a valid score
is present, and their difference is the set of samples whose answer is gradable but whose score is
missing. \textbf{Convention 2} keeps, per seed, only the questions for which every main-table column has
at least 5 of 10 valid scores, and recomputes ECE, AUROC and Brier on that common question set.
\textbf{Convention 3} assigns a score of 0 to every missing score and recomputes ECE and Brier
over all samples. Because this creates a large tie at 0, AUROC is not reported.

Both the common-subset and full-sample results pool the three training seeds 42/43/44. Verbal has
seed 42 only and is marked $^{*}$. Error terms are question-level bootstrap half-widths with
$B{=}5000$. Qwen's main-table columns are \probesd{}, Probe-SD-LRM, \probesd{} w/o CW, SC-SD and Verbal.
Ministral omits the \emph{w/o CW} variant, and its rows use the submitted checkpoints, trained with the
original label probe. The readout probe has a different coverage mechanism and is
reported separately. The standalone probe results in Table~\ref{tab:probe-vs-verbal} use the Phase-A
evaluation batch, whereas the readout-probe check below reuses the Probe baseline from the SFT
evaluation suite. They use the same L24 residual, $C{=}0.005$ recipe, but not the same evaluation
batch or coverage, so their point estimates need not match.

\begin{table}[h]
\caption{Convention 2 (common subset). Common question counts are: EntityQ 982/998, NQ-open
690/893, SimpleQA 846/929, TriviaQA 989/981 for Qwen/Ministral. Bold marks the best method per
column within a model.}
\label{tab:i1}
\centering\small
\setlength{\tabcolsep}{3pt}
\begin{tabular}{llccccc}
\toprule
Model & Method & TriviaQA & EntityQ & NQ-open & SimpleQA & Macro\\
\midrule
\multicolumn{7}{l}{\emph{ECE} $\downarrow$}\\
Qwen & \probesd{}        & \bci{.023}{.006} & \bci{.090}{.008} & \bci{.124}{.012} & \bci{.114}{.007} & \bci{.088}{.004}\\
Qwen & Verbal$^{*}$      & \ci{.177}{.021} & \ci{.531}{.022} & \ci{.423}{.032} & \ci{.665}{.015} & \ci{.449}{.012}\\
Qwen & Probe-SD-LRM      & \ci{.030}{.007} & \ci{.102}{.008} & \ci{.163}{.013} & \ci{.136}{.007} & \ci{.108}{.004}\\
Qwen & \probesd{} w/o CW  & \ci{.032}{.006} & \ci{.105}{.008} & \ci{.146}{.012} & \ci{.126}{.007} & \ci{.102}{.004}\\
Qwen & SC-SD             & \ci{.063}{.008} & \ci{.224}{.009} & \ci{.241}{.013} & \ci{.237}{.009} & \ci{.191}{.005}\\
Min. & \probesd{}        & \bci{.043}{.006} & \bci{.087}{.009} & \bci{.141}{.011} & \ci{.162}{.007} & \bci{.108}{.004}\\
Min. & Verbal$^{*}$      & \ci{.246}{.024} & \ci{.627}{.022} & \ci{.515}{.027} & \ci{.767}{.014} & \ci{.539}{.011}\\
Min. & Probe-SD-LRM      & \ci{.064}{.007} & \ci{.108}{.009} & \ci{.164}{.011} & \bci{.158}{.007} & \ci{.123}{.004}\\
Min. & SC-SD             & \ci{.051}{.005} & \ci{.150}{.009} & \ci{.175}{.011} & \ci{.184}{.007} & \ci{.140}{.004}\\
\addlinespace[2pt]
\multicolumn{7}{l}{\emph{AUROC} $\uparrow$}\\
Qwen & \probesd{}        & \bci{.929}{.007} & \bci{.881}{.009} & \bci{.809}{.013} & \bci{.668}{.024} & \bci{.821}{.007}\\
Qwen & Verbal$^{*}$      & \ci{.826}{.017} & \ci{.811}{.017} & \ci{.716}{.022} & \ci{.625}{.043} & \ci{.745}{.013}\\
Qwen & Probe-SD-LRM      & \ci{.926}{.007} & \ci{.868}{.010} & \ci{.804}{.013} & \ci{.654}{.027} & \ci{.813}{.008}\\
Qwen & \probesd{} w/o CW  & \ci{.919}{.007} & \ci{.872}{.009} & \ci{.800}{.013} & \ci{.644}{.024} & \ci{.809}{.007}\\
Qwen & SC-SD             & \ci{.877}{.009} & \ci{.847}{.009} & \ci{.758}{.012} & \ci{.658}{.020} & \ci{.785}{.007}\\
Min. & \probesd{}        & \ci{.891}{.008} & \bci{.879}{.009} & \bci{.811}{.012} & \bci{.683}{.025} & \bci{.816}{.008}\\
Min. & Verbal$^{*}$      & \ci{.669}{.015} & \ci{.734}{.016} & \ci{.640}{.018} & \ci{.523}{.041} & \ci{.641}{.012}\\
Min. & Probe-SD-LRM      & \bci{.904}{.007} & \ci{.875}{.010} & \bci{.811}{.012} & \ci{.641}{.027} & \ci{.808}{.008}\\
Min. & SC-SD             & \ci{.809}{.009} & \ci{.821}{.011} & \ci{.733}{.011} & \ci{.612}{.021} & \ci{.744}{.007}\\
\addlinespace[2pt]
\multicolumn{7}{l}{\emph{Brier} $\downarrow$}\\
Qwen & \probesd{}        & \bci{.092}{.005} & \bci{.131}{.006} & \bci{.197}{.008} & \bci{.106}{.006} & \bci{.132}{.003}\\
Qwen & Verbal$^{*}$      & \ci{.183}{.017} & \ci{.457}{.018} & \ci{.404}{.028} & \ci{.536}{.013} & \ci{.395}{.010}\\
Qwen & Probe-SD-LRM      & \ci{.096}{.005} & \ci{.137}{.006} & \ci{.213}{.009} & \ci{.109}{.006} & \ci{.139}{.003}\\
Qwen & \probesd{} w/o CW  & \ci{.096}{.005} & \ci{.140}{.006} & \ci{.208}{.009} & \ci{.116}{.006} & \ci{.140}{.003}\\
Qwen & SC-SD             & \ci{.116}{.006} & \ci{.212}{.007} & \ci{.263}{.010} & \ci{.185}{.006} & \ci{.194}{.004}\\
Min. & \probesd{}        & \bci{.122}{.005} & \bci{.126}{.005} & \bci{.200}{.007} & \ci{.122}{.005} & \bci{.142}{.003}\\
Min. & Verbal$^{*}$      & \ci{.262}{.021} & \ci{.564}{.019} & \ci{.493}{.024} & \ci{.678}{.012} & \ci{.499}{.010}\\
Min. & Probe-SD-LRM      & \ci{.127}{.005} & \ci{.127}{.005} & \ci{.204}{.007} & \bci{.113}{.005} & \ci{.143}{.003}\\
Min. & SC-SD             & \ci{.153}{.006} & \ci{.174}{.006} & \ci{.237}{.006} & \ci{.166}{.005} & \ci{.183}{.003}\\
\bottomrule
\end{tabular}
\end{table}

On Qwen, \probesd{} is best on all four datasets and on the macro average for all three metrics.
Ministral has local exceptions---Probe-SD-LRM is better on TriviaQA AUROC and on SimpleQA
ECE/Brier, and the two tie on NQ-open AUROC---but the hierarchy is unchanged: probe-based methods
beat SC-SD, and both beat Verbal by a wide margin.

\begin{table}[h]
\caption{Convention 3 (full-sample attribution, missing scores set to 0). AUROC is not reported
because of the tie mass at 0.}
\label{tab:i2}
\centering\small
\setlength{\tabcolsep}{3pt}
\begin{tabular}{llccccc}
\toprule
Model & Method & TriviaQA & EntityQ & NQ-open & SimpleQA & Macro\\
\midrule
\multicolumn{7}{l}{\emph{ECE} $\downarrow$}\\
Qwen & \probesd{}        & \bci{.024}{.009} & \bci{.089}{.013} & \bci{.162}{.018} & \bci{.104}{.010} & \bci{.095}{.006}\\
Qwen & Verbal$^{*}$      & \ci{.182}{.020} & \ci{.493}{.020} & \ci{.410}{.022} & \ci{.545}{.018} & \ci{.408}{.010}\\
Qwen & Probe-SD-LRM      & \ci{.031}{.010} & \ci{.101}{.014} & \ci{.205}{.018} & \ci{.122}{.010} & \ci{.115}{.007}\\
Qwen & \probesd{} w/o CW  & \ci{.033}{.010} & \ci{.105}{.013} & \ci{.182}{.017} & \ci{.114}{.010} & \ci{.109}{.006}\\
Qwen & SC-SD             & \ci{.064}{.013} & \ci{.222}{.016} & \ci{.266}{.019} & \ci{.208}{.012} & \ci{.190}{.007}\\
Min. & \probesd{}        & \bci{.045}{.008} & \bci{.087}{.014} & \bci{.146}{.018} & \ci{.159}{.011} & \bci{.109}{.007}\\
Min. & Verbal$^{*}$      & \ci{.245}{.023} & \ci{.596}{.021} & \ci{.469}{.020} & \ci{.680}{.016} & \ci{.498}{.010}\\
Min. & Probe-SD-LRM      & \ci{.066}{.011} & \ci{.106}{.014} & \ci{.165}{.018} & \bci{.152}{.010} & \ci{.122}{.007}\\
Min. & SC-SD             & \ci{.054}{.007} & \ci{.147}{.014} & \ci{.178}{.016} & \ci{.175}{.010} & \ci{.138}{.006}\\
\addlinespace[2pt]
\multicolumn{7}{l}{\emph{Brier} $\downarrow$}\\
Qwen & \probesd{}        & \bci{.092}{.009} & \bci{.130}{.010} & \bci{.224}{.013} & \bci{.097}{.008} & \bci{.136}{.005}\\
Qwen & Verbal$^{*}$      & \ci{.188}{.017} & \ci{.424}{.017} & \ci{.397}{.020} & \ci{.439}{.015} & \ci{.362}{.009}\\
Qwen & Probe-SD-LRM      & \ci{.097}{.009} & \ci{.135}{.009} & \ci{.245}{.015} & \ci{.099}{.008} & \ci{.144}{.005}\\
Qwen & \probesd{} w/o CW  & \ci{.097}{.009} & \ci{.139}{.010} & \ci{.237}{.014} & \ci{.105}{.008} & \ci{.144}{.005}\\
Qwen & SC-SD             & \ci{.116}{.010} & \ci{.210}{.011} & \ci{.285}{.015} & \ci{.165}{.008} & \ci{.194}{.006}\\
Min. & \probesd{}        & \bci{.124}{.009} & \bci{.125}{.009} & \bci{.203}{.011} & \ci{.121}{.008} & \ci{.143}{.005}\\
Min. & Verbal$^{*}$      & \ci{.260}{.020} & \ci{.537}{.019} & \ci{.451}{.017} & \ci{.601}{.015} & \ci{.462}{.009}\\
Min. & Probe-SD-LRM      & \ci{.129}{.008} & \ci{.125}{.009} & \ci{.205}{.011} & \bci{.110}{.008} & \bci{.142}{.005}\\
Min. & SC-SD             & \ci{.154}{.010} & \ci{.171}{.010} & \ci{.238}{.010} & \ci{.161}{.008} & \ci{.181}{.005}\\
\bottomrule
\end{tabular}
\end{table}

Full-sample attribution preserves the overall ordering. This convention is not an upper bound: if
the missing samples are mostly wrong answers, assigning them 0 makes their ECE contribution close
to zero and can \emph{lower} the total. Qwen's Verbal ECE on NQ-open, for instance, drops from
.436 under the conditional convention to .410 under full-sample attribution. The common
subset is therefore our primary test for selection bias, with full-sample attribution as a
supplementary view.

\paragraph{Readout probe under all three conventions.} We evaluate the Qwen readout-probe baseline
(L24 residual, $C{=}0.005$) on the seed-42 run of the SFT evaluation suite, using the same common
question set as the main-table columns and again assigning 0 to missing probe scores, with bootstrap
$B{=}5000$. This is distinct from the standalone probe evaluation in Table~\ref{tab:probe-vs-verbal}:
the recipe is the same, but the evaluation batch and coverage differ. Table~\ref{tab:i3} therefore
tests within-run stability across denominator conventions rather than reproducing Table~\ref{tab:probe-vs-verbal}.

\begin{table}[h]
\caption{Qwen readout-probe baseline from the SFT evaluation suite under the three denominator
conventions. Brackets are 95\% bootstrap intervals.}
\label{tab:i3}
\centering\small
\setlength{\tabcolsep}{4pt}
\begin{tabular}{llccc}
\toprule
Dataset & Convention & ECE $\downarrow$ & AUROC $\uparrow$ & Brier $\downarrow$\\
\midrule
\multirow{3}{*}{TriviaQA} & scorable subset & .014 [.010,.026] & .948 [.938,.957] & .080 [.072,.089]\\
 & common subset ($n_q{=}989$) & .013 [.010,.025] & .948 [.938,.957] & .080 [.071,.089]\\
 & full-sample attribution & .015 [.010,.027] & --- & .080 [.072,.088]\\
\midrule
\multirow{3}{*}{EntityQ} & scorable subset & .022 [.015,.038] & .898 [.882,.914] & .109 [.099,.120]\\
 & common subset ($n_q{=}982$) & .023 [.015,.039] & .899 [.883,.915] & .109 [.099,.120]\\
 & full-sample attribution & .020 [.014,.035] & --- & .099 [.090,.109]\\
\midrule
\multirow{3}{*}{NQ-open} & scorable subset & .074 [.058,.093] & .817 [.796,.837] & .181 [.168,.193]\\
 & common subset ($n_q{=}690$) & .067 [.048,.090] & .834 [.809,.856] & .172 [.157,.187]\\
 & full-sample attribution & .070 [.054,.089] & --- & .175 [.163,.187]\\
\midrule
\multirow{3}{*}{SimpleQA} & scorable subset & .089 [.077,.103] & .670 [.615,.720] & .089 [.080,.099]\\
 & common subset ($n_q{=}846$) & .092 [.080,.108] & .658 [.600,.712] & .096 [.085,.107]\\
 & full-sample attribution & .081 [.069,.094] & --- & .079 [.070,.088]\\
\midrule
\multirow{3}{*}{\textbf{Macro}} & scorable subset & .050 & .833 & .115\\
 & common subset & .049 & .835 & .114\\
 & full-sample attribution & .047 & --- & .108\\
\bottomrule
\end{tabular}
\end{table}

The probe's macro ECE is .050, .049 and .047 under the three conventions and its Brier .115, .114
and .108, so its results are stable under changes of denominator. Combining the main-table columns
with the readout probe, we find that \probesd{}'s calibration advantage cannot be explained by missing-score
selection or by the probe's higher coverage. This conclusion concerns selection bias induced by
missing scores only. It does not address grading error, answer-extraction error, or the causal
origin of the missingness mechanism itself.
\clearpage
\section{Training Hyperparameters and Probe Configurations}
\label{app:hparams}

All SFT methods use full-parameter finetuning with \texttt{swift}
\citep{zhao2024swiftascalablelightweightinfrastructure} and three training seeds (42/43/44), and
differ only in the source of the training data. The two models share most hyperparameters. The
differences are confined to the learning rate and confidence weighting
(Table~\ref{tab:c1}).

\begin{table}[h]
\caption{Supervised finetuning hyperparameters.}
\label{tab:c1}
\centering\small
\setlength{\tabcolsep}{5pt}
\begin{tabular}{llll}
\toprule
Hyperparameter & Qwen3-14B & Ministral-3-8B & Note\\
\midrule
Training set size & 10{,}000 & 10{,}000 & identical across methods\\
Learning rate & $5\times10^{-6}$ & $3\times10^{-6}$ & \\
Epochs & 2 & 2 & ${\approx}625$ steps\\
LR schedule & cosine to 0 & cosine to 0 & 50 warmup steps\\
Weight decay & 0.1 & 0.1 & \\
Adam $\beta_1/\beta_2$ & 0.9 / 0.95 & 0.9 / 0.95 & \\
Per-device batch & 1 & 1 & \\
Gradient accumulation & 4 & 4 & effective batch $1{\times}4{\times}8{=}32$\\
Max sequence length & 8192 & 8192 & \\
Precision & bf16 & bf16 & \\
Optimizer sharding & DeepSpeed ZeRO-1 & DeepSpeed ZeRO-1 & 8 GPUs\\
Confidence weighting & $\omega{=}3$ & not used & applied to score tokens only\\
Initialization & Qwen3-14B-Base & Ministral-3-8B-Base-2512 & base checkpoint\\
\bottomrule
\end{tabular}
\end{table}

Confidence weighting multiplies the cross-entropy loss on the verbalized score tokens by
$\omega$. Qwen3-14B uses $\omega{=}3$ to offset the gradient dilution caused by long chains of
thought. Ministral-3-8B rehearses the score inside and outside its thinking block, which already
duplicates the score tokens and acts as implicit weighting, so no extra weighting is applied.

\begin{table}[h]
\caption{Probe configurations selected independently for each purpose and sampling temperature.
Selection uses validation data only (Appendix~\ref{app:grid}).}
\label{tab:c2}
\centering\small
\begin{tabular}{llll}
\toprule
Purpose & $T$ & Qwen3-14B & Ministral-3-8B\\
\midrule
Evaluation probe scoring & 0.6 & L24 residual, $C{=}0.005$ & L20 MLP, $C{=}1.0$\\
\probesd{} training labels & 1.5 & L24 residual, $C{=}0.005$ & L24 residual, $C{=}0.25$\\
\bottomrule
\end{tabular}
\end{table}

Each temperature is searched independently with the same all-layer procedure on the TriviaQA
validation split: we compute selection accuracy, ECE and AUROC, rank the three metrics separately
and take the configuration with the lowest mean rank. The submitted version also ranked OOD-macro
ECE and AUROC on EntityQuestions and NQ-open validation data. Dropping them leaves Qwen unchanged and
moves Ministral by a few layers (Appendix~\ref{app:grid}). Qwen selects L24 residual with
$C{=}0.005$ at both temperatures. Ministral's two winners are layers 20 and 24 with
different readouts and $C$. The all-layer sweep shows that the calibration signal does not
emerge abruptly at a particular relative depth but is readable across many layers and forms a
broad plateau over the middle of the network. L24/40 and L20/34 are the specific winners picked
on that plateau by the validation composite rank and should not be read as the only effective
layers. Transferring the $T{=}0.6$ probe directly to $T{=}1.5$ samples instead of matching
temperatures costs Qwen almost nothing (macro ECE .043 against .047 for the temperature-matched
probe, i.e.\ direct transfer is even marginally better) but cost Ministral substantially under the
original selection (.100 against .054), which is why we reselect per temperature rather than reuse
the evaluation probe.

\begin{table}[h]
\caption{Training-data generation temperature ablation for Qwen3-14B (\S\ref{sec:sampling}). All
resulting models are evaluated directly at $T{=}0.6$. Errors are 95\% question-level bootstrap
half-widths.}
\label{tab:c3}
\centering\small
\begin{tabular}{lcccc}
\toprule
Training pool & TriviaQA & NQ-open & EntityQ & SimpleQA\\
\midrule
$T{=}0.6$ & \ci{.739}{.024} & \ci{.470}{.027} & \ci{.289}{.025} & \ci{.077}{.013}\\
$T{=}1.0$ & \ci{.738}{.024} & \ci{.467}{.026} & \ci{.288}{.025} & \ci{.078}{.013}\\
$T{=}1.5$ & \ci{.744}{.024} & \ci{.470}{.026} & \ci{.291}{.025} & \ci{.076}{.012}\\
$T{=}1.5$, no top-$k$/top-$p$ & \ci{.745}{.024} & \ci{.471}{.026} & \ci{.286}{.024} & \ci{.078}{.012}\\
\bottomrule
\end{tabular}
\end{table}

High-temperature sampling gives a modest, dataset-dependent gain at best: relative to $T{=}0.6$,
$T{=}1.5$ improves TriviaQA by 0.5 points and EntityQ by 0.2 points, leaves NQ-open unchanged,
and lowers SimpleQA by 0.1 points. All differences fall within the bootstrap half-widths. We
therefore keep $T{=}1.5$ for candidate diversity rather than claim a reliable accuracy gain, and
apply top-$k$/top-$p$ truncation to suppress generation breakdown, in particular Ministral's
formatting and repetition failures at high temperature.

After training, Ministral-3-8B weights must be converted from HF format to Ministral-native
format for vLLM deployment. The conversion has to apply the inverse of the wq/wk head
interleaving permutation. Without it the Q/K heads are scrambled and the output degenerates into
repetitive noise. Qwen3-14B is deployed directly in HF format.
\clearpage
\section{Complete Per-Dataset Results}
\label{app:full-results}

Tables~\ref{tab:d1} and~\ref{tab:d2} expand Table~\ref{tab:main} into all four datasets. SFT
methods (SC-SD, EA-SD, Probe-SD-LRM, \probesd{}) are averaged over the three training seeds 42/43/44.
Verbal, the frozen Probe, Verbal-Iso, Verbal-TS+bias, Answer Prob.\ and Surface-Feat use the seed-42 evaluation
run, with $^{*}$ marking the single-run baselines. All $\pm$ values are 95\% question-level bootstrap half-widths with $B{=}5000$
and bootstrap seed 42. For SFT methods each training seed is evaluated first and the three seeds
are then pooled. Out-of-domain macro half-widths in Table~\ref{tab:main} are composed from the
three datasets' bootstrap variances rather than resampled from the macro average itself. Where ECE
is close to 0 the bootstrap interval is right-skewed, so the reported $\pm$ half-width is a
symmetric approximation. ECE uses 10 equal-width bins. AUROC and Brier are computed on the scorable
subset. A dash marks a readout or post-hoc method that reuses the Verbal answers and therefore
has no accuracy of its own.
The frozen Probe scores the original model's sampled answers rather than the student's outputs.
Ministral's \probesd{} rows use the TriviaQA-only label probe (Appendix~\ref{app:grid}). Its Frozen
Probe rows and the Ministral rows of Appendix~\ref{app:missing} come from the submitted evaluation
suite and checkpoints, which used the original probes.

\begin{table}[h]
\caption{Qwen3-14B, complete per-dataset results. TriviaQA is in-domain. The other three are
held-out test datasets under distribution shift. Bold marks the best value per column within a
dataset.}
\label{tab:d1}
\centering\small
\begin{tabular}{llcccc}
\toprule
Dataset & Method & Acc $\uparrow$ & ECE $\downarrow$ & AUROC $\uparrow$ & Brier $\downarrow$\\
\midrule
\multirow{10}{*}{TriviaQA}
 & Verbal          & \ci{.729}{.024}$^{*}$ & \ci{.178}{.020} & \ci{.825}{.017} & \ci{.184}{.017}\\
 & Frozen Probe$^{*}$ & ---                 & \ci{.014}{.008} & \bci{.948}{.010} & \bci{.080}{.009}\\
 & Verbal-Iso      & ---                   & \bci{.013}{.009} & \ci{.825}{.017} & \ci{.128}{.010}\\
 & Verbal-TS+bias  & ---                   & \ci{.047}{.011} & \ci{.825}{.017} & \ci{.131}{.010}\\
 & Answer Prob.    & ---                   & \ci{.251}{.024} & \ci{.636}{.029} & \ci{.253}{.023}\\
 & Surface-Feat    & ---                   & \ci{.042}{.010} & \ci{.858}{.016} & \ci{.128}{.010}\\
 & EA-SD           & \bci{.750}{.024}      & \ci{.093}{.011} & \ci{.878}{.014} & \ci{.119}{.011}\\
 & SC-SD           & \ci{.748}{.023}       & \ci{.063}{.013} & \ci{.876}{.014} & \ci{.116}{.010}\\
 & Probe-SD-LRM    & \ci{.729}{.024}       & \ci{.031}{.011} & \ci{.926}{.011} & \ci{.096}{.009}\\
 & \probesd{}      & \ci{.749}{.024}       & \ci{.024}{.009} & \bci{.928}{.011} & \bci{.092}{.009}\\
\midrule
\multirow{10}{*}{EntityQ}
 & Verbal          & \ci{.267}{.024}$^{*}$ & \ci{.529}{.022} & \ci{.812}{.017} & \ci{.455}{.018}\\
 & Frozen Probe$^{*}$ & ---                 & \bci{.022}{.012} & \bci{.898}{.016} & \bci{.109}{.011}\\
 & Verbal-Iso      & ---                   & \ci{.267}{.019} & \ci{.812}{.017} & \ci{.229}{.011}\\
 & Verbal-TS+bias  & ---                   & \ci{.285}{.019} & \ci{.812}{.017} & \ci{.238}{.011}\\
 & Answer Prob.    & ---                   & \ci{.689}{.026} & \ci{.724}{.027} & \ci{.680}{.026}\\
 & Surface-Feat    & ---                   & \ci{.340}{.021} & \ci{.777}{.020} & \ci{.294}{.012}\\
 & EA-SD           & \ci{.288}{.025}       & \ci{.215}{.013} & \ci{.842}{.016} & \ci{.211}{.012}\\
 & SC-SD           & \bci{.296}{.025}      & \ci{.224}{.016} & \ci{.847}{.015} & \ci{.212}{.011}\\
 & Probe-SD-LRM    & \ci{.268}{.024}       & \ci{.102}{.014} & \ci{.868}{.016} & \ci{.136}{.010}\\
 & \probesd{}      & \ci{.291}{.025}       & \bci{.090}{.013} & \bci{.881}{.015} & \bci{.130}{.010}\\
\midrule
\multirow{10}{*}{NQ-open}
 & Verbal          & \ci{.440}{.027}$^{*}$ & \ci{.436}{.029} & \ci{.705}{.021} & \ci{.416}{.025}\\
 & Frozen Probe$^{*}$ & ---                 & \bci{.074}{.018} & \bci{.817}{.021} & \bci{.181}{.013}\\
 & Verbal-Iso      & ---                   & \ci{.245}{.027} & \ci{.705}{.021} & \ci{.276}{.017}\\
 & Verbal-TS+bias  & ---                   & \ci{.250}{.027} & \ci{.705}{.021} & \ci{.275}{.016}\\
 & Answer Prob.    & ---                   & \ci{.500}{.027} & \ci{.557}{.029} & \ci{.498}{.025}\\
 & Surface-Feat    & ---                   & \ci{.251}{.026} & \ci{.707}{.024} & \ci{.286}{.016}\\
 & EA-SD           & \bci{.472}{.026}      & \ci{.273}{.019} & \ci{.751}{.019} & \ci{.283}{.017}\\
 & SC-SD           & \ci{.470}{.026}       & \ci{.251}{.020} & \ci{.748}{.018} & \ci{.273}{.015}\\
 & Probe-SD-LRM    & \ci{.440}{.026}       & \ci{.170}{.020} & \bci{.797}{.021} & \ci{.218}{.014}\\
 & \probesd{}      & \ci{.471}{.026}       & \bci{.137}{.018} & \ci{.796}{.020} & \bci{.207}{.013}\\
\midrule
\multirow{10}{*}{SimpleQA}
 & Verbal          & \ci{.066}{.012}$^{*}$ & \ci{.661}{.016} & \ci{.627}{.042} & \ci{.532}{.013}\\
 & Frozen Probe$^{*}$ & ---                 & \bci{.089}{.014} & \ci{.670}{.052} & \bci{.089}{.010}\\
 & Verbal-Iso      & ---                   & \ci{.310}{.016} & \ci{.626}{.043} & \ci{.203}{.010}\\
 & Verbal-TS+bias  & ---                   & \ci{.348}{.016} & \ci{.627}{.042} & \ci{.223}{.009}\\
 & Answer Prob.    & ---                   & \ci{.897}{.014} & \ci{.477}{.047} & \ci{.880}{.013}\\
 & Surface-Feat    & ---                   & \ci{.405}{.016} & \ci{.608}{.042} & \ci{.276}{.010}\\
 & EA-SD           & \ci{.074}{.012}       & \ci{.193}{.011} & \ci{.648}{.034} & \ci{.165}{.010}\\
 & SC-SD           & \bci{.076}{.012}      & \ci{.234}{.014} & \ci{.665}{.031} & \ci{.180}{.009}\\
 & Probe-SD-LRM    & \ci{.066}{.011}       & \ci{.134}{.011} & \ci{.658}{.042} & \ci{.108}{.009}\\
 & \probesd{}      & \bci{.076}{.012}      & \bci{.113}{.011} & \bci{.671}{.037} & \bci{.105}{.009}\\
\bottomrule
\end{tabular}
\end{table}

\begin{table}[h]
\caption{Ministral-3-8B, complete per-dataset results. Conventions as in Table~\ref{tab:d1}.}
\label{tab:d2}
\centering\small
\begin{tabular}{llcccc}
\toprule
Dataset & Method & Acc $\uparrow$ & ECE $\downarrow$ & AUROC $\uparrow$ & Brier $\downarrow$\\
\midrule
\multirow{10}{*}{TriviaQA}
 & Verbal          & \ci{.664}{.025}$^{*}$ & \ci{.247}{.024} & \ci{.669}{.015} & \ci{.263}{.021}\\
 & Frozen Probe$^{*}$ & ---                 & \bci{.019}{.008} & \bci{.915}{.013} & \bci{.106}{.009}\\
 & Verbal-Iso      & ---                   & \bci{.020}{.014} & \ci{.669}{.015} & \ci{.188}{.009}\\
 & Verbal-TS+bias  & ---                   & \ci{.067}{.016} & \ci{.669}{.015} & \ci{.193}{.008}\\
 & Answer Prob.    & ---                   & \ci{.302}{.025} & \ci{.569}{.027} & \ci{.306}{.025}\\
 & Surface-Feat    & ---                   & \ci{.042}{.010} & \ci{.770}{.016} & \ci{.174}{.009}\\
 & EA-SD           & \ci{.696}{.024}       & \ci{.123}{.015} & \ci{.826}{.014} & \ci{.160}{.012}\\
 & SC-SD           & \bci{.698}{.025}      & \ci{.052}{.007} & \ci{.809}{.015} & \ci{.153}{.010}\\
 & Probe-SD-LRM    & \ci{.644}{.026}       & \ci{.064}{.011} & \bci{.904}{.012} & \ci{.128}{.008}\\
 & \probesd{}      & \ci{.693}{.024}       & \ci{.029}{.010} & \ci{.887}{.014} & \bci{.123}{.009}\\
\midrule
\multirow{10}{*}{EntityQ}
 & Verbal          & \ci{.207}{.023}$^{*}$ & \ci{.627}{.022} & \ci{.734}{.016} & \ci{.564}{.019}\\
 & Frozen Probe$^{*}$ & ---                 & \bci{.027}{.012} & \bci{.901}{.017} & \bci{.099}{.010}\\
 & Verbal-Iso      & ---                   & \ci{.320}{.020} & \ci{.734}{.016} & \ci{.258}{.009}\\
 & Verbal-TS+bias  & ---                   & \ci{.340}{.021} & \ci{.734}{.016} & \ci{.271}{.009}\\
 & Answer Prob.    & ---                   & \ci{.761}{.025} & \ci{.562}{.028} & \ci{.755}{.024}\\
 & Surface-Feat    & ---                   & \ci{.331}{.019} & \ci{.790}{.019} & \ci{.267}{.010}\\
 & EA-SD           & \ci{.216}{.023}       & \ci{.219}{.014} & \ci{.836}{.017} & \ci{.207}{.012}\\
 & SC-SD           & \bci{.220}{.023}      & \ci{.150}{.014} & \ci{.821}{.018} & \ci{.174}{.010}\\
 & Probe-SD-LRM    & \ci{.197}{.022}       & \ci{.108}{.014} & \bci{.875}{.017} & \bci{.127}{.009}\\
 & \probesd{}      & \ci{.219}{.023}       & \bci{.105}{.014} & \ci{.872}{.017} & \ci{.131}{.009}\\
\midrule
\multirow{10}{*}{NQ-open}
 & Verbal          & \ci{.370}{.026}$^{*}$ & \ci{.513}{.026} & \ci{.643}{.018} & \ci{.490}{.022}\\
 & Frozen Probe$^{*}$ & ---                 & \ci{.114}{.019} & \ci{.805}{.021} & \bci{.195}{.013}\\
 & Verbal-Iso      & ---                   & \ci{.225}{.025} & \ci{.643}{.018} & \ci{.274}{.011}\\
 & Verbal-TS+bias  & ---                   & \ci{.236}{.025} & \ci{.643}{.018} & \ci{.277}{.010}\\
 & Answer Prob.    & ---                   & \ci{.589}{.027} & \ci{.542}{.023} & \ci{.587}{.026}\\
 & Surface-Feat    & ---                   & \ci{.253}{.025} & \ci{.687}{.021} & \ci{.283}{.011}\\
 & EA-SD           & \bci{.410}{.026}      & \ci{.280}{.019} & \ci{.757}{.017} & \ci{.286}{.014}\\
 & SC-SD           & \ci{.405}{.026}       & \ci{.174}{.017} & \ci{.732}{.017} & \ci{.237}{.010}\\
 & Probe-SD-LRM    & \ci{.366}{.025}       & \ci{.163}{.019} & \bci{.811}{.019} & \bci{.204}{.011}\\
 & \probesd{}      & \ci{.407}{.026}       & \bci{.161}{.019} & \ci{.803}{.019} & \ci{.211}{.012}\\
\midrule
\multirow{10}{*}{SimpleQA}
 & Verbal          & \ci{.061}{.011}$^{*}$ & \ci{.761}{.014} & \ci{.531}{.040} & \ci{.671}{.012}\\
 & Frozen Probe$^{*}$ & ---                 & \ci{.159}{.014} & \ci{.615}{.050} & \ci{.126}{.010}\\
 & Verbal-Iso      & ---                   & \ci{.403}{.014} & \ci{.531}{.040} & \ci{.268}{.008}\\
 & Verbal-TS+bias  & ---                   & \ci{.446}{.014} & \ci{.531}{.040} & \ci{.293}{.007}\\
 & Answer Prob.    & ---                   & \ci{.909}{.012} & \ci{.457}{.041} & \ci{.895}{.012}\\
 & Surface-Feat    & ---                   & \ci{.403}{.014} & \ci{.542}{.041} & \ci{.279}{.008}\\
 & EA-SD           & \ci{.070}{.011}       & \ci{.253}{.011} & \ci{.610}{.038} & \ci{.210}{.010}\\
 & SC-SD           & \ci{.070}{.012}       & \ci{.183}{.011} & \ci{.612}{.033} & \ci{.165}{.008}\\
 & Probe-SD-LRM    & \ci{.056}{.010}       & \ci{.156}{.010} & \ci{.647}{.042} & \bci{.111}{.008}\\
 & \probesd{}      & \bci{.071}{.012}      & \bci{.155}{.011} & \bci{.674}{.039} & \ci{.120}{.008}\\
\bottomrule
\end{tabular}
\end{table}

The frozen Probe provides a lower-ECE reference than the distilled student. On the three held-out
datasets, Qwen's frozen Probe has macro ECE .062 and AUROC .795, compared with .113 and .783 for
\probesd{}. Ministral has .100 and .774, compared with .140 and .783. The student therefore does not
fully match the teacher's calibration profile, although it remains substantially better calibrated than
the verbal baseline. The two families agree: \probesd{} is best or close to best on most dataset--metric cells, and SC-SD occasionally
overtakes it on accuracy (EntityQ, SimpleQA). The same ordering survives the alternative
missing-score conventions of Appendix~\ref{app:brier}. Per-dataset results for \probesdq{} and
ConfTuner are in Table~\ref{tab:newfull}.
\clearpage
\section{Training-Set Construction and Downsampling}
\label{app:downsample}

This appendix supports the data-downsampling step of Section~\ref{sec:exp-setup}. The
probe-coherent selection and downsampling pipeline is shared across the two models: 15{,}000
questions $\times$ 20 samples $=$ 300{,}000 candidates, of which the coherence filter keeps one
trajectory per question (Ministral 13{,}840, Qwen 14{,}951), which is then downsampled to 10{,}000
training trajectories.

\paragraph{Probe-score distribution before and after selection.} Table~\ref{tab:ds-dist} reports
the probe-score distribution of the candidate pool and the retained set. The pool statistics are
over a random sample of 15{,}000 scored candidates, so its size is directly comparable to the
retained set's one trajectory per question. The filter biases the retained set
toward mid-to-high confidence: on Ministral the retained probe mean rises from 0.636 to 0.672,
whereas on Qwen it is nearly unchanged (0.722 to 0.723).

\begin{table}[h]
\caption{Probe-score distribution of the candidate pool and the coherently selected set.}
\label{tab:ds-dist}
\centering\small
\begin{tabular}{llccccc}
\toprule
Model & Set & $n$ & mean & p25 & p50 & p75\\
\midrule
\multirow{2}{*}{Ministral-3-8B}
 & Pool (random 15k) & 15{,}000 & 0.636 & 0.317 & 0.777 & 0.943\\
 & Retained & 13{,}840 & 0.672 & 0.495 & 0.782 & 0.930\\
\multirow{2}{*}{Qwen3-14B}
 & Pool (random 15k) & 15{,}000 & 0.722 & 0.460 & 0.936 & 0.992\\
 & Retained & 14{,}951 & 0.723 & 0.485 & 0.915 & 0.989\\
\bottomrule
\end{tabular}
\end{table}

\paragraph{Why the shift is large on Ministral.} The filter examines candidates in order of
$|p_i-\bar p_q|$ and keeps the first one whose relabeled triple is judged coherent. Candidates
with multiple or inconsistent in-chain confidence numbers are rejected before the coherence check
(Section~\ref{sec:relabel}). Ministral's low-probe trajectories fail this stage disproportionately:
trajectories with probe below 0.2 pass at only $\sim$19\%, because Ministral's uncertain answers tend
to have long chains of thought (median 2{,}064 tokens for probe$<$0.2 versus 399 for
probe$\ge$0.8) in which the model re-evaluates and inserts several, often conflicting, confidence
numbers, and such trajectories are discarded by the multiple-score check. Format failures likewise concentrate in
long, low-probe trajectories. The average position of the first passing candidate is therefore much higher on
Ministral (3.11) than on Qwen (1.43), and the retained set shifts toward mid-to-high confidence.
This makes Ministral the harder case for testing whether the shift matters.

\paragraph{Downsampling variants.} To assess how the selection shift affects performance, we
downsample the retained trajectories to 10{,}000 under four policies: uniform random
(\emph{rand\_sub}), a quota by difficulty decile replicating the pool's difficulty mix
(\emph{diffdist\_rp}), the difficulty quota plus a correct/incorrect quota aligning the training
accuracy to the global accuracy (\emph{calibrated}), and the difficulty quota plus a within-decile
sliding window aligning each decile's probe mean (\emph{probe\_align}). Table~\ref{tab:ds-variants}
reports their 4-dataset macro calibration (3 seeds; $\Delta$pp is the largest difficulty-decile
deviation from the pool and Train Acc the aligned training accuracy), together with the Qwen
calibrated-versus-random comparison discussed below.

\begin{table}[h]
\caption{Downsampling policies across models, 4-dataset macro (3 seeds). $\Delta$pp is the largest
difficulty-decile deviation from the pool and Train Acc the aligned training accuracy, both from
the Ministral variant study.}
\label{tab:ds-variants}
\centering\small
\setlength{\tabcolsep}{4.5pt}
\begin{tabular}{llcccccc}
\toprule
Model & Downsampling & ECE $\downarrow$ & AUROC $\uparrow$ & Brier $\downarrow$ & Sample Acc & $\Delta$pp & Train Acc\\
\midrule
\multirow{2}{*}{Qwen3-14B}
 & calibrated (\probesd{}) & 0.091 & 0.819 & 0.134 & 0.397 & --- & ---\\
 & random (\emph{rand\_sub}) & 0.094 & 0.820 & 0.134 & 0.394 & --- & ---\\
\multirow{4}{*}{Ministral-3-8B}
 & random        & 0.123 & 0.808 & 0.150 & 0.346 & 4.26 & 70.0\%\\
 & diffdist\_rp  & 0.112 & 0.813 & 0.144 & 0.347 & 0.01 & 65.5\%\\
 & calibrated    & \textbf{0.105} & \textbf{0.815} & \textbf{0.142} & 0.348 & 1.17 & 61.8\%\\
 & probe\_align  & 0.118 & 0.807 & 0.149 & 0.348 & 0.01 & 66.2\%\\
\bottomrule
\end{tabular}
\end{table}

\paragraph{Robustness to the selection shift.} The alignment matters only where the filter induces
a large distribution shift. On Qwen, whose selection barely changes the probe distribution, random
subsampling (\emph{rand\_sub}) gives macro ECE/AUROC/Brier of 0.094/0.820/0.134, essentially
indistinguishable from the calibrated \probesd{} (0.091/0.819/0.134). On Ministral, whose shift is
large, calibrated improves ECE from 0.123 to 0.105 (Table~\ref{tab:ds-variants}). The gain from
alignment is thus modest and confined to the large-shift case, and the method is overall robust to
the coherence filter's distribution shift. Sample Acc is essentially unchanged in either model.

\clearpage

\clearpage
\section{Supporting Data for the Mechanistic Analysis}
\label{app:mechanism}

This appendix reports the teacher-to-student fidelity and label-source results underlying
\S\ref{sec:fidelity}. All experiments use Qwen3-14B with at most 10 samples per
question. \probesd{}, SC-SD and Verbal-SD share the same source sampling, filtering pipeline and
Base initialization, and differ only in the confidence target: the frozen probe probability, the
self-consistency answer-cluster frequency, and the source model's own verbalized score. Question-
level metrics average over the samples of a question first. Trace-level metrics are computed per
trace. Probe is the L24 residual, $C{=}0.005$, three-seed ensemble used to generate training
labels. Correlations and intervals use question-level bootstrap with $B{=}5000$.

\begin{table}[h]
\caption{Per-decile values behind Figure~\ref{fig:fidelity}(a): each source's mean, aggregated by
decile of the source verbalized score $q_{src}$ on TriviaQA ($n{=}100$ per decile). Deciles are
equal-frequency in $q_{src}$, so that column is monotone by construction.}
\label{tab:h1}
\centering\small
\begin{tabular}{ccccccc}
\toprule
Decile & $q_{src}$ & Probe & Observed acc & \probesd{} & SC-SD & Verbal-SD\\
\midrule
1  & .675 & .191 & .141 & .295 & .374 & .756\\
2  & .803 & .337 & .331 & .424 & .522 & .817\\
3  & .866 & .461 & .423 & .537 & .637 & .867\\
4  & .907 & .601 & .546 & .645 & .752 & .906\\
5  & .938 & .836 & .891 & .865 & .878 & .936\\
6  & .952 & .900 & .916 & .904 & .926 & .945\\
7  & .959 & .935 & .940 & .957 & .940 & .950\\
8  & .966 & .953 & .977 & .967 & .967 & .955\\
9  & .975 & .977 & .985 & .984 & .980 & .960\\
10 & .989 & .986 & .989 & .992 & .993 & .972\\
\bottomrule
\end{tabular}
\end{table}

The four lowest deciles are the region where the source is overconfident: in decile 1 the source
reports .675 against an observed accuracy of .141, Probe reports .191, \probesd{} follows down to
.295, and Verbal-SD still reports .756. In the high deciles all curves converge, because the two
teachers already agree there.

\begin{table}[h]
\caption{Per-quartile values behind Figure~\ref{fig:fidelity}(b): mean absolute error of each
student relative to Probe, by quartile of the overconfidence gap $q_{src}-p_{\mathrm{probe}}$ on
TriviaQA ($n{=}250$ per quartile). The last column is $\lvert
q_{src}-p_{\mathrm{probe}}\rvert$, the distance between the verbalized score and Probe with no
distillation at all.}
\label{tab:h2}
\centering\small
\begin{tabular}{clcccc}
\toprule
Quartile & Gap range & \probesd{} & SC-SD & Verbal-SD & $q_{src}$ itself\\
\midrule
Q0 & $[-.560,-.011)$ & .018 & .034 & .037 & .030\\
Q1 & $[-.011,+.044)$ & .025 & .030 & .019 & .013\\
Q2 & $[+.044,+.373)$ & .120 & .134 & .202 & .188\\
Q3 & $[+.373,+.832]$ & \textbf{.164} & .263 & .588 & .575\\
\bottomrule
\end{tabular}
\end{table}

Verbal-SD nearly coincides with the no-distillation reference at every quartile (.588 versus .575
in Q3), i.e.\ it simply reproduces the original verbalized score, whereas \probesd{} stays at .164
in the quartile of largest disagreement. The three students differ mainly in Q2--Q3 and are almost
identical in Q0.

\begin{table}[h]
\caption{Question-level Pearson correlation between each student's verbalized score and the two
candidate teachers, Probe and the source verbalized score $q_{src}$. Bold marks, for each
student, the teacher it correlates with more strongly.}
\label{tab:h3}
\centering\small
\setlength{\tabcolsep}{4pt}
\begin{tabular}{lcccccc}
\toprule
& \multicolumn{2}{c}{\probesd{}} & \multicolumn{2}{c}{SC-SD} & \multicolumn{2}{c}{Verbal-SD}\\
\cmidrule(lr){2-3}\cmidrule(lr){4-5}\cmidrule(lr){6-7}
Dataset & Probe & $q_{src}$ & Probe & $q_{src}$ & Probe & $q_{src}$\\
\midrule
TriviaQA & \textbf{.913} & .771 & \textbf{.872} & .791 & .767 & \textbf{.822}\\
EntityQ  & \textbf{.895} & .676 & \textbf{.810} & .738 & .658 & \textbf{.803}\\
NQ-open  & \textbf{.823} & .550 & \textbf{.700} & .590 & .526 & \textbf{.725}\\
SimpleQA & \textbf{.762} & .494 & \textbf{.642} & .641 & .468 & \textbf{.757}\\
\bottomrule
\end{tabular}
\end{table}

All four datasets agree in direction: \probesd{} always correlates more strongly with Probe than
with the source verbalized score, Verbal-SD does the reverse, and SC-SD is intermediate. The
confidence structure a student inherits is determined by its training label, and this holds
outside the training domain as well.

\begin{figure}[h]
\centering
\includegraphics[width=0.98\linewidth]{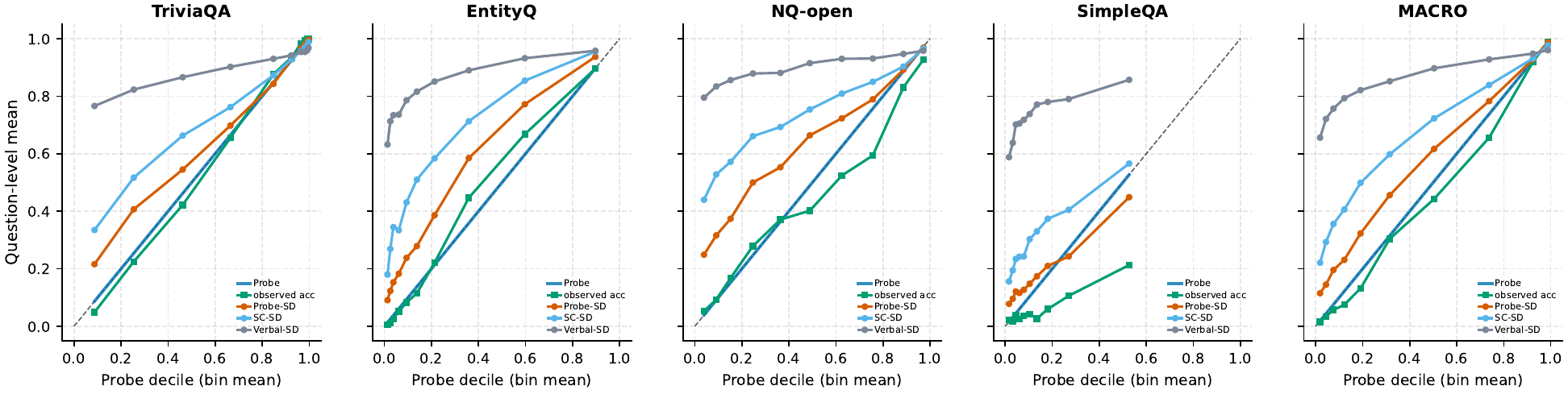}
\caption{Probe, observed accuracy and the three students' verbalized confidence, conditioned on
Probe decile rather than on $q_{src}$. Single datasets have $n{=}100$ per decile, MACRO $n{=}400$.}
\label{fig:shape}
\end{figure}

Binning by Probe rather than by $q_{src}$ (Figure~\ref{fig:shape}) makes the shape argument
explicit: \probesd{}'s output follows Probe's difficulty profile decile by decile, whereas
Verbal-SD is a nearly flat high-confidence line almost independent of question difficulty---on
EntityQ it rises only from .632 in the lowest decile to .958 in the highest, while Probe rises
from .013 to .897 over the same range. If \probesd{} were merely applying a global downward shift
to the verbalized score, its curve would run parallel to Verbal-SD. The measured difference in
shape rules that explanation out.

\begin{figure}[h]
\centering
\includegraphics[width=0.52\linewidth]{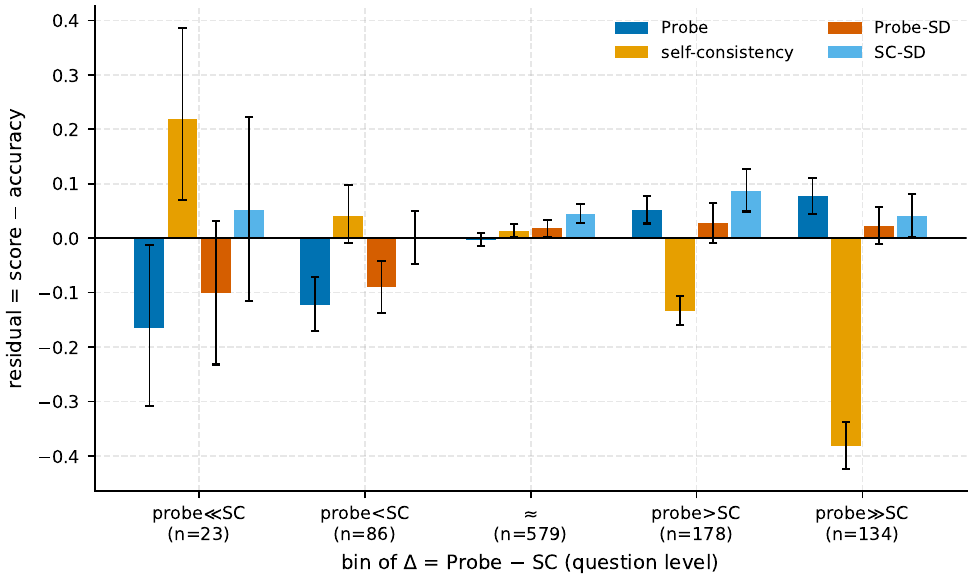}
\caption{Calibration residuals on TriviaQA, binned by the disagreement between Probe and
self-consistency (\S\ref{sec:label-source}). Positive values indicate overconfidence, negative
values underconfidence. Per-bin values are in Table~\ref{tab:h4}.}
\label{fig:labelsource}
\end{figure}

\begin{table}[h]
\caption{Per-bin values behind Figure~\ref{fig:labelsource}: calibration residuals on TriviaQA
($N{=}1{,}000$ questions), binned by the question-level disagreement
$\Delta=p_{\mathrm{probe}}-f_{\mathrm{SC}}$. Residual $=$ score $-$ corresponding accuracy, so
positive values are overconfidence. Probe and SC are referenced to the source accuracy and each
student to its own. Brackets give 95\% question-level bootstrap intervals ($B{=}5000$).}
\label{tab:h4}
\centering\small
\setlength{\tabcolsep}{3pt}
\begin{tabular}{lcccccc}
\toprule
Bin & $n$ & $\mathrm{acc}_{src}$ & Probe & SC & \probesd{} & SC-SD\\
\midrule
Probe$\ll$SC & 23  & .635 & \makecell{$-.165$\\\mbox{[$-.309,-.012$]}} & \makecell{$+.219$\\\mbox{[$.070,.386$]}} & \makecell{$-.100$\\\mbox{[$-.232,.032$]}} & \makecell{$+.051$\\\mbox{[$-.116,.222$]}}\\
Probe$<$SC   & 86  & .700 & \makecell{$-.123$\\\mbox{[$-.171,-.071$]}} & \makecell{$+.041$\\\mbox{[$-.009,.098$]}} & \makecell{$-.090$\\\mbox{[$-.138,-.042$]}} & \makecell{$-.000$\\\mbox{[$-.047,.050$]}}\\
$\approx$    & 579 & .739 & \makecell{$-.003$\\\mbox{[$-.014,.009$]}}  & \makecell{$+.013$\\\mbox{[$.002,.026$]}}  & \makecell{$+.018$\\\mbox{[$.003,.033$]}}  & \makecell{$+.045$\\\mbox{[$.028,.062$]}}\\
Probe$>$SC   & 178 & .610 & \makecell{$+.052$\\\mbox{[$.027,.078$]}}   & \makecell{$-.133$\\\mbox{[$-.160,-.106$]}} & \makecell{$+.028$\\\mbox{[$-.008,.064$]}} & \makecell{$+.087$\\\mbox{[$.049,.127$]}}\\
Probe$\gg$SC & 134 & .766 & \makecell{$+.078$\\\mbox{[$.044,.111$]}}   & \makecell{$-.381$\\\mbox{[$-.424,-.337$]}} & \makecell{$+.023$\\\mbox{[$-.011,.058$]}} & \makecell{$+.041$\\\mbox{[$.003,.082$]}}\\
\bottomrule
\end{tabular}
\end{table}

The middle bin holds 57.9\% of the questions. Both teachers are close to calibrated there and the
students differ least, so the ECE gaps of Table~\ref{tab:main} can be attributed bin by bin to the
two disagreement tails. SC-SD fails to reproduce its teacher's direction at the Probe$\gg$SC end
(SC is $-.381$, SC-SD is $+.041$). The same mismatch shows up as a distance to its own label: in
the top quintile of $\lvert\mathrm{Probe}-\mathrm{SC}\rvert$, SC-SD's MAE to the SC label is .415
while its MAE to Probe is only .126. Answer frequency is too noisy a label in that region for the
student to fit, which places a ceiling on the fidelity of SC distillation.

\begin{table}[h]
\caption{Mean absolute error of each student relative to Probe, by quintile of
$\lvert\mathrm{Probe}-\mathrm{SC}\rvert$ on TriviaQA ($n{=}200$ per quintile). This binning is by
the magnitude of teacher disagreement and is not the same convention as
Table~\ref{tab:h2}.}
\label{tab:h5}
\centering\small
\begin{tabular}{clcccc}
\toprule
Quintile & $\lvert\mathrm{Probe}-\mathrm{SC}\rvert$ & mean $\Delta$ & \probesd{} & SC-SD & Verbal-SD\\
\midrule
Q0 & $[.000,.013)$ & $-.004$ & \textbf{.030} & .055 & .115\\
Q1 & $[.013,.044)$ & $-.018$ & \textbf{.053} & .079 & .173\\
Q2 & $[.044,.107)$ & $-.023$ & \textbf{.123} & .171 & .327\\
Q3 & $[.107,.242)$ & $+.060$ & \textbf{.103} & .146 & .263\\
Q4 & $[.242,.855]$ & $+.295$ & \textbf{.100} & .126 & .179\\
\bottomrule
\end{tabular}
\end{table}

\probesd{}'s MAE to Probe stays in .030--.123 and is the lowest in every disagreement quintile, so
the fidelity difference is not driven by the low-disagreement samples alone.
\clearpage
\section{The Source of the Base-Initialization Accuracy Gain}
\label{app:priming}

\S\ref{sec:ablation} ruled out probe relabeling as the source of the small accuracy gain, which
leaves the initialization. Factual priming explains it
\citep{gekhman2026thinkingrecallreasoningunlocks}: a reasoning model gains on factual QA less by
decomposing the task than by \emph{retrieving into its own context}---stating relevant facts in the
chain of thought lowers the threshold for the answer that follows. Correctness then turns on two
things, whether the key fact surfaces and whether it is used.

We measure the first with the \emph{gold recall hit rate} $r$: a trace counts as a hit when the gold
answer or an alias appears in the chain of thought, by string match, with no LLM in the loop. With
$a_1$ and $a_0$ the accuracies conditional on hit and on miss,
\begin{equation}
\mathrm{acc}=r\,a_1+(1-r)\,a_0 ,
\end{equation}
so a difference between checkpoints splits into a retrieval term ($\Delta r$) and a usage term
($\Delta a_1$). A separate audit labels every extracted fact for truth and for whether it bridges to
the answer, and \emph{retrieval difficulty} counts hits over a question's 10 samples: 10 is
\emph{easy}, 1--9 \emph{medium}, 0 \emph{hard}.

\textbf{Qwen3-14B: retrieval accounts for nearly everything.} SFT-Base beats Instruct by 2.36
points (.7508 versus .7272). Gold recall explains 2.21 of them, 93.7\%, while the usage term
contributes 0.8\%: accuracy after a hit is identical (.8915 versus .8912) and the two checkpoints
produce equally many true bridging facts (4.95 versus 4.98). The gap is about getting a stored fact
into context, not about reasoning over it.

\textbf{Ministral-3-8B: retrieval dominates but does not exhaust.} SFT-Base leads by 3.56 points
(.6995 versus .6639), of which retrieval supplies 2.60 (73.0\%). Here accuracy after a hit still
differs (.8889 versus .8791), the usage term takes another 19.9\%, and Reasoning produces fewer true
bridging facts (4.32 versus 4.81) while hallucinating slightly more (27.6\% versus 26.7\%)---the
direction \citet{gekhman2026thinkingrecallreasoningunlocks} report for hallucinated facts inflating
final error.

Difficulty transitions agree across families: exactly one question falls from \emph{easy} to
\emph{hard} in each, so finetuning erases nothing firmly encoded, while the middle band moves up
(net $+42$ questions for Qwen, $+52$ for Ministral). Post-training reduces the retrievability of
marginal facts, not knowledge. Gold recall is a string metric and misses paraphrases outside the
alias list, so its level is an underestimate. Both checkpoints are measured the same way, so
$\Delta r$ survives the bias, and the fact audit, which uses no string matching, points the same
way.

The rest of this appendix gives the supporting data. We compare two pairs---Qwen3-14B's SFT-Base
student against Instruct, and Ministral-3-8B's SFT-Base student against Reasoning---on the
1{,}000 questions of the TriviaQA validation split, with 10 symmetric samples per question and
answers graded by Qwen3-32B. The Ministral student here was trained with the earlier label probe
(Appendix~\ref{app:grid}). Its accuracy matches that of the students in Table~\ref{tab:main} to within
0.001 on every benchmark, so the retrieval analysis carries over. The chain-of-thought
quality audit uses glm-5-2-260617 \citep{glm5team2026glm5vibecodingagentic} to extract facts from 200 traces (100 matched pairs) and label
\emph{truth} and \emph{relevance}. Gold recall only checks whether the gold answer or one of its
aliases occurs as a substring of the chain of thought and does not depend on any LLM judgment.

\begin{table}[h]
\caption{Sample-level accuracy difference and its decomposition into a retrieval term, a usage
(conditional accuracy) term, and a miss-branch term, for both families.}
\label{tab:g1}
\centering\small
\begin{tabular}{lcc}
\toprule
& Qwen3-14B: SFT vs Instruct & Ministral-3-8B: SFT vs Reasoning\\
\midrule
Sample accuracy      & .7508 vs .7272 ($+.0236$) & .6995 vs .6639 ($+.0356$)\\
Retrieval term       & \textbf{$+.0221$ (93.7\%)} & \textbf{$+.0260$ (73.0\%)}\\
Usage term           & $+.0002$ (0.8\%) & $+.0071$ (19.9\%)\\
Miss-branch term     & $+.0013$ (5.5\%) & $+.0025$ (7.0\%)\\
\bottomrule
\end{tabular}
\end{table}

Writing $\mathrm{acc}=r\,a_1+(1-r)\,a_0$, the total difference is first split into a hit branch
$\Delta(r a_1)$ and a miss branch $\Delta[(1-r)a_0]$, and the hit branch is then split into a
retrieval term $\Delta r\cdot a_1^{(F2)}$ and a usage term $r^{(F1)}\cdot\Delta a_1$. Both
families are dominated by the retrieval term, but the residual differs: for Qwen the usage term
contributes only 0.8\%, whereas for Ministral it accounts for a further 19.9\%, so the Ministral
mechanism cannot be reduced to a pure retrieval effect.

\begin{table}[h]
\caption{Chain-of-thought quality audit, 200 traces per family. Facts are counted per trace.}
\label{tab:g2}
\centering\small
\begin{tabular}{lcccc}
\toprule
& \multicolumn{2}{c}{Qwen3-14B} & \multicolumn{2}{c}{Ministral-3-8B}\\
\cmidrule(lr){2-3}\cmidrule(lr){4-5}
Metric & SFT-Base & Instruct & SFT-Base & Reasoning\\
\midrule
Facts per trace                & 9.24 & 9.02 & 9.33 & 9.29\\
truth: true                    & 80.1\% & 78.0\% & 69.7\% & 67.8\%\\
truth: false (hallucination)   & 13.5\% & 15.2\% & 26.7\% & 27.6\%\\
relevance: bridge              & 5.83 & 5.97 & 6.13 & 5.57\\
true $\wedge$ bridge           & 4.95 & 4.98 & 4.81 & 4.32\\
\bottomrule
\end{tabular}
\end{table}

On Qwen the two checkpoints are tied on effective bridges: true$\wedge$bridge counts are 4.95 and
4.98, and the 2.1-point advantage of SFT in true-fact rate comes from context or tangential facts
that do not lead directly to the answer, while Instruct produces slightly more bridge facts
overall without producing more \emph{true} ones. There is thus no evidence on Qwen that better
fact quality explains the accuracy gap. Ministral behaves differently: its SFT student is higher
on true-fact rate, bridge coverage and true$\wedge$bridge alike. Conditioning on correctness,
true$\wedge$bridge is 5.76 versus 5.43 on correct traces and 2.12 versus 2.06 on incorrect ones,
and on incorrect traces Reasoning has the \emph{higher} true-fact rate (52.0\% versus 44.4\%),
indicating that part of its errors occur in the reasoning or synthesis step after a correct fact
has already surfaced.

\begin{table}[h]
\caption{Gold recall hit rate and conditional accuracy.}
\label{tab:g3}
\centering\small
\setlength{\tabcolsep}{4pt}
\begin{tabular}{lcccccc}
\toprule
& \multicolumn{3}{c}{Qwen3-14B} & \multicolumn{3}{c}{Ministral-3-8B}\\
\cmidrule(lr){2-4}\cmidrule(lr){5-7}
Metric & SFT-Base & Instruct & $\Delta$ & SFT-Base & Reasoning & $\Delta$\\
\midrule
Gold recall hit rate $r$          & \textbf{.7888} & .7640 & $+2.48$pp & \textbf{.7294} & .6998 & $+2.96$pp\\
Accuracy $\mid$ gold recalled     & .8915 & .8912 & $+0.03$pp & \textbf{.8889} & .8791 & $+0.98$pp\\
Accuracy $\mid$ gold not recalled & .2254 & .1962 & $+2.92$pp & .1888 & .1622 & $+2.66$pp\\
Sample accuracy                   & .7508 & .7272 & $+2.36$pp & .6995 & .6639 & $+3.56$pp\\
\bottomrule
\end{tabular}
\end{table}

Once the gold fact has surfaced, Qwen's two checkpoints are indistinguishable (.8915 versus
.8912). At the question level, recall on at least one of the 10 samples is .8660 versus .8530 for
Qwen and .8370 versus .8080 for Ministral. Qwen's SFT student uniquely recalls 23 questions
against Instruct's 10, with 843 recalled by both and 124 by neither, and Ministral's SFT student
uniquely recalls 51 against 22, with 786 recalled by both and 141 by neither.

\begin{table}[h]
\caption{Retrieval difficulty over 1{,}000 questions (left) and the difficulty transition matrix
across the two checkpoints of the same family (right). Difficulty is defined by the number of
gold-recall hits over the 10 samples of a question: \emph{easy} = all 10, \emph{medium} = 1--9,
\emph{hard} = never.}
\label{tab:g4}
\centering\small
\setlength{\tabcolsep}{4pt}
\begin{tabular}{lcccc@{\hskip 16pt}lccc}
\toprule
& \multicolumn{2}{c}{Qwen3-14B} & \multicolumn{2}{c}{Ministral-3-8B} & & \multicolumn{3}{c}{$\to$ SFT-Base}\\
\cmidrule(lr){2-3}\cmidrule(lr){4-5}\cmidrule(lr){7-9}
Difficulty & SFT & Instr. & SFT & Reas. & Source side & easy & med. & hard\\
\midrule
easy   & 687 & 663 & 582 & 563 & \emph{Qwen} Instruct easy (663)  & 635 & 27  & \textbf{1}\\
medium & 179 & 190 & 255 & 245 & \phantom{\emph{Qwen}} medium (190) & \textbf{51} & 130 & 9\\
hard   & 134 & 147 & 163 & 192 & \phantom{\emph{Qwen}} hard (147)   & 1   & 22  & 124\\
\addlinespace[2pt]
       &     &     &     &     & \emph{Min.} Reasoning easy (563) & 506 & 56  & \textbf{1}\\
       &     &     &     &     & \phantom{\emph{Min.}} medium (245) & \textbf{73} & 151 & 21\\
       &     &     &     &     & \phantom{\emph{Min.}} hard (192)   & 3   & 48  & 141\\
\bottomrule
\end{tabular}
\end{table}

Firmly encoded knowledge is almost never lost: exactly one question moves from easy to hard in
each family. The net flow is concentrated in the middle band---Qwen moves 51 medium questions to
easy against 9 to hard (net $+42$) and Ministral 73 against 21 (net $+52$)---and the hard band
largely persists, with an overlap of 124/147 (84\%) for Qwen and 141/192 (73\%) for Ministral.
Ministral has both a larger hard band and a larger difference between checkpoints, consistent with
more marginal facts changing retrievability under a change of training regime at the smaller
scale. Taken together, both families show post-training acting mainly on the retrieval difficulty
of marginal facts rather than systematically erasing firm knowledge: Qwen's $+2.36$pp is almost
entirely gold recall with identical post-recall behavior, whereas Ministral's $+3.56$pp is still
retrieval-dominated but includes fact quality and conditional reasoning as well. Gold surfacing is
an auditable string metric but can miss semantically equivalent phrasings outside the alias list,
which is why we read it jointly with the fact-quality audit, the conditional accuracies, and the
question-level transition matrix rather than treating any single metric as complete mechanistic
evidence.
\section{Test-Time Behavior of the Internalized Confidence}
\label{sec:tts}

We adapt two diagnostics that \citet{damani2025binaryrewardstraininglms} apply to
confidence-reward models and ask whether analogous behaviors can arise from SFT alone. We train no
RL baseline and our traces carry no separate \texttt{<analysis>} segment, so these diagnostics only
approximate the original protocol. They test whether analogous behaviors can arise without online
RL in our setting. They do not constitute a controlled comparison with RLCR.\footnote{RLCR (Qwen2.5-7B
Base, trained on HotPotQA, so TriviaQA and SimpleQA are both OOD for it) reports TriviaQA Acc
60.8\% / ECE 0.06 / AUROC 0.73 and SimpleQA Acc 12.1\% / ECE 0.34 / AUROC 0.60. Our \probesd{} on
Qwen3-14B reaches TriviaQA Acc 74.9\% / ECE 0.024 / AUROC 0.928 and SimpleQA Acc 7.6\% / ECE 0.113
/ AUROC 0.671. Base model, training distribution, SimpleQA subset (RLCR uses all 4{,}326 questions,
we use the 1{,}000-question Verified set) and grading protocol all differ, so these are
order-of-magnitude reference points, not a controlled comparison.}

\begin{figure}[h]
\centering
\includegraphics[width=0.94\linewidth]{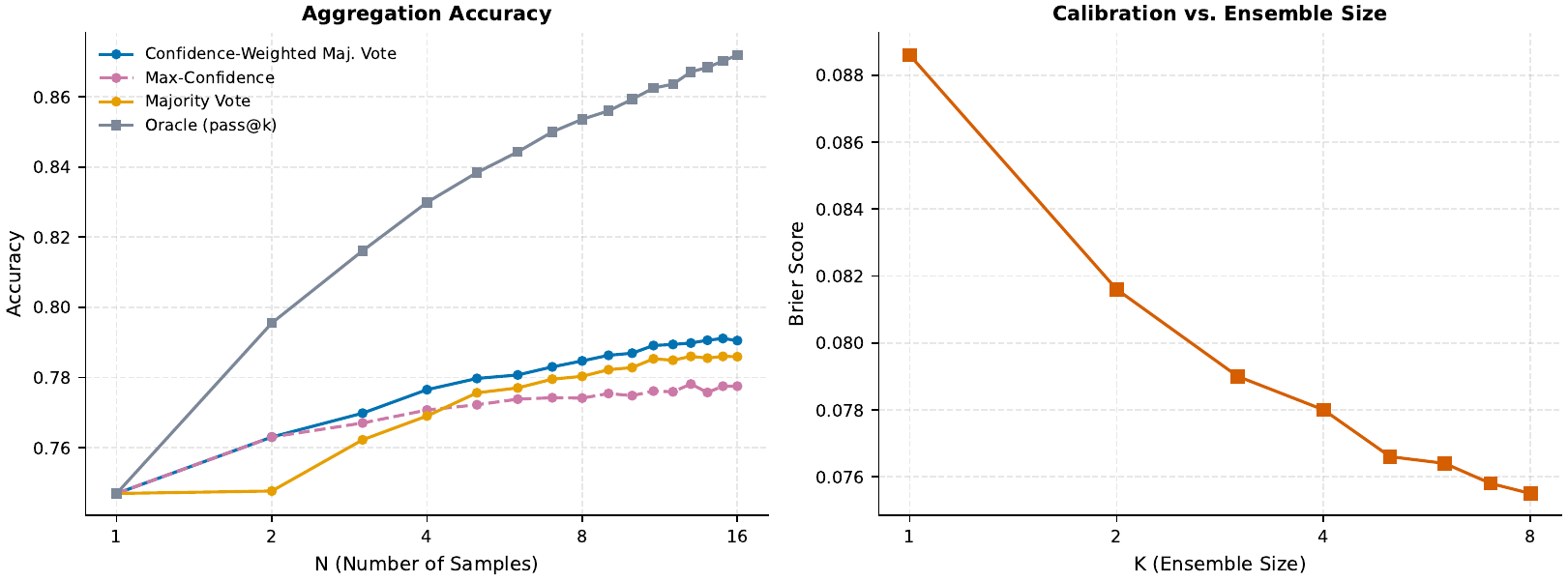}
\caption{(a) Majority voting, max-confidence selection, confidence-weighted voting, and the oracle
on TriviaQA as a function of the number of samples $N$. (b) Brier score of the confidence averaged
over $K$ traces of the same question that share an answer.}
\label{fig:tts}
\end{figure}

Confidence works better as a weight than as a target. On TriviaQA at $N{=}16$, majority voting,
max-confidence selection, confidence-weighted voting and the oracle reach 78.59\%, 77.75\%,
79.05\% and 87.19\%, i.e.\ $+3.90$, $+3.06$, $+4.36$ and $+12.50$ points over $N{=}1$
(Figure~\ref{fig:tts}a). Weighted voting leads plain vote counting. Picking the single most
confident sample does not, and everything saturates past $N\ge8$---the pattern
\S\ref{sec:not-selection} predicts, since confidence aggregates candidates and cannot invent ones
the pool never produced.

Averaging the score over $K$ traces that agree on an answer, across a fixed set of 900 questions,
lowers Brier from 0.0886 at $K{=}1$ to 0.0755 at $K{=}8$, near the group-mean bound of 0.0748,
while ECE holds at 0.033--0.036 (Figure~\ref{fig:tts}b). Averaging removes variance, not
within-bin bias. These traces were sampled independently and merely happened to agree, rather than
being resampled with the answer fixed, so the check only approximates the original.

\begin{figure}[h]
\centering
\includegraphics[width=0.94\linewidth]{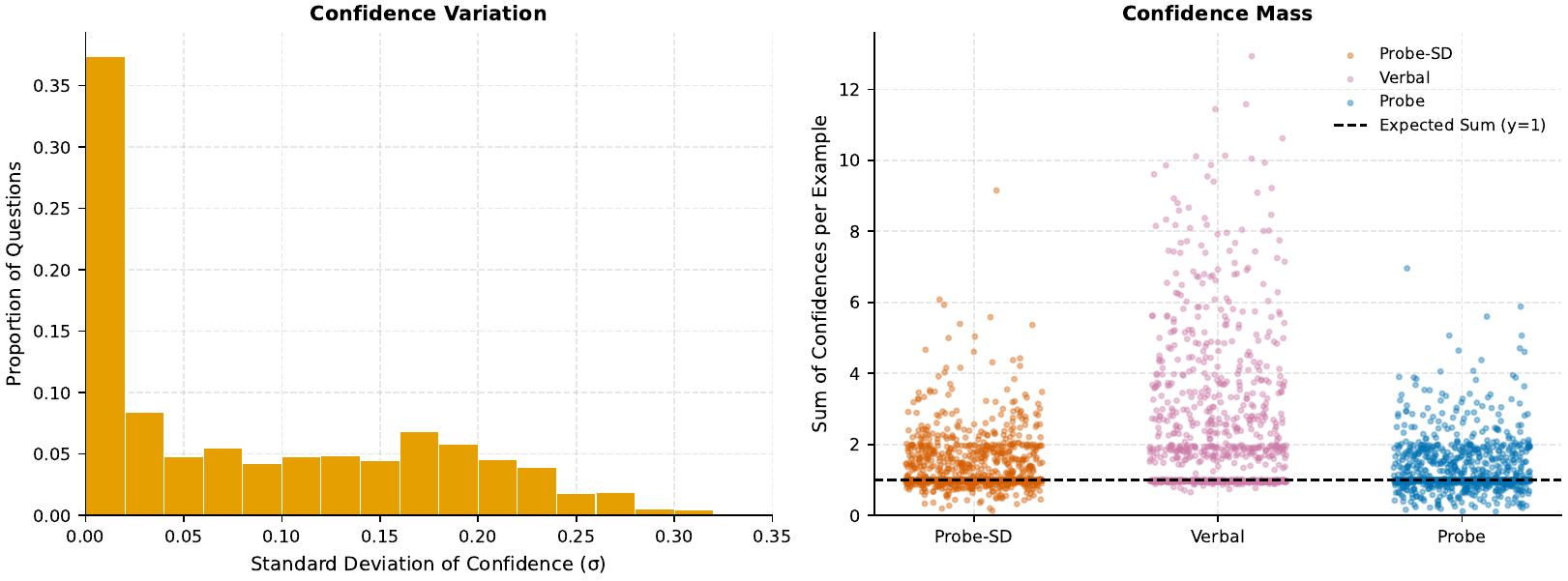}
\caption{Self-consistency of the internalized confidence (\S\ref{sec:tts}). (a) Distribution of the
per-question standard deviation of verbalized confidence over $K{=}16$ repeated samples of the same
answer on TriviaQA. (b) Sum of the mean confidences assigned to mutually exclusive answers of the
same question, for the original model and \probesd{}.}\label{fig:selfcons}
\end{figure}

Two consistency checks close the analysis (Figure~\ref{fig:selfcons}). Over 949 questions at
$K{=}16$, 37.4\% of scores vary by $\sigma<0.02$ and 13.1\% form a real long tail at
$\sigma\ge0.20$. Asked about mutually exclusive answers, the original Instruct model assigns
confidences that sum to 2.287 on average and exceed 2 on 33.3\% of questions, whereas \probesd{}
sums to 1.383 on average and exceeds 2 on 13.3\% of questions. Low variance alone proves
nothing---saturated overconfidence is
equally consistent---but together with the low ECE of \S\ref{sec:experiments} it rules that case
out. These behaviors arise in our setting without online RL. We do not claim equivalence to an
RL-trained model.

\clearpage
\section{Additional Experiments}
\label{app:extra}

This appendix reports additional experiments: within-question ranking (\ref{app:within}),
question-level labels (\ref{app:qmean}), the source of the probe label's advantage (\ref{app:mech}),
the ConfTuner and RL baselines (\ref{app:conftuner}--\ref{app:rl}), a nonlinear probe
(\ref{app:mlp}), zero-shot transfer to MATH-500 (\ref{app:math}), the questions on which the probe
ordering helps (\ref{app:bins}) and multiple-choice QA with shuffled options (\ref{app:mcqa}). Unless noted, settings follow
\S\ref{sec:exp-setup}, and $\pm$ values are 95\% question-level bootstrap half-widths.

\subsection{Within-Question AUROC}
\label{app:within}

For every question whose 10 samples ($T{=}0.6$, original model) contain both correct and incorrect
answers, we compute the AUROC of a score at ranking that question's correct samples above its
incorrect ones, and pool over such questions. We compare the probe, the answer-cluster frequency and
the verbalized score (Table~\ref{tab:within}). The probe is well above chance within a question but
below its overall AUROC on every dataset, and answer frequency is about as good. Its overall discrimination
therefore mainly reflects differences between questions, which explains why Max-Probe nearly ties
majority voting (\S\ref{sec:not-selection}).

\begin{table}[h]
\caption{Within-question AUROC on questions with both correct and incorrect samples. Brackets are
95\% question-level bootstrap intervals. ``Overall'' is the probe's ordinary AUROC on all samples.}
\label{tab:within}
\centering\small
\setlength{\tabcolsep}{4pt}
\begin{tabular}{llccccc}
\toprule
Model & Dataset & \# questions & Probe & Answer freq. & Verbal & Overall (probe)\\
\midrule
\multirow{5}{*}{Qwen3-14B}
 & TriviaQA & 266 & .743 [.707,.779] & .706 & .639 & .948\\
 & EntityQ  & 234 & .700 [.660,.738] & .683 & .601 & .913\\
 & NQ-open  & 361 & .587 [.555,.619] & .610 & .558 & .826\\
 & SimpleQA & 158 & .561 [.511,.610] & .512 & .576 & .705\\
 & Pooled   & 1{,}019 & .650 [.631,.669] & .637 & .596 & ---\\
\midrule
\multirow{5}{*}{Ministral-3-8B}
 & TriviaQA & 302 & .723 [.689,.756] & .773 & .585 & .919\\
 & EntityQ  & 175 & .672 [.628,.715] & .704 & .568 & .902\\
 & NQ-open  & 362 & .587 [.557,.618] & .630 & .521 & .816\\
 & SimpleQA & 173 & .555 [.512,.595] & .533 & .487 & .628\\
 & Pooled   & 1{,}012 & .637 [.618,.655] & .669 & .544 & ---\\
\bottomrule
\end{tabular}
\end{table}

\FloatBarrier
\subsection{Question-Level Labels (\probesdq)}
\label{app:qmean}

\probesdq{} (\S\ref{sec:relabel}) keeps the training traces, the in-chain relabeling and the recipe of
\probesd{}, and only replaces each trace's label with its question's mean probe score $\bar p_q$ over
the 20 training samples. Because the coherence filter scans candidates in order of closeness to $\bar p_q$
(\S\ref{sec:relabel}), the per-trace labels already track $\bar p_q$ on Qwen, where the first or
second candidate usually passes (mean rank 1.4, $r{=}.987$, mean absolute difference 2.5 points). On
Ministral they drift more ($r{=}.935$, 7.7 points). Both Ministral variants use the TriviaQA-only
label probe.

\probesdq{} lowers out-of-domain ECE on every dataset, significantly on three of four
datasets for each model, and leaves accuracy unchanged (Table~\ref{tab:qmean}). In-domain ECE falls on
Qwen and is unchanged on Ministral. The gain is stable across seeds: out-of-domain ECE is .101, .098
and .098 on Qwen and .091, .088 and .090 on Ministral. In-chain and final scores stay consistent, with
0.1\% of Qwen responses differing by more than 20 points. Table~\ref{tab:newfull} gives the full
per-dataset results.

\begin{table}[h]
\caption{\probesd{} versus \probesdq{} (three seeds each, seed means). $\Delta$ECE is
\probesdq{} minus \probesd{} with a 95\% paired question-bootstrap interval. AUROC is \probesd{} /
\probesdq{}.}
\label{tab:qmean}
\centering\footnotesize
\setlength{\tabcolsep}{3pt}
\begin{tabular}{llcccc}
\toprule
Model & Dataset & ECE \probesd{} & ECE \probesdq{} & $\Delta$ECE [95\% CI] & AUROC\\
\midrule
\multirow{4}{*}{Qwen3-14B}
 & TriviaQA & .024 & .012 & $-.011$ [$-.015$, $-.003$] & .928 / .930\\
 & EntityQ  & .090 & .077 & $-.013$ [$-.018$, $-.009$] & .881 / .880\\
 & NQ-open  & .137 & .110 & $-.027$ [$-.033$, $-.021$] & .796 / .800\\
 & SimpleQA & .113 & .111 & $-.002$ [$-.007$, $+.003$] & .671 / .673\\
\midrule
\multirow{4}{*}{Ministral-3-8B}
 & TriviaQA & .029 & .031 & $+.001$ [$-.021$, $+.021$] & .887 / .894\\
 & EntityQ  & .105 & .050 & $-.055$ [$-.061$, $-.045$] & .872 / .882\\
 & NQ-open  & .161 & .084 & $-.077$ [$-.084$, $-.069$] & .803 / .800\\
 & SimpleQA & .155 & .134 & $-.021$ [$-.029$, $-.013$] & .674 / .684\\
\bottomrule
\end{tabular}
\end{table}

\begin{table}[h]
\caption{Per-dataset results of the students added in this appendix, in the format of
Tables~\ref{tab:d1}--\ref{tab:d2}. \probesdq{} averages three seeds. ConfTuner is a single run
(Appendix~\ref{app:conftuner}).}
\label{tab:newfull}
\centering\small
\setlength{\tabcolsep}{4pt}
\begin{tabular}{lllcccc}
\toprule
Model & Method & Dataset & Acc $\uparrow$ & ECE $\downarrow$ & AUROC $\uparrow$ & Brier $\downarrow$\\
\midrule
\multirow{8}{*}{Qwen3-14B} & \multirow{4}{*}{\probesdq{}}
   & TriviaQA & \ci{.747}{.025} & \ci{.012}{.006} & \ci{.930}{.010} & \ci{.090}{.008}\\
 & & EntityQ  & \ci{.290}{.025} & \ci{.077}{.015} & \ci{.880}{.015} & \ci{.124}{.009}\\
 & & NQ-open  & \ci{.473}{.029} & \ci{.110}{.020} & \ci{.800}{.021} & \ci{.197}{.012}\\
 & & SimpleQA & \ci{.077}{.012} & \ci{.111}{.010} & \ci{.673}{.037} & \ci{.101}{.008}\\
\cmidrule(lr){2-7}
 & \multirow{4}{*}{ConfTuner}
   & TriviaQA & \ci{.737}{.025} & \ci{.048}{.014} & \ci{.913}{.015} & \ci{.100}{.009}\\
 & & EntityQ  & \ci{.273}{.024} & \ci{.079}{.016} & \ci{.899}{.016} & \ci{.114}{.009}\\
 & & NQ-open  & \ci{.449}{.029} & \ci{.185}{.024} & \ci{.777}{.024} & \ci{.226}{.016}\\
 & & SimpleQA & \ci{.053}{.011} & \ci{.109}{.012} & \ci{.769}{.040} & \ci{.068}{.008}\\
\midrule
\multirow{4}{*}{Ministral-3-8B} & \multirow{4}{*}{\probesdq{}}
   & TriviaQA & \ci{.693}{.024} & \ci{.031}{.013} & \ci{.894}{.013} & \ci{.119}{.008}\\
 & & EntityQ  & \ci{.218}{.024} & \ci{.050}{.012} & \ci{.882}{.017} & \ci{.109}{.009}\\
 & & NQ-open  & \ci{.411}{.027} & \ci{.084}{.019} & \ci{.800}{.019} & \ci{.190}{.011}\\
 & & SimpleQA & \ci{.071}{.012} & \ci{.134}{.011} & \ci{.684}{.041} & \ci{.093}{.008}\\
\bottomrule
\end{tabular}
\end{table}

\FloatBarrier
\subsection{What the Probe Label Contributes}
\label{app:mech}

\textbf{The probe label is not the more accurate estimate of accuracy.} We split each training
question's 20 samples into halves A and B, build a label from $K$ samples of A, and score it against
the accuracy of B. B's own sampling noise adds the same constant to every label's error, so we
subtract its unbiased estimate and obtain the error against the question's true accuracy
(Table~\ref{tab:halfsplit}). Ten-sample empirical accuracy is clearly closer than the ten-sample
probe mean. More probe samples help little, so the probe's error is mostly bias, not noise. The
probe mean is calibrated across questions but compressed toward the middle: 41.6\% of Qwen questions
have a probe mean of at least .95, against 57.5\% for empirical accuracy (Ministral 21.7\% against
51.5\%). Its correlation with empirical accuracy is .87 on Qwen and .85 on Ministral.

\begin{table}[h]
\caption{Error of question-level labels against held-out accuracy (RMSE, half-split of the 20 training
samples per question, sampling noise of the held-out half removed). $K$ is the number of samples the
label uses. $K{=}1$ probe is the per-trace label.}
\label{tab:halfsplit}
\centering\footnotesize
\begin{tabular}{lcccc|cccc}
\toprule
& \multicolumn{4}{c|}{Qwen3-14B} & \multicolumn{4}{c}{Ministral-3-8B}\\
Label & $K{=}1$ & 2 & 5 & 10 & $K{=}1$ & 2 & 5 & 10\\
\midrule
Empirical accuracy & .239 & .171 & .108 & .077 & .260 & .185 & .115 & .084\\
Probe mean         & .229 & .208 & .192 & .187 & .238 & .219 & .205 & .200\\
\bottomrule
\end{tabular}
\end{table}

\textbf{Ordering and distribution.} The two labels differ in which traces they rank higher (ordering)
and in how their values are spread (distribution). We separate the two on Qwen with two controls.
EA-QM keeps the EA-SD traces and the ordering of their labels, but maps the labels by quantile onto
the Probe-SD label distribution. Probe-QM does the reverse. Ties are broken at random, the recipe is
that of the main Qwen students, and each cell averages three seeds (Table~\ref{tab:2x2}).

The probe's distribution, less extreme than empirical accuracy, lowers in-domain ECE under either
ordering. Out of domain, the distribution alone does not help: EA-QM stays near EA-SD. The probe's
ordering does: Probe-QM recovers most of the out-of-domain ECE gap to \probesd{}. AUROC rises with
the probe ordering, and further when the probe's distribution is added. Every seed shows the same
pattern, with out-of-domain ECE of .224--.231 for EA-SD, .232--.234 for EA-QM, .143--.147 for
Probe-QM and .110--.116 for \probesd{}.

\textbf{Learnability.} Why does the less accurate label teach better? The probe is a calibrated
readout of the model's own hidden state, so it estimates accuracy from what that state encodes.
Empirical accuracy tracks true accuracy more closely, but partly through variation the hidden state
does not encode. A student that sees only the question and its trace can learn the first part. The
rest it can only memorize per training question, which does not carry to new questions. Swapping
values is a monotone rescaling of the labels and cannot change what is learnable, so it helps only
where the test distribution matches the training one. To test learnability directly, we ask how
closely each student reproduces its own training labels. On 1,000 training questions shared by EA-SD
and \probesd{}, we sample each student ten times and correlate its mean stated confidence with the
question's mean training label (fit $r$ in Table~\ref{tab:2x2}). Students trained on probe-ordered
labels fit their targets more closely and are better calibrated on these questions. The student
starts from the Base checkpoint, not the model the probe reads, so this account assumes that the two
share the relevant representation. The experiment is consistent with the account but does not prove
it, and it covers Qwen only.

\begin{table}[h]
\caption{Label ordering $\times$ label distribution on Qwen3-14B. Each cell averages seeds 42--44,
with \probesd{} taken from Table~\ref{tab:main}. EA-SD is retrained here and differs slightly from
Table~\ref{tab:main}. Fit $r$ correlates a student's mean stated confidence with its mean training
label on 1,000 training questions (\probesd{}: students retrained in the same environment, which
reproduce Table~\ref{tab:main}).}
\label{tab:2x2}
\centering\small
\setlength{\tabcolsep}{5pt}
\begin{tabular}{lllccccc}
\toprule
& \multicolumn{2}{c}{Label} & \multicolumn{2}{c}{In-domain} & \multicolumn{2}{c}{Out-of-domain} & \\
\cmidrule(lr){2-3}\cmidrule(lr){4-5}\cmidrule(lr){6-7}
Student & Ordering & Distribution & ECE $\downarrow$ & AUROC $\uparrow$ & ECE $\downarrow$ & AUROC $\uparrow$ & Fit $r$ $\uparrow$\\
\midrule
EA-SD      & EA    & EA    & .096 & .882 & .228 & .745 & .858\\
EA-QM      & EA    & probe & .047 & .887 & .233 & .751 & .872\\
Probe-QM   & probe & EA    & .075 & .905 & .145 & .759 & .903\\
\probesd{} & probe & probe & \textbf{.024} & \textbf{.928} & \textbf{.113} & \textbf{.783} & \textbf{.920}\\
\bottomrule
\end{tabular}
\end{table}

\FloatBarrier
\subsection{ConfTuner Baseline}
\label{app:conftuner}

ConfTuner \citep{li2025conftuner} was proposed for instruct models that answer without a chain of
thought, and it optimizes only the final score token with a tokenized Brier loss. We train it on
Qwen3-14B-Instruct with the original Qwen recipe (LoRA $r{=}8$, $\alpha{=}32$ on q/v projections,
learning rate $10^{-5}$, 3 epochs, batch 24) on the model's own judged traces for the 10{,}000
TriviaQA training questions used to fit the probe and the post-hoc calibrators. Its per-dataset
results are in Table~\ref{tab:newfull}.

A reasoning model often also states a confidence in its chain of thought. Table~\ref{tab:conflict}
counts the responses that contain an in-chain confidence number and, among them, those whose in-chain
number differs from the final score by more than 20 points. A typical ConfTuner case ends its
reasoning with ``The confidence here is maybe around 60\%'' and then states 20.

\begin{table}[h]
\caption{Consistency between the confidence stated in the chain of thought and the final score
(Qwen3-14B, all four test sets).}
\label{tab:conflict}
\centering\small
\begin{tabular}{lccc}
\toprule
Model & Responses & With in-chain confidence & Differ by $>$20 points\\
 & & (share of responses) & (share of in-chain)\\
\midrule
Instruct (original) & 34{,}156 & 7{,}818 (22.9\%) & 42 (0.5\%)\\
\probesd{} (seed 42) & 37{,}635 & 8{,}949 (23.8\%) & 13 (0.1\%)\\
Probe-SD-LRM (Instruct init, 3 seeds) & 110{,}757 & 23{,}561 (21.3\%) & 24 (0.1\%)\\
ConfTuner (Instruct init) & 36{,}750 & 8{,}126 (22.1\%) & 2{,}279 (28.0\%)\\
\bottomrule
\end{tabular}
\end{table}

\FloatBarrier
\subsection{An RLCR-Style Reinforcement-Learning Attempt}
\label{app:rl}

Before choosing SFT we tried several GRPO configurations on Qwen3-14B-Instruct with an RLCR-style
reward \citep{damani2025binaryrewardstraininglms}: binary correctness plus a Brier term on the
stated confidence, 32 rollouts per prompt, TriviaQA training questions. None moved the stated
confidence. Table~\ref{tab:rl} shows the longest run (437 steps): mean stated
confidence stays at 0.91--0.93 while rollout accuracy varies between 0.69 and 0.81. From the same
checkpoint, Probe-SD-LRM reaches in-domain ECE .031 with SFT alone. These runs were lightly tuned, so
they show only that calibration did not emerge easily in our setting.

\begin{table}[h]
\caption{Training-rollout calibration of the longest GRPO run, in windows of about 43 steps
(scored rollouts only).}
\label{tab:rl}
\centering\small
\begin{tabular}{lcccccc}
\toprule
Steps & Rollouts & Acc & Mean conf. & ECE & Brier & AUROC\\
\midrule
0--42    & 27{,}535 & .723 & .917 & .194 & .200 & .687\\
43--85   & 27{,}528 & .773 & .921 & .148 & .166 & .704\\
86--128  & 27{,}431 & .758 & .914 & .156 & .172 & .700\\
129--171 & 28{,}004 & .785 & .931 & .146 & .161 & .667\\
172--214 & 27{,}562 & .766 & .925 & .159 & .170 & .701\\
215--257 & 27{,}716 & .809 & .930 & .121 & .138 & .704\\
258--300 & 27{,}625 & .760 & .918 & .157 & .165 & .748\\
301--343 & 27{,}778 & .741 & .919 & .178 & .184 & .711\\
344--386 & 26{,}282 & .688 & .910 & .222 & .223 & .689\\
387--429 & 16{,}794 & .787 & .922 & .135 & .149 & .744\\
430--437 & 1{,}647 & .736 & .929 & .194 & .198 & .664\\
\bottomrule
\end{tabular}
\end{table}

\FloatBarrier
\subsection{Nonlinear Probe}
\label{app:mlp}

We train a one-hidden-layer MLP probe (256 units, early stopping) on the same features and labels as
the linear probe, with weight decay chosen on TriviaQA validation only (Ministral at the L20 MLP
readout of Table~\ref{tab:probe-vs-verbal}). It is not consistently better (Table~\ref{tab:mlp}). Surface-Feat in
Table~\ref{tab:main}, the same logistic readout on text features, is clearly worse than either.

\begin{table}[h]
\caption{Linear versus MLP probe, ECE / AUROC / Brier on the four test sets.}
\label{tab:mlp}
\centering\small
\setlength{\tabcolsep}{4pt}
\begin{tabular}{llcccc}
\toprule
Model & Probe & TriviaQA & EntityQ & NQ-open & SimpleQA\\
\midrule
\multirow{2}{*}{Qwen3-14B} & Linear & .015/.948/.081 & .014/.913/.098 & .079/.826/.177 & .075/.705/.076\\
 & MLP & .024/.946/.082 & .025/.911/.102 & .039/.835/.166 & .072/.745/.071\\
\midrule
\multirow{2}{*}{Ministral-3-8B} & Linear & .023/.919/.105 & .043/.902/.095 & .106/.816/.186 & .145/.628/.110\\
 & MLP & .026/.917/.107 & .065/.897/.101 & .129/.809/.195 & .140/.648/.099\\
\bottomrule
\end{tabular}
\end{table}

\FloatBarrier
\subsection{Zero-Shot Transfer to MATH-500}
\label{app:math}

We evaluate Qwen3-14B on MATH-500 with 10 samples per question at $T{=}0.6$, with no math data in
training and no recalibration (Table~\ref{tab:math}). The original model writes a parsable score on
70\% of its traces (most others write it as \texttt{\{95\}}), so both of its confidence sources are
scored on those traces. The frozen TriviaQA probe still ranks traces well, but its probability scale,
fit at TriviaQA's accuracy of .73, does not transfer to MATH-500's .94. The verbal ECE is low only
because accuracy is near ceiling, so MATH-500 does not test overconfidence for this model. The
\probesd{} student, trained from Base on TriviaQA only, carries the ranking signal into its own verbal
score: its AUROC far exceeds that of the identically trained student without relabeling. We treat
this as a reference point and claim calibration only within factual QA, consistent with
\citet{jang2026confidence}.

\begin{table}[h]
\caption{MATH-500, Qwen3-14B, zero-shot.}
\label{tab:math}
\centering\small
\begin{tabular}{lccccc}
\toprule
Source & Acc & ECE $\downarrow$ & AUROC $\uparrow$ & Brier $\downarrow$ & Cov.\\
\midrule
Instruct, verbal & .944 & .045 & .638 & .052 & .699\\
Instruct, frozen TriviaQA probe (same traces) & .944 & .398 & \textbf{.864} & .236 & .699\\
Student w/o relabel, verbal & .897 & .081 & .718 & .101 & .996\\
\probesd{} student, verbal & .904 & .053 & \textbf{.872} & .077 & .996\\
\bottomrule
\end{tabular}
\end{table}

\FloatBarrier
\subsection{Where the Probe Ordering Helps}
\label{app:bins}

Appendix~\ref{app:mech} shows that the probe's ordering, not its value distribution, carries the
out-of-domain gain. To see on which questions, we bin the 3{,}000 out-of-domain test questions by
their accuracy pooled over the four Qwen students of Table~\ref{tab:2x2} and their three seeds. We
then split the Brier gap between EA-SD and Probe-QM (probe ordering, EA distribution) into per-bin
contributions, each weighted by the bin's share of questions (Table~\ref{tab:bins}).

The gain comes from questions the model never answers correctly, not from mid-accuracy questions,
where sampling noise in the empirical-accuracy label is largest. On high-accuracy questions the EA
ordering is slightly better. The same pattern holds on each out-of-domain dataset separately: the
bin below .05 contributes most of the gap, and the bins above .65 contribute small negative amounts.
In domain the gap is smaller (Brier .121 against .103) and less concentrated, with 49\% coming from
the bin below .05.

The training questions show the same split. Of the 1{,}000 training questions used for the fit
analysis of Appendix~\ref{app:mech}, 104 have empirical accuracy below .05. On them, EA-SD's mean
stated confidence misses its own training label by .27 on average (EA-QM .28), against .14 for
Probe-QM and \probesd{}. Students trained on probe-ordered labels thus learn to say that they do not
know, while EA-SD does not learn to follow its near-zero labels on these questions.

\begin{table}[h]
\caption{Out-of-domain test questions (EntityQ, NQ-open, SimpleQA; Qwen3-14B) binned by accuracy
pooled over all students and seeds. Stated confidence and Brier are EA-SD / Probe-QM / \probesd{}
(three seeds each). The last column is each bin's share of the question-weighted Brier gap between
EA-SD and Probe-QM (.222 against .166).}
\label{tab:bins}
\centering\small
\setlength{\tabcolsep}{4pt}
\begin{tabular}{lccccc}
\toprule
Pooled acc. & Questions & Acc & Stated confidence & Brier & Share of gap\\
\midrule
$<$.05     & 1{,}668 (56\%) & .004 & .292 / .178 / .223 & .219 / .127 / .118 & 92\%\\
.05--.35   & 419 (14\%)     & .157 & .410 / .286 / .321 & .338 / .271 / .234 & 17\%\\
.35--.65   & 201 (7\%)      & .496 & .579 / .464 / .474 & .374 / .376 / .318 & 0\%\\
.65--.95   & 253 (8\%)      & .820 & .718 / .638 / .619 & .239 / .269 / .231 & $-$5\%\\
$\geq$.95  & 459 (15\%)     & .992 & .922 / .895 / .856 & .050 / .065 / .061 & $-$4\%\\
\bottomrule
\end{tabular}
\end{table}

\FloatBarrier
\subsection{Multiple-Choice QA with Shuffled Options}
\label{app:mcqa}

On multiple-choice questions a model can be right by guessing, so empirical accuracy counts lucky
guesses, and the option order can be changed without changing the question. We use this setting to
ask how much stated confidence moves when only the option order changes, and whether the probe label
still teaches better than empirical accuracy.

\textbf{Setup.} We rerun the full pipeline on MMLU \citep{hendrycks2021mmlu} with Qwen3-14B. We keep
40 knowledge subjects and drop 16 computation-heavy subjects and moral\_scenarios, whose four fixed
option combinations cannot be shuffled. After removing duplicate questions, a split stratified by
subject gives 10{,}235 training and 1{,}502 test questions. Each of the 20 training samples per
question ($T{=}1.5$) sees its own random option order, so empirical accuracy includes guessing. The
prompt is that of the main experiments with the options appended, and correctness is read from the
chosen option without a grader. The label probe keeps the readout of Table~\ref{tab:c2} and is refit
on MMLU training samples. It is cross-fitted over two folds of questions, so no trace is scored by a
probe that saw its question. EA-SD, \probesd{} and \probesdq{} share the coherence filter and the
recipe, and both label sets are downsampled to 9{,}500 traces. Each test question is asked in four
cyclic shifts of a random option order, so the correct option appears once in each position, with
five samples per order at $T{=}0.6$. MMLU-Pro \citep{wang2024mmlupro} serves as an out-of-domain
test: 1{,}000 questions from its eight knowledge categories, with ten options, asked in two orders
with five samples each, after removing questions whose normalized text matches a training question.

\textbf{Order sensitivity.} For test question $q$, let $m_{q,j}$ be the mean stated confidence over
the five samples in order $j$, and $w_q$ the mean within-order variance. Even without an order
effect, the order means differ by chance, with expected variance $w_q/5$. Over the $Q$ test
questions, the across-order standard deviation removes this part,
\[
\sigma_{\text{order}}=\sqrt{\frac{1}{Q}\sum_{q}\max\Big(\operatorname{Var}_j(m_{q,j})-\frac{w_q}{5},\,0\Big)},
\]
and the within-order standard deviation $\sigma_{\text{sample}}=\sqrt{\bar w}$, with $\bar w$ the
mean of $w_q$, is the spread across repeated samples of the same prompt.

\textbf{Calibration and stability.} All three students reach the same accuracy, so they differ only
in the confidence they state (Table~\ref{tab:mcqa}). \probesd{} lowers ECE about five-fold relative to
EA-SD on MMLU and two-fold on MMLU-Pro, and raises AUROC. Its confidence moves half as much across
option orders, and its spread across repeated samples of the same prompt is less than half that of
EA-SD, so EA-SD's confidence is unstable even when nothing in the prompt changes. The original model's
verbal score barely moves across orders, but it is near .9 for almost every answer. \probesdq{}
matches \probesd{}. Across seeds, the standard deviation of every student metric is at most .004.
Ministral-3-8B (one seed, label probe of Table~\ref{tab:c2}) shows the same pattern.

\textbf{Guessing and knowing.} We group each model's test questions by its majority answer in each
option order (Table~\ref{tab:mcqa-groups}). Where the majority answer changes with the order
(accuracy about .44), all students state far less than the original model's .85: EA-SD .51 and
\probesd{} .62. EA-SD's mean is closer to the accuracy there, yet \probesd{} has the lower Brier in
every accuracy bin, by the largest margin at mid accuracy (Table~\ref{tab:mcqa-bins}). The
difference lies in the spread: EA-SD's statements scatter widely around the right mean
($\sigma_{\text{sample}}$ .198), which Brier penalizes. Unlike factual QA (Appendix~\ref{app:bins}),
the gain extends to mid-accuracy questions, where guessing is possible.

\textbf{Better estimate, worse teacher.} As on TriviaQA (Table~\ref{tab:halfsplit}), ten-sample
empirical accuracy tracks held-out accuracy far more closely than the ten-sample probe mean (RMSE
.066 against .255 on Qwen and .076 against .233 on Ministral, half-split of the 20 training samples),
yet it trains the worse student. Noise in the empirical-accuracy label therefore does not explain
the gap.

\textbf{Limit.} Where the model gives the same wrong answer in every order (accuracy .04--.07),
every confidence source stays high: .65--.89, with the probe at .67. On training questions that the
model never answers correctly, the mean probe score is .71. The probe does not detect confident
misconceptions, and neither do the students trained on it.

\begin{table}[h]
\caption{Multiple-choice QA. Qwen students average three seeds, Ministral uses one seed.
$\sigma_{\text{order}}$: standard deviation of stated confidence across option orders, net of
sampling noise. $\sigma_{\text{sample}}$: standard deviation across samples of the same prompt. The
original model's verbal and probe rows are scored on traces with a parsable score and with
extracted features, respectively, so their accuracies differ slightly.}
\label{tab:mcqa}
\centering\small
\setlength{\tabcolsep}{4pt}
\begin{tabular}{llcccccc}
\toprule
Setting & Source & Acc & ECE $\downarrow$ & AUROC $\uparrow$ & Brier $\downarrow$ & $\sigma_{\text{order}}$ $\downarrow$ & $\sigma_{\text{sample}}$\\
\midrule
\multirow{5}{*}{\makecell[l]{Qwen3-14B\\MMLU}}
 & Original, verbal & .836 & .085 & .729 & .130 & .018 & .042\\
 & Original, probe  & .837 & .030 & .857 & .103 & .051 & .084\\
 & EA-SD            & .837 & .109 & .783 & .137 & .072 & .198\\
 & \probesd{}       & .841 & .023 & .854 & .102 & .036 & .088\\
 & \probesdq{}      & .841 & .023 & .856 & .102 & .035 & .085\\
\midrule
\multirow{5}{*}{\makecell[l]{Qwen3-14B\\MMLU-Pro}}
 & Original, verbal & .805 & .127 & .697 & .160 & .016 & .040\\
 & Original, probe  & .806 & .089 & .828 & .137 & .066 & .098\\
 & EA-SD            & .804 & .130 & .773 & .149 & .085 & .203\\
 & \probesd{}       & .806 & .058 & .843 & .122 & .043 & .097\\
 & \probesdq{}      & .807 & .057 & .848 & .120 & .041 & .094\\
\midrule
\multirow{5}{*}{\makecell[l]{Ministral-3-8B\\MMLU}}
 & Original, verbal & .803 & .117 & .639 & .160 & .011 & .035\\
 & Original, probe  & .827 & .032 & .836 & .112 & .037 & .080\\
 & EA-SD            & .815 & .113 & .722 & .148 & .050 & .153\\
 & \probesd{}       & .818 & .027 & .817 & .119 & .025 & .067\\
 & \probesdq{}      & .817 & .025 & .815 & .119 & .026 & .066\\
\bottomrule
\end{tabular}
\end{table}

\begin{table}[h]
\caption{Mean stated confidence (accuracy in parentheses) on MMLU test questions grouped by each
model's majority answer per option order (Qwen3-14B, seed 42). Stable: the same answer in all four
orders.}
\label{tab:mcqa-groups}
\centering\small
\begin{tabular}{lccc}
\toprule
Source & Stable, correct & Order-dependent & Stable, wrong\\
\midrule
Original, verbal & .938 (.987) & .848 (.420) & .889 (.044)\\
Original, probe  & .909 (.987) & .573 (.420) & .674 (.044)\\
EA-SD            & .922 (.984) & .514 (.440) & .652 (.066)\\
\probesd{}       & .908 (.983) & .622 (.450) & .694 (.060)\\
\probesdq{}      & .909 (.983) & .619 (.456) & .702 (.062)\\
\bottomrule
\end{tabular}
\end{table}

\begin{table}[h]
\caption{MMLU test questions binned by accuracy pooled over the three students (Qwen3-14B, seed 42).
Stated confidence and Brier are EA-SD / \probesd{}.}
\label{tab:mcqa-bins}
\centering\small
\begin{tabular}{lcccc}
\toprule
Pooled acc. & Questions & Acc & Stated confidence & Brier\\
\midrule
$<$.05     & 83      & .006 & .675 / .708 & .597 / .540\\
.05--.35   & 102     & .178 & .534 / .625 & .449 / .395\\
.35--.65   & 81      & .502 & .495 / .607 & .387 / .276\\
.65--.95   & 171     & .821 & .625 / .682 & .264 / .166\\
$\geq$.95  & 1{,}065 & .997 & .951 / .935 & .032 / .016\\
\bottomrule
\end{tabular}
\end{table}

\clearpage
\section{Limitations}
\label{app:lim}

This appendix expands the limitations summarized in \S\ref{sec:conclusion}.

\textbf{Scope.} We validate on two open-weight reasoning families of at most 14B parameters and
four factual QA benchmarks with short, checkable answers. Larger or closed models are untested, as
is long-form or subjective generation---hard for a structural reason, since probe training needs
stable trace-level correctness labels that such tasks do not supply. Replacing binary correctness
with graded factuality is the natural next step. Zero-shot on MATH-500 the probe's ranking transfers
but its scale does not (Appendix~\ref{app:math}), so we claim calibration only within factual QA.
On multiple-choice knowledge QA the method carries over, but neither the probe nor the student flags
answers that the model gives consistently and wrongly (Appendix~\ref{app:mcqa}).
Our RL attempt was lightly tuned (Appendix~\ref{app:rl}), so we do not compare against RL methods on
benefit or on training cost.

\textbf{White-box access.} The deployed student needs no probe, but constructing its training data
requires hidden states for candidate traces and a per-model choice of layer and readout. The
pipeline therefore cannot be built from a text-only interface.

\textbf{Low base-rate tasks.} As accuracy falls the probe keeps its probability scale but loses
ranking power (\S\ref{sec:probe-calibrated}), matching the general difficulty of uncertainty
estimation on knowledge-intensive tasks \citep{tao2025revisitinguncertaintyestimationcalibration}.
Calibration also acts only on candidates that were generated: if the pool contains no correct
answer, no confidence estimate can create one. The ceiling on such tasks is set on the generation
side, and calibration has to be combined with retrieval or stronger generation rather than used in
its place.

\textbf{Grader dependence.} Probe labels and correctness metrics use Qwen3-32B-Instruct. A human
audit confirmed 199/200 extractions and 198/199 answerable grading decisions. Correcting two labels
in a separate 500-trace audit left Probe ECE unchanged and changed AUROC and Brier by less than
0.001 (Appendix~\ref{app:judge}). These checks make large grading artifacts unlikely, but cannot
rule out errors correlated with probe errors, such as shared failures on long-tail entities or
ambiguous questions.

\end{document}